\documentclass{article} 
\usepackage{iclr2027_conference,times}

\usepackage{amsmath,amsfonts,bm}

\def\eqref#1{equation~\ref{#1}}

\def\1{\bm{1}}

\DeclareMathAlphabet{\mathsfit}{\encodingdefault}{\sfdefault}{m}{sl}
\SetMathAlphabet{\mathsfit}{bold}{\encodingdefault}{\sfdefault}{bx}{n}

\usepackage{amsthm}

\usepackage{hyperref}
\usepackage{array}
\usepackage{tabularx}
\usepackage{graphicx}   
\usepackage[labelformat=parens,labelsep=space,font=footnotesize,skip=1pt]{subcaption}
\usepackage{url}
\usepackage{booktabs}
\usepackage{multirow}
\usepackage{xcolor}
\usepackage{colortbl}   
\definecolor{cNone}{HTML}{F2F2F2}   
\definecolor{cChunk}{HTML}{E8F1FA}  
\definecolor{cGraph}{HTML}{E6F4EA}  
\definecolor{cAgGr}{HTML}{EEF7E0}   
\definecolor{cKGC}{HTML}{FFF6DB}    
\definecolor{cHyper}{HTML}{FDECE0}  
\definecolor{cMem}{HTML}{F3EAF7}    
\definecolor{cEvo}{HTML}{FBE9EF}    
\definecolor{cAgKE}{HTML}{E4F4F4}   
\definecolor{cOmni}{HTML}{ECEBFA}   
\newif\ifrevision\revisionfalse
\definecolor{rvCommon}{HTML}{D50000}
\definecolor{rvNJ}{HTML}{9C27B0}
\definecolor{rvPV}{HTML}{1E8E3E}
\definecolor{rvFu}{HTML}{1565C0}
\newcommand{\rC}[1]{\ifrevision{\color{rvCommon}#1}\else#1\fi}
\newcommand{\rF}[1]{\ifrevision{\color{rvFu}#1}\else#1\fi}
\newcommand{\rP}[1]{\ifrevision{\color{rvPV}#1}\else#1\fi}
\newcommand{\rN}[1]{\ifrevision{\color{rvNJ}#1}\else#1\fi}
\definecolor{cAbl}{HTML}{EDEDED}    
\definecolor{cOurs}{HTML}{D6E9FF}   
\definecolor{cOKE}{gray}{0.93}      
\usepackage{enumitem}
\usepackage{wrapfig}
\usepackage{pifont}    
\usepackage{algorithm}
\usepackage{algpseudocode}

\usepackage[T1,OT1]{fontenc}
\usepackage[skins]{tcolorbox}
\usepackage{float}      
\definecolor{pmCa}{HTML}{FBE9DF}\definecolor{pmCaF}{HTML}{8C3F22}   
\definecolor{pmCc}{HTML}{E3EDF8}\definecolor{pmCcF}{HTML}{25507F}   
\definecolor{pmSlot}{HTML}{A1165B}                                  
\newcommand{\promptfont}{\fontencoding{T1}\fontfamily{phv}\fontsize{7.2}{8.7}\selectfont
  \raggedright\setlength{\parindent}{0pt}\setlength{\parskip}{0pt}\hyphenpenalty=10000\exhyphenpenalty=10000}
\newtcolorbox{promptbox}[3]{enhanced,colback=#1,colframe=#2,coltitle=white,title={#3},
  fonttitle=\sffamily\bfseries\scriptsize,boxrule=0.5pt,arc=1.2pt,left=4pt,right=4pt,top=2.5pt,bottom=2.5pt,
  toptitle=0.6pt,bottomtitle=0.6pt,shadow={0.8pt}{-0.8pt}{0pt}{black!28},fontupper=\promptfont,before skip=0pt,after skip=0pt}
\newcommand{\slot}[1]{{\color{pmSlot}\bfseries\{#1\}}}
\newcommand{\media}[1]{{\color{pmSlot}\bfseries$\langle$#1$\rangle$}}
\newcommand{\pgap}{\par\vspace{2.6pt}}
\newcommand{\pj}[2]{\par\hangindent=\dimexpr#1+1.6em\relax\hangafter=1\hspace*{#1}#2}

\usepackage{fontawesome5}
\usepackage{tikz}
\definecolor{tgMod}{HTML}{4F46B8}    
\definecolor{tgStr}{HTML}{0B7E80}    
\definecolor{tgAg}{HTML}{E0663D}     
\colorlet{tgM1}{tgMod!70}\colorlet{tgM2}{tgMod!85}\colorlet{tgM3}{tgMod}
\colorlet{tgS0}{black!45}\colorlet{tgS1}{tgStr!75}\colorlet{tgS2}{tgStr!88}\colorlet{tgS3}{tgStr}
\colorlet{tgA0}{black!38}
\DeclareRobustCommand{\tgbadge}[2]{\tikz[baseline=-0.62ex]{\node[inner sep=0pt,minimum width=0.98em,minimum height=0.98em,rounded corners=1.6pt,fill=#1!13,text=#1]{\scalebox{0.64}{#2}};}}
\newcommand{\tgglyphN}{\tikz[line width=0.55pt]{\draw (0,0) circle (0.12);\draw (-0.085,-0.085)--(0.085,0.085);}}
\newcommand{\tgglyphC}{\tikz[line width=0.45pt]{\foreach \x in {0,0.13}\foreach \y in {0,0.13}\draw[fill=.!30] (\x,\y) rectangle ++(0.1,0.1);}}
\newcommand{\tgglyphG}{\tikz[line width=0.6pt]{\draw (0,0)--(0.13,0.22)--(0.26,0)--cycle;\foreach \p in {(0,0),(0.13,0.22),(0.26,0)}\fill \p circle (0.048);}}
\newcommand{\tgglyphH}{\tikz[line width=0.55pt]{\draw[fill=.!35,rounded corners=2.5pt] (-0.06,-0.05)--(0.13,0.28)--(0.32,-0.05)--cycle;\foreach \p in {(0.04,0.01),(0.13,0.16),(0.22,0.01)}\fill \p circle (0.045);}}
\DeclareRobustCommand{\tmT}{\tgbadge{tgM1}{\faAlignLeft}}      
\DeclareRobustCommand{\tmM}{\tgbadge{tgM2}{\faImage}}          
\DeclareRobustCommand{\tmO}{\tgbadge{tgM3}{\faPhotoVideo}}     
\DeclareRobustCommand{\tsN}{\tgbadge{tgS0}{\tgglyphN}}         
\DeclareRobustCommand{\tsC}{\tgbadge{tgS1}{\tgglyphC}}         
\DeclareRobustCommand{\tsG}{\tgbadge{tgS2}{\tgglyphG}}         
\DeclareRobustCommand{\tsH}{\tgbadge{tgS3}{\tgglyphH}}         
\DeclareRobustCommand{\taY}{\tgbadge{tgAg}{\faToggleOn}}
\DeclareRobustCommand{\taN}{\tgbadge{tgA0}{\faToggleOff}}
\DeclareRobustCommand{\tg}[3]{#1\hspace{0.8pt}#2\hspace{0.8pt}#3\,}

\title{{\spaceskip=0.2em plus 0.05em minus 0.05em KnowledgeClaw:\ Towards Agentic Knowledge}\\ Engineering via Omnimodal Hypergraphs}

\author{{\bf Haoran Luo\textsuperscript{\rm 1}\textsuperscript{$\dagger$}, Jiapu Wang\textsuperscript{\rm 2}\textsuperscript{$*$}, Linhao Luo\textsuperscript{\rm 3}, Luu Anh Tuan\textsuperscript{\rm 1,4}}\\
         \textsuperscript{1}Nanyang Technological University, Singapore\\
         \textsuperscript{2}Nanjing University of Science and Technology\ \ \textsuperscript{3}Monash University\ \ \textsuperscript{4}VinUniversity\\
         {\textsuperscript{$\dagger$}Project Lead.\ \ \textsuperscript{$*$}Corresponding Author.}\\
         {\small\texttt{haoran.luo@ntu.edu.sg, jiapu.wang@njust.edu.cn, anhtuan.luu@ntu.edu.sg}}}

\iclrfinalcopy 
\begin{document}

\maketitle

\begin{abstract}
Knowledge engineering (KE) turns raw files into a reasoning-ready knowledge base, a task now within reach of LLM agents.
KE has progressed from chunk and graph retrieval-augmented generation (RAG) to memory agents, agentic KE and multimodal KE.
However, a live knowledge base needs omnimodality, a closed-loop lifecycle and multi-granular application at once, yet existing systems meet at most two and no benchmark measures all three.
We formalize \emph{Omnimodal Knowledge Engineering} (OmniKE) and release \emph{OmniKE-Bench}, six tasks with $1{,}829$ model-free items over $3{,}960$ real files in five modalities.
We further propose \textbf{\mbox{KnowledgeClaw}}, an agentic system that builds an omnimodal typed hypergraph, keeps it current with a closed-loop lifecycle agent and routes each query to the granularity it needs.
Against $30$ baselines, it ranks first on all six tasks with an overall OKE of $55.9$, $+14.6$ over the strongest baseline.
Our software and data are publicly available at\ \url{https://github.com/LHRLAB/KnowledgeClaw}.
\end{abstract}

\section{Introduction}
\label{sec:intro}

\begin{wrapfigure}[19]{r}{0.5\textwidth}
\vspace{-4.5mm}
\centering
\includegraphics[width=0.5\textwidth]{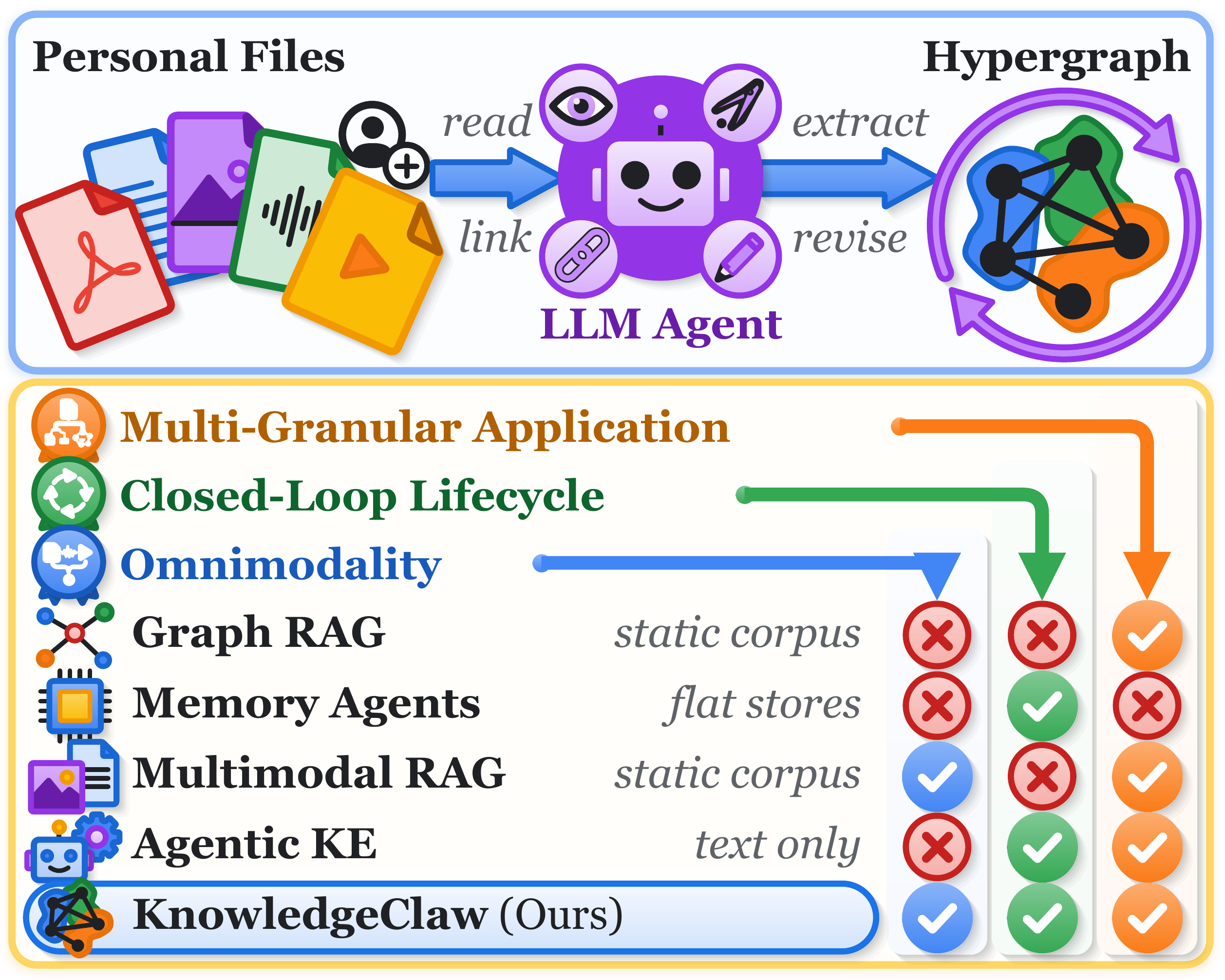}
\setlength{\abovecaptionskip}{5pt}
\caption{Existing KE systems each miss at least one of the three properties that a live knowledge base needs, while the proposed KnowledgeClaw achieves all three of the above at the same time.}
\label{fig:task}
\vspace{5pt}
\vspace{-2mm}
\end{wrapfigure}

Knowledge engineering (KE) distills raw files into a reasoning-ready knowledge base~\citep{KEclassic}, but manual construction long kept it from scaling~\citep{KGLLMsurvey}.
LLM agents lift this limit~\citep{GPT4,ReAct,AgentSurvey}: an agent can read files, extract entities, link records and revise its store~\citep{LLMIEsurvey}, curating all of a person's growing files into a single live knowledge base.

Knowledge structures grew from passages in RAG~\citep{RAG} to entity graphs in GraphRAG~\citep{GraphRAG} and $n$-ary hyperedges in HyperGraphRAG~\citep{HyperGraphRAG}. Memory agents~\citep{MemGPT} keep a writable store that MemPro~\citep{MemPro} evolves, agentic KE~\citep{OpenClaw} lets one agent build and keep it, and Omni-SimpleMem~\citep{Omni-SimpleMem} reaches audio and video. Yet no system unites them all.

\begin{figure}[t]
\centering
\includegraphics[width=\textwidth]{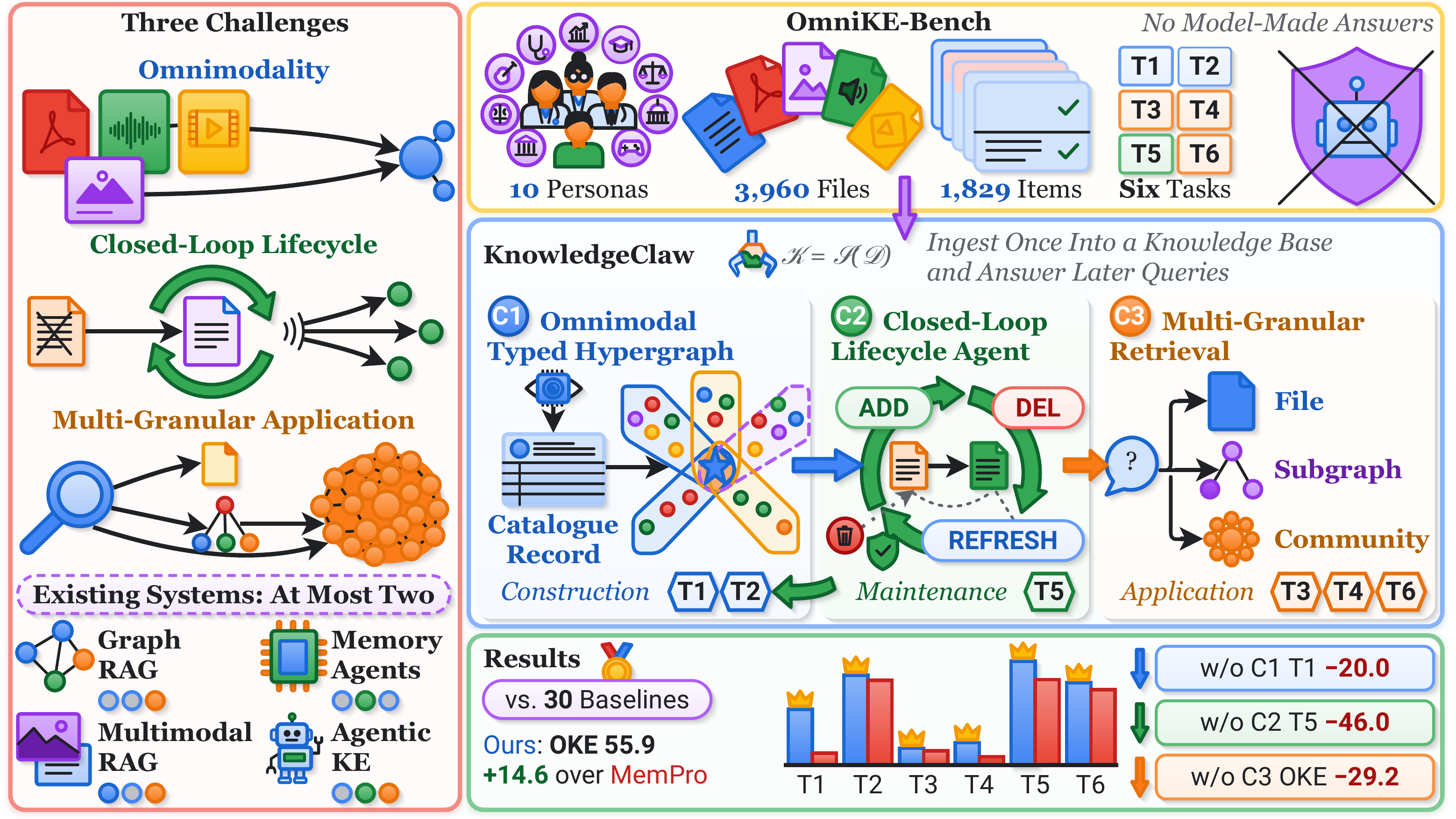}
\setlength{\abovecaptionskip}{5pt}
\caption{Overview of KnowledgeClaw on OmniKE. C1 builds an omnimodal hypergraph (T1, T2), C2 maintains it (T5) and C3 routes each query (T3, T4, T6); it leads on all six tasks (OKE\,$=$\,55.9).}
\label{fig:overview}
\vspace{5pt}
\end{figure}

A personal knowledge base needs all of this at once: papers arrive as PDFs, talks as audio or video and notes as text; papers are revised or withdrawn; queries range from one-file lookups to multi-file joins and topic synthesis.
This raises three challenges (Figure~\ref{fig:task}); each existing line meets at most two.
\textbf{(i) Omnimodality:} an entity in a PDF and a recording should be one node, yet only multimodal RAG reads beyond text.
\textbf{(ii) Closed-loop lifecycle:} every edit should reach everything derived from its file, yet graph and multimodal RAG index a static corpus.
\textbf{(iii) Multi-granular application:} each query should use the file, subgraph or community level, yet memory agents keep flat stores.

No benchmark measures all three, so we formalize \emph{Omnimodal Knowledge Engineering} (\textbf{OmniKE}): a system ingests a person's files once into a knowledge base and answers later queries from it.
We release \textbf{OmniKE-Bench}, six tasks on construction (T1, T2), application (T3, T4, T6) and maintenance (T5).
Its $1{,}829$ model-free items cover $3{,}960$ real files in five modalities from ten researchers.

On top of OmniKE-Bench, we propose \textbf{KnowledgeClaw}, an agentic system of three components (Figure~\ref{fig:overview}).
\textbf{(C1) Omnimodal typed hypergraph} \rC{reads each file natively into a catalogue record and merges all records into one hypergraph, where files of any modality meet at shared entities.}
\textbf{(C2) Closed-loop lifecycle agent} \rF{tracks every node's source files and commits each \textsc{Add}, \textsc{Del} or \textsc{Refresh} along them, so no stale knowledge survives an edit.}
\textbf{(C3) Multi-granular retrieval with intent routing} draws on just the tier that each query needs, so all named files enter the evidence.

Against $30$ baselines in ten groups, KnowledgeClaw ranks first on all six tasks with an overall OKE of \textbf{55.9}, \textbf{+14.6} over the strongest baseline, MemPro. Removing C1, C2 or C3 lowers T1 by $20.0$, T5 by $46.0$ or OKE by $29.2$. Even with only $256$ evidence tokens, it beats every full-context baseline.

\section{Related Work}
\label{sec:related_work}

Our setting joins the two lines of work reviewed below, and Appendix~\ref{app:extended_rw} surveys both lines in full.

\textbf{LLM-based knowledge engineering.}
KE structures grew from chunks (StandardRAG, \citealp{RAG}; RAPTOR, \citealp{RAPTOR}) to entity graphs (GraphRAG, \citealp{GraphRAG}; LightRAG, \citealp{LightRAG}; HippoRAG\,2, \citealp{HippoRAG2}), agentic graph RAG (KAG, \citealp{KAG}; Youtu-GraphRAG, \citealp{Youtu-GraphRAG}), curated KGs (iText2KG, \citealp{iText2KG}; OntoKG, \citealp{OntoKG}) and hypergraphs (HyperGraphRAG, \citealp{HyperGraphRAG}; Hyper-KGGen, \citealp{Hyper-KGGen}; HyperRAG, \citealp{HyperRAG}).
Memory agents (MemGPT, \citealp{MemGPT}; A-MEM, \citealp{A-MEM}; Mem0, \citealp{Mem0}; Graphiti, \citealp{Zep}; MIRIX, \citealp{MIRIX}; MemOS, \citealp{MemOS}), self-evolving systems (MemEvolve, \citealp{MemEvolve}; ATOM, \citealp{ATOM}; MemPro, \citealp{MemPro}) and agentic KE (OpenClaw, \citeyear{OpenClaw}; Graphify, \citeyear{Graphify}; Cognee, \citealp{Cognee}; LLM Wiki, \citealp{LLMWiki}; OKF, \citealp{OKF}) \rF{make the store writable, but edit single facts or pages, not everything derived from a changed source file.}

\textbf{Multimodal retrieval-augmented generation.}
Multimodal RAG extends this line to visual--textual graphs (MMGraphRAG, \citealp{MMGraphRAG}), audio-visual KGs (M$^3$KG-RAG, \citealp{M3KG-RAG}) and multimodal memories (Omni-SimpleMem, \citealp{Omni-SimpleMem}), but none maintains its store.
\mbox{KnowledgeClaw} joins both lines in one omnimodal hypergraph kept consistent under file edits.

\clearpage

\begin{figure}[t]
\centering
\includegraphics[width=\textwidth]{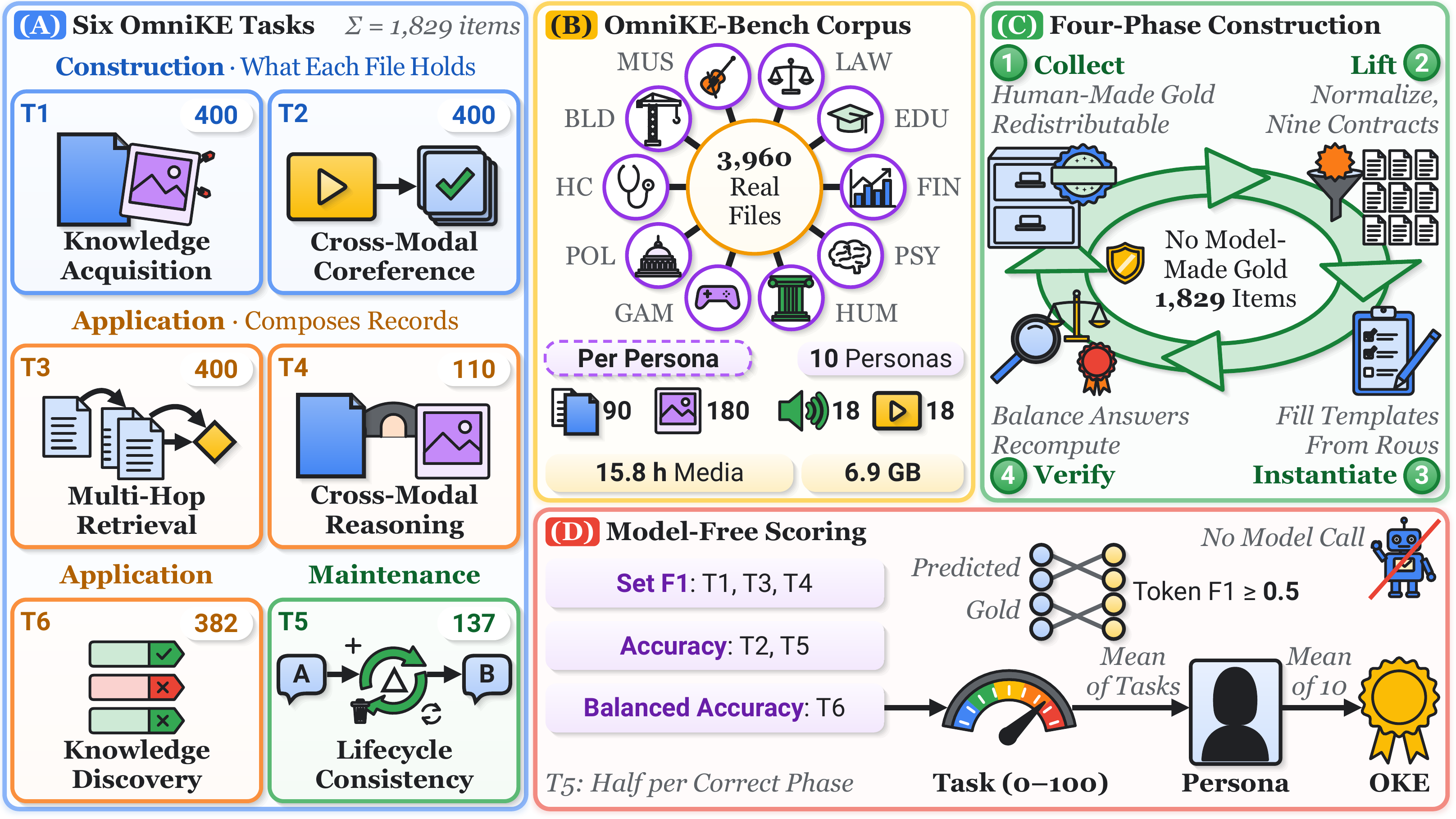}
\setlength{\abovecaptionskip}{5pt}
\caption{\textbf{(A)} Six OmniKE tasks in three groups: construction (T1, T2), application (T3, T4, T6) and maintenance (T5). \textbf{(B)} OmniKE-Bench spans ten personas, five modalities and $3{,}960$ real files. \textbf{(C)} Four phases build its $1{,}829$ items with no model-made gold. \textbf{(D)} Model-free scoring and OKE.}
\label{fig:omnike}
\vspace{5pt}
\end{figure}

\section{The OmniKE Tasks and Benchmark}
\label{sec:bench}

\textbf{Task definition.}\label{sec:taskdef}
A \emph{persona} is one researcher's file collection $\mathcal{D}$ over the modalities
$\mathcal{M}=\{\textit{text},\textit{pdf},\textit{image},\textit{audio},\textit{video}\}$. A system $\mathcal{S}$ reads $\mathcal{D}$ once, at
\textsc{ingest}, and commits an artefact $\mathcal{K}=\mathcal{S}(\mathcal{D})$, its knowledge store; every later query is answered through $\mathcal{K}$
alone, each in a fresh context. The six tasks $\mathcal{T}_1$--$\mathcal{T}_6$ (Figure~\ref{fig:omnike}A, Table~\ref{tab:tasks}) form three
groups, one per property. \emph{Construction} asks whether $\mathcal{K}$ captures what each file holds: extract a file's
entities of given types ($\mathcal{T}_1$) and match a media file to its authoritative description ($\mathcal{T}_2$).
\emph{Application} asks whether $\mathcal{K}$ composes records: answer 2--4-hop joins ($\mathcal{T}_3$), joins across modalities
($\mathcal{T}_4$) and a file's controlled-vocabulary labels ($\mathcal{T}_6$). \emph{Maintenance} ($\mathcal{T}_5$) asks whether $\mathcal{K}$
absorbs an edit $\Delta\in\{\textsc{Add},\textsc{Del},\textsc{Refresh}\}$: the system answers, commits the
next version as a \textsc{Refresh} that changes the answer, and answers it again.

\textbf{Benchmark and statistics.}\label{sec:dataset}
OmniKE-Bench holds ten personas from healthcare to games, \rC{each with $90$ documents as PDF and text, $180$ images, $18$ audio and $18$ video clips}: $3{,}960$ files,
$15.8$\,h of media and $6.9$\,GB in all (Figure~\ref{fig:omnike}B, Table~\ref{tab:corpus}). Its $1{,}829$ items comprise
$400$ each for $\mathcal{T}_1$--$\mathcal{T}_3$, $110$ for $\mathcal{T}_4$, $137$ for $\mathcal{T}_5$ and $382$ for $\mathcal{T}_6$, and,
unlike prior omnimodal benchmarks, \rC{every answer traces to human-made metadata, deterministic parsing, official
transcripts or version chains, never to a model's output.}

\textbf{Construction.}\label{sec:construction}
Four phases (Figure~\ref{fig:omnike}C) follow three rules: \rF{answers come only from authoritative structure,
a model only paraphrases, and no human edits or picks an item.} \rC{\emph{Collect} admits redistributable sources with
human-made annotation, so ASR, OCR and model descriptions may be system input but never gold.} \emph{Lift} normalizes every
file and maps each source onto nine canonical contracts (records, alignments, events, \dots), so each task is written once
for all personas. \emph{Instantiate} fills templates from contract rows; \rF{GPT-5-nano only
rewrites a stem into prose, and deterministic checks discard any item whose slots changed or whose stem leaks the answer.}
\emph{Verify} balances answers so that priors carry no signal and recomputes them with an independent implementation that agrees on all $1{,}829$ items,
so every answer is reproducible from the raw source files alone (Appendix~\ref{app:bench-build}).

\textbf{Evaluation metrics.}\label{sec:metrics}
\rF{Scoring calls no model (Figure~\ref{fig:omnike}D). Free-form answers ($\mathcal{T}_1,\mathcal{T}_3,\mathcal{T}_4$) are matched
to gold values one-to-one and leniently (a date at the gold's precision in any format, any other value by normalized
equality or a word-token F1 of at least $0.5$) and scored by the set F1 of the matches.} $\mathcal{T}_2$ and $\mathcal{T}_5$ are four-way
choices, a $\mathcal{T}_5$ item scoring half per correct phase, and $\mathcal{T}_6$ scores balanced accuracy over positive and
negative candidates. A task score is the mean item score ($0$--$100$), a persona score the mean over the tasks it holds,
and OKE the mean of the ten persona scores, so each persona weighs the same whatever its number of items
(full definitions in Appendix~\ref{app:bench-metric}).

\clearpage

\begin{figure}[t]
\centering
\includegraphics[width=\textwidth]{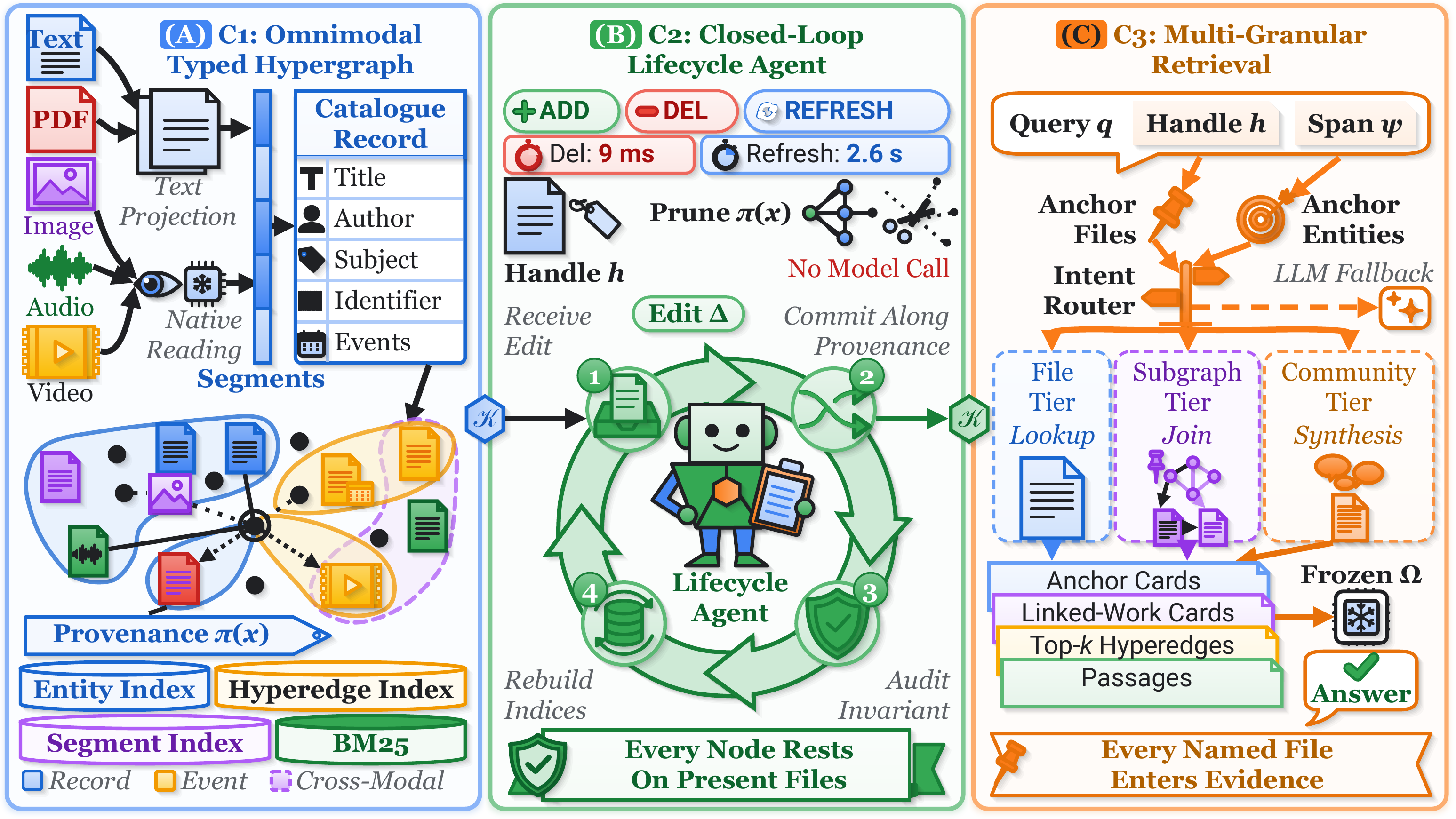}
\setlength{\abovecaptionskip}{5pt}
\caption{\textbf{KnowledgeClaw.} \textbf{(A)} C1 merges catalogue records into one omnimodal typed hypergraph $\mathcal{K}$. \textbf{(B)} C2 commits each edit in place along provenance. \textbf{(C)} C3 routes each query to the right tier.}
\label{fig:method}
\vspace{5pt}
\end{figure}

\section{Methodology: KnowledgeClaw}
\label{sec:method}
\setlength{\abovedisplayskip}{6pt plus 2pt minus 2pt}
\setlength{\belowdisplayskip}{6pt plus 2pt minus 2pt}

KnowledgeClaw (Figure~\ref{fig:method}) realizes the system $\mathcal{S}$ of \S\ref{sec:taskdef} with three components,
one per task group: C1 reads a persona $\mathcal{D}$ once into an omnimodal
typed hypergraph $\mathcal{K}$ (\S\ref{sec:c1}), C2 commits each edit $\Delta$ to $\mathcal{K}$ along source files
(\S\ref{sec:c2}), and C3 answers each query $q$ from just the tier of $\mathcal{K}$ it needs (\S\ref{sec:c3}).

\subsection{C1: Omnimodal Typed Hypergraph}
\label{sec:c1}

\textbf{The artefact.} Nodes are the files $a\in\mathcal{D}$ and their entities $v\in\mathcal{V}$; a
hyperedge $e\in\mathcal{E}$ joins nodes and is typed by a kind $\kappa(e)$, and the provenance $\pi(x)$ gives the source files of an entity or hyperedge $x$:
\begin{equation}
\mathcal{K}=(\mathcal{D},\mathcal{V},\mathcal{E},\pi),\quad
\mathcal{E}\subseteq2^{\mathcal{D}\cup\mathcal{V}},\quad
\kappa:\mathcal{E}\rightarrow\{\textit{record},\textit{event},\textit{cross-modal}\},\quad
\pi:\mathcal{V}\cup\mathcal{E}\rightarrow2^{\mathcal{D}}.
\label{eq:artefact}
\end{equation}

\textbf{Native reading.} A file $a$ of modality $\mu(a)\in\mathcal{M}$ becomes a text surface $\tilde a$: text and
PDF via a shared text projection $\mathrm{proj}$, media natively via the frozen omnimodal model $\Omega$ on the raw
file $\mathrm{raw}(a)$ under a prompt $p_{\mu(a)}$ (prompts $p_{\cdot}$ in Appendix~\ref{app:prompt}); $\mathrm{split}$ cuts $\tilde a$ into $n_a$ segments $\sigma_{a,j}$ of ${\sim}1{,}000$ tokens:
\begin{equation}
\tilde a=
\begin{cases}
\mathrm{proj}(a), & \mu(a)\in\{\textit{text},\textit{pdf}\},\\
\Omega\bigl(\mathrm{raw}(a);\,p_{\mu(a)}\bigr), & \mu(a)\in\{\textit{image},\textit{audio},\textit{video}\},
\end{cases}
\qquad
(\sigma_{a,1},\dots,\sigma_{a,n_a})=\mathrm{split}(\tilde a).
\label{eq:read}
\end{equation}

\textbf{Catalogue records and identity.} One call $\Omega(\tilde a;p_{\mathrm{rec}})$ writes the catalogue record of $a$:
dated events and a value set $Z_{a,\gamma}$ per field $\gamma$ of \rN{a fixed schema $\Gamma$ (title, author,
subject, id, \dots)}. A value $z$ becomes the entity $\iota(z)$, each field a \emph{record}
hyperedge and each event an \emph{event} hyperedge $\{a\}$ with its date:
\begin{equation}
\begin{gathered}
e_{a,\gamma}=\{a\}\cup\bigl\{\iota(z):z\in Z_{a,\gamma}\bigr\},\quad
\pi(e_{a,\gamma})=\{a\},\quad
\pi(v)\leftarrow\pi(v)\cup\{a\}\ \ \forall\,v\in e_{a,\gamma}\cap\mathcal{V},\\
\iota(z)=\mathrm{norm}(z),\qquad
\mathcal{V}(a)=\textstyle\bigcup_{\gamma\in\Gamma}\bigl(e_{a,\gamma}\cap\mathcal{V}\bigr),\qquad
[v]=\bigl\{v'\in\mathcal{V}:\iota_{\mathrm{p}}(v')=\iota_{\mathrm{p}}(v)\bigr\},
\end{gathered}
\label{eq:record}
\end{equation}
where $\mathrm{norm}$ case-folds a value and trims its edge punctuation, so records meet on shared entities, and the
name key $\iota_{\mathrm{p}}$ (first initial, last name) puts ``M.~Langtry'' and ``Max Langtry'' in one alias class.

\textbf{Cross-modal coreference.} \rC{An entity whose files span ${\ge}2$ modalities ($\mu$ applied elementwise) gets a
\emph{cross-modal} hyperedge}, and files of different modalities sharing a title or identifier are \emph{twins}:
\begin{equation}
e^{\times}_{v}=\{v\}\cup\pi(v)\ \text{if}\ |\mu(\pi(v))|\ge2,\quad
a\approx b\iff\mu(a)\neq\mu(b)\,\wedge\,b\in\pi\bigl(\mathcal{V}\cap(e_{a,\mathrm{title}}\cup e_{a,\mathrm{id}})\bigr),
\label{eq:xmodal}
\end{equation}
and an encoder $\phi$ embeds entities, hyperedges and segments into dense indices $I_{\mathcal{V}},I_{\mathcal{E}},I_{\sigma}$;
BM25 indexes every surface $\tilde a$, and each index key names the entity, hyperedge or segment that it encodes.

\textbf{Proposition 1.} \textit{In $\mathcal{K}$, any two files $a,b\in\mathcal{D}$, of the same or different modalities, whose records name a common entity $v$ are joined within two hyperedges, namely their record hyperedges containing $v$.}%
\vspace{-2mm}
\begin{proof}
Empirical support is in \S\ref{sec:rq3} (Figure~\ref{fig:rq3}) and the proof in Appendix~\ref{app:proof1}.
\end{proof}

\subsection{C2: Closed-Loop Lifecycle Agent}
\label{sec:c2}

\textbf{Edits.} A handle $h$ names a document across its versions. Let $\mathcal{D}_t$ be the files
present after the $t$-th edit $\Delta_t$ and $\eta_t(h)$ the files of the newest version of $h$; the three edits of \S\ref{sec:taskdef} act as
\begin{equation}
\mathcal{D}_t=
\begin{cases}
\mathcal{D}_{t-1}\cup A, & \Delta_t=\textsc{Add}(A),\\
\mathcal{D}_{t-1}\setminus\eta_{t-1}(h), & \Delta_t=\textsc{Del}(h),\\
\bigl(\mathcal{D}_{t-1}\setminus\eta_{t-1}(h)\bigr)\cup A', & \Delta_t=\textsc{Refresh}(h\!\to\!A'),\ \text{after which}\ \eta_t(h)=A',
\end{cases}
\label{eq:edits}
\end{equation}
where $A$ holds added files, $A'$ a new version, $\mathcal{K}_t=(\mathcal{D}_t,\mathcal{V}_t,\mathcal{E}_t,\pi_t)$ is the artefact after $\Delta_t$
and $\Sigma_t=\{\sigma_{a,j}:a\in\mathcal{D}_t\}$ its segments. \rF{$\textsc{Add}(A)$ runs C1 on $A$ alone; $\textsc{Del}(h)$ calls no model:
it prunes $\eta_{t-1}(h)$ from each $\pi(x)$, so an edit touches only what rests on changed files;
$\textsc{Refresh}(h\!\to\!A')$ is $\textsc{Del}(h)$ then $\textsc{Add}(A')$. Commits restore \emph{provenance soundness}}, with $\mathrm{keys}(I)$ being the keys of $I$:
\begin{gather}
\mathcal{V}_t\cup\mathcal{E}_t=\bigl\{x\in\mathcal{V}_{t-1}\cup\mathcal{E}_{t-1}:\ \pi_{t-1}(x)\not\subseteq\eta_{t-1}(h)\bigr\},\qquad
\pi_t(x)=\pi_{t-1}(x)\setminus\eta_{t-1}(h),
\label{eq:del}\\
\forall x\in\mathcal{V}_t\cup\mathcal{E}_t:\ \ \emptyset\neq\pi_t(x)\subseteq\mathcal{D}_t,\qquad
\mathrm{keys}(I_{\mathcal{V}})\cup\mathrm{keys}(I_{\mathcal{E}})\cup\mathrm{keys}(I_{\sigma})
\subseteq\mathcal{V}_t\cup\mathcal{E}_t\cup\Sigma_t.
\label{eq:invariant}
\end{gather}

\textbf{Proposition 2.} \textit{After any sequence of edits $\Delta_1,\dots,\Delta_t$, all of $\mathcal{K}_t$ rests on present files, i.e., it satisfies Eq.~(\ref{eq:invariant}), so no evidence is stale: no later query draws evidence from a deleted or superseded file.}%
\vspace{-2mm}
\begin{proof}
Empirical support is in \S\ref{sec:rq4} (Figure~\ref{fig:rq4}) and the proof in Appendix~\ref{app:proof2}.
\end{proof}

\subsection{C3: Multi-Granular Retrieval with Intent Routing}
\label{sec:c3}

\textbf{Anchors and intent routing.} Handles $\mathrm{hdl}(q)$ in $q$ resolve via $\eta_t$ to anchor files
$H_q$, and each quoted span $\psi\in\Psi_q$ to the entity $\iota(\psi)$ or its nearest entity in $I_{\mathcal{V}}$ at
cosine ${\ge}\,\theta=0.86$, giving anchor entities $V_q$. Relational, lookup and synthesis cues
$\mathrm{rel}(q),\mathrm{look}(q),\mathrm{syn}(q)\in\{0,1\}$ route $q$ to a tier $g(q)$:
\begin{equation}
H_q=\!\bigcup_{h\in\mathrm{hdl}(q)}\!\eta_t(h),\quad
g(q)=
\begin{cases}
\textit{file}, & H_q\neq\emptyset,\ V_q=\emptyset,\ \neg\mathrm{rel}(q),\\
\textit{subgraph}, & \text{else if}\ H_q\cup V_q\neq\emptyset\ \vee\ (\mathrm{look}\wedge\neg\mathrm{syn})(q),\\
\textit{community}, & H_q\cup V_q=\emptyset\ \wedge\ \mathrm{syn}(q),\\
\Omega(q;\,p_{\mathrm{route}}), & \text{otherwise}.
\end{cases}
\label{eq:route}
\end{equation}

\textbf{Tiers and answering.} The \textit{file} tier reads only the cards $\mathrm{card}(a)$ (record, events, opening
text) of $H_q$. A join walks from the anchors, scoring each other file $b$ with $\mathbf{1}[\cdot]$ the indicator,
$\pi([v])$ the files of $[v]$, $\sqsubseteq$ containment in $\mathrm{head}(b)$ (title, subject) or text $\tilde b$, and $\beta_\psi(b)\in[0,1]$ a BM25 score:
\begin{equation}
\begin{aligned}
s_q(b)=\sum\nolimits_{v\in V_q}10\,\mathbf{1}\bigl[b\in\pi([v])\bigr]
&+\sum\nolimits_{\psi\in\Psi_q}\Bigl(6\,\mathbf{1}\bigl[\psi\sqsubseteq\mathrm{head}(b)\bigr]
+3\,\mathbf{1}\bigl[\psi\sqsubseteq\tilde b\bigr]+2\,\beta_\psi(b)\Bigr)\\
&+\sum\nolimits_{a\in H_q}\sum\nolimits_{u\in\mathcal{V}(a)}\log\bigl(1+|\mathcal{D}_t|/|\pi(u)|\bigr)\,\mathbf{1}\bigl[b\in\pi([u])\bigr],
\end{aligned}
\label{eq:link}
\end{equation}
skipping hubs, entities in over $\max(20,|\mathcal{D}_t|/5)$ files. Twins form works $w$; $W_q$ keeps up to six
within $0.6$ of the best score $\bar s_q(w)$. For synthesis, $\Omega$ reports on the $64$ largest Louvain clusters under
$p_{\mathrm{com}}$. The evidence $\mathcal{X}_q$ holds the anchor cards, those of $W_q$, the four best $I_\sigma$ segments and the top-$k$
hyperedges by rank fusion $\mathrm{rrf}_q$ of lists $L_{q,\mathcal{V}}$ (hyperedges of the $2k$ entities nearest $q$ in
$I_{\mathcal{V}}$) and $L_{q,\mathcal{E}}$ (the $2k$ nearest in $I_{\mathcal{E}}$), or the nearest reports; $\hat y_q=\Omega(q,\mathcal{X}_q;p_{\mathrm{ans}})$ answers $q$ (with $\lambda=4$,~$k=10$):
\begin{equation}
\bar s_q(w)=\sum\nolimits_{b\in w}s_q(b),\qquad
\mathrm{rrf}_q(e)=\sum\nolimits_{L\in\{L_{q,\mathcal{V}},\,L_{q,\mathcal{E}}\}}\mathbf{1}[e\in L]\,\big/\bigl(\lambda+\mathrm{rank}_{L}(e)\bigr).
\label{eq:rrf}
\end{equation}

\textbf{Proposition 3.} \rN{\textit{For every query $q$, each file $a\in H_q$ that $q$ names enters its evidence $\mathcal{X}_q$ as $\mathrm{card}(a)$, whatever $k$, $\lambda$, $\theta$ and the size of $\mathcal{D}_t$, which top-$k$ similarity with any $k$ below that size cannot ensure.}}%
\vspace{-2mm}
\begin{proof}
Empirical support is in \S\ref{sec:rq5} (Figure~\ref{fig:rq5}) and the proof in Appendix~\ref{app:proof3}.
\end{proof}
\setlength{\abovedisplayskip}{7pt plus 2pt minus 5pt}
\setlength{\belowdisplayskip}{7pt plus 2pt minus 5pt}

\pagebreak
\clearpage

\begin{table}[t]
\centering
\caption{\textbf{Main results on OmniKE-Bench.} $\mathcal{T}_1$--$\mathcal{T}_6$: per-task scores over ten personas;
\mbox{\textsc{HC}--\textsc{Gam}}: per-persona means over tasks (descriptive); \textbf{OKE}: mean of persona columns.
\textbf{Bold}: best; \underline{underline}: best baseline; \textit{italic}: ablation equal to the full model. Tags:
input (\tmT\,text, \tmM\,multimodal, \tmO\,omnimodal), structure (\tsN\,none, \tsC\,chunk, \tsG\,graph, \tsH\,hypergraph)
and lifecycle (\taY\,yes, \taN\,no).}
\label{tab:main}
\setlength{\tabcolsep}{2.1pt}\renewcommand{\arraystretch}{1.1}
\scriptsize
\begin{tabular}{l*{6}{c}*{10}{c}>{\columncolor{cOKE}}c}
\toprule
 & \multicolumn{6}{c}{\textit{Tasks}} & \multicolumn{10}{c}{\textit{Personas}} & \multicolumn{1}{c}{} \\
\cmidrule(lr){2-7}\cmidrule(lr){8-17}
\textbf{Method} & $\mathcal{T}_1$ & $\mathcal{T}_2$ & $\mathcal{T}_3$ & $\mathcal{T}_4$ & $\mathcal{T}_5$ & $\mathcal{T}_6$
 & \textsc{HC} & \textsc{Fin} & \textsc{Edu} & \textsc{Law} & \textsc{Psy} & \textsc{Pol} & \textsc{Hum} & \textsc{Bld} & \textsc{Mus} & \textsc{Gam}
 & \textbf{OKE} \\
\midrule
\rowcolor{cNone}\multicolumn{18}{l}{\textit{No knowledge engineering}} \\
\tg{\tmT}{\tsN}{\taN}\quad NaiveGeneration & 0.0 & 28.0 & 0.8 & 1.8 & 37.2 & 59.8 & 17.9 & 13.4 & 22.0 & 10.8 & 22.8 & 9.2 & 10.0 & 16.9 & 18.4 & 18.7 & 16.0 \\
\rowcolor{cChunk}\multicolumn{18}{l}{\textit{Chunk / hierarchical retrieval}} \\
\tg{\tmT}{\tsC}{\taN}\quad StandardRAG & 6.1 & 27.2 & 9.5 & 12.7 & 51.8 & 55.7 & 30.4 & 18.6 & 29.2 & 13.9 & 32.8 & 7.5 & 14.0 & 19.3 & 23.9 & 20.6 & 21.0 \\
\tg{\tmT}{\tsC}{\taN}\quad RAPTOR & 6.4 & 28.5 & 6.2 & 0.9 & 39.4 & 58.0 & 24.6 & 17.5 & 25.0 & 15.0 & 30.1 & 11.0 & 12.0 & 16.2 & 21.7 & 20.2 & 19.3 \\
\rowcolor{cGraph}\multicolumn{18}{l}{\textit{Graph RAG}} \\
\tg{\tmT}{\tsG}{\taN}\quad LightRAG & 13.6 & 31.0 & 9.5 & 2.7 & 50.0 & 58.0 & 30.6 & 24.5 & 29.7 & 22.2 & 33.3 & 6.7 & 18.9 & 20.7 & 24.7 & 27.7 & 23.9 \\
\tg{\tmT}{\tsG}{\taN}\quad GraphRAG & 3.5 & 26.8 & 10.5 & 6.4 & 32.8 & 52.9 & 26.0 & 15.3 & 22.6 & 12.8 & 33.3 & 8.1 & 13.3 & 14.8 & 15.8 & 18.2 & 18.0 \\
\tg{\tmT}{\tsG}{\taN}\quad HippoRAG\,2 & 3.1 & 24.5 & 8.8 & 4.5 & 46.7 & 53.1 & 27.7 & 18.1 & 22.1 & 11.6 & 23.8 & 7.5 & 11.2 & 19.2 & 17.3 & 26.2 & 18.5 \\
\rowcolor{cAgGr}\multicolumn{18}{l}{\textit{Agentic graph RAG}} \\
\tg{\tmT}{\tsG}{\taN}\quad KAG / OpenSPG & 6.7 & 22.8 & 7.0 & 2.7 & 46.7 & 53.4 & 24.9 & 18.4 & 28.7 & 14.2 & 23.7 & 8.3 & 12.8 & 18.3 & 21.5 & 16.1 & 18.7 \\
\tg{\tmT}{\tsG}{\taN}\quad Youtu-GraphRAG & 1.6 & 27.5 & 9.0 & 4.5 & 32.8 & 53.0 & 23.1 & 18.3 & 20.8 & 8.7 & 30.8 & 7.8 & 11.7 & 17.4 & 15.7 & 17.8 & 17.2 \\
\rowcolor{cKGC}\multicolumn{18}{l}{\textit{Knowledge graph construction}} \\
\tg{\tmT}{\tsG}{\taN}\quad iText2KG & 1.3 & 25.2 & 2.2 & 0.0 & 27.7 & 53.8 & 17.4 & 9.5 & 20.2 & 10.8 & 21.7 & 5.0 & 12.4 & 17.8 & 14.5 & 14.8 & 14.4 \\
\tg{\tmT}{\tsG}{\taN}\quad OntoKG & 4.8 & 29.2 & 2.0 & 1.8 & 31.8 & 56.2 & 19.6 & 13.2 & 25.0 & 7.5 & 24.7 & 7.8 & 15.8 & 14.6 & 19.2 & 20.9 & 16.8 \\
\rowcolor{cHyper}\multicolumn{18}{l}{\textit{$n$-ary / hypergraph knowledge engineering}} \\
\tg{\tmT}{\tsH}{\taN}\quad HyperGraphRAG & 4.4 & 29.5 & 7.8 & 5.5 & 40.1 & 53.0 & 25.1 & 15.6 & 27.2 & 15.6 & 30.1 & 8.5 & 12.0 & 17.4 & 23.9 & 17.5 & 19.3 \\
\tg{\tmT}{\tsH}{\taN}\quad Hyper-KGGen & 1.4 & 29.8 & 3.0 & 0.0 & 27.7 & 54.6 & 18.8 & 12.1 & 21.2 & 16.3 & 22.5 & 10.0 & 13.7 & 12.5 & 16.7 & 17.9 & 16.2 \\
\tg{\tmT}{\tsH}{\taN}\quad HyperRAG & 1.1 & 30.2 & 6.5 & 2.7 & 31.8 & 53.1 & 25.4 & 13.3 & 23.7 & 9.2 & 30.4 & 10.8 & 13.3 & 13.8 & 18.0 & 13.0 & 17.1 \\
\rowcolor{cMem}\multicolumn{18}{l}{\textit{Memory agents}} \\
\tg{\tmT}{\tsC}{\taY}\quad MemGPT / Letta & 8.8 & 27.2 & 4.8 & 4.5 & 34.3 & 55.4 & 26.7 & 18.2 & 21.9 & 17.1 & 25.6 & 10.1 & 11.4 & 18.6 & 18.7 & 16.2 & 18.4 \\
\tg{\tmT}{\tsG}{\taY}\quad A-MEM & 7.8 & 25.2 & 10.2 & 10.0 & 49.3 & 58.1 & 29.2 & 20.1 & 28.0 & 12.1 & 31.5 & 8.9 & 13.1 & 20.2 & 26.3 & 20.2 & 21.0 \\
\tg{\tmT}{\tsC}{\taY}\quad Mem0 & 2.4 & 29.5 & 3.5 & 2.7 & 38.0 & 53.0 & 21.9 & 14.5 & 23.6 & 15.1 & 23.3 & 6.2 & 13.3 & 19.6 & 18.7 & 18.2 & 17.4 \\
\tg{\tmT}{\tsG}{\taY}\quad Graphiti (Zep) & 10.9 & 66.0 & 12.0 & 5.5 & 58.8 & 68.5 & 37.0 & 35.6 & 43.1 & 35.0 & 50.1 & 18.3 & 37.2 & 26.6 & 30.5 & 38.8 & 35.2 \\
\tg{\tmT}{\tsC}{\taY}\quad MIRIX & 1.6 & 34.0 & 3.2 & 1.8 & 38.7 & 55.1 & 21.7 & 13.9 & 23.9 & 13.8 & 25.5 & 12.0 & 13.7 & 19.1 & 24.1 & 18.5 & 18.6 \\
\tg{\tmT}{\tsC}{\taY}\quad MemOS & 13.6 & 28.8 & 5.0 & 2.7 & 47.8 & 54.8 & 26.8 & 22.2 & 31.8 & 15.3 & 24.0 & 9.2 & 18.7 & 17.4 & 24.2 & 26.6 & 21.6 \\
\rowcolor{cEvo}\multicolumn{18}{l}{\textit{Self-evolving / closed-loop lifecycle}} \\
\tg{\tmT}{\tsG}{\taY}\quad MemEvolve & 0.0 & 27.0 & 0.5 & 4.5 & 32.5 & 60.6 & 19.9 & 13.0 & 22.1 & 10.0 & 23.7 & 11.7 & 8.3 & 15.9 & 17.6 & 10.0 & 15.2 \\
\tg{\tmT}{\tsG}{\taY}\quad ATOM & 5.1 & 30.5 & 4.2 & 0.0 & 30.3 & 54.8 & 22.4 & 14.8 & 23.7 & 10.8 & 21.7 & 11.7 & 11.6 & 16.7 & 22.3 & 19.7 & 17.5 \\
\tg{\tmT}{\tsC}{\taY}\quad MemPro & 11.4 & \underline{78.2} & \underline{13.5} & 8.2 & \underline{78.8} & \underline{69.5} & 43.5 & \underline{45.2} & \underline{50.7} & 33.4 & \underline{53.6} & \underline{23.6} & 36.5 & \underline{37.4} & \underline{43.4} & \underline{45.5} & \underline{41.3} \\
\rowcolor{cAgKE}\multicolumn{18}{l}{\textit{Agentic knowledge engineering}} \\
\tg{\tmT}{\tsC}{\taY}\quad OpenClaw & 2.9 & 61.5 & 2.8 & 0.9 & 41.2 & 65.2 & 27.1 & 23.0 & 36.9 & 22.5 & 36.2 & 19.9 & 31.8 & 17.2 & 29.0 & 28.3 & 27.2 \\
\tg{\tmT}{\tsG}{\taY}\quad Graphify & 0.9 & 35.8 & 1.2 & 0.0 & 35.8 & 52.4 & 18.3 & 20.2 & 22.8 & 15.9 & 28.3 & 10.2 & 10.5 & 14.2 & 20.7 & 16.4 & 17.8 \\
\tg{\tmT}{\tsG}{\taY}\quad Cognee & 12.8 & 44.0 & 5.0 & 2.7 & 45.3 & 59.7 & 28.2 & 28.2 & 38.1 & 18.1 & 30.9 & 21.5 & 11.7 & 19.4 & 28.9 & 30.4 & 25.6 \\
\tg{\tmT}{\tsC}{\taY}\quad LLM Wiki & 20.0 & 66.2 & 6.5 & 0.0 & 50.7 & 68.1 & 33.9 & 36.5 & 45.8 & 32.5 & 44.5 & 21.6 & \underline{42.4} & 24.2 & 36.5 & 34.8 & 35.3 \\
\tg{\tmT}{\tsC}{\taY}\quad OKF & \underline{27.2} & 68.5 & 8.8 & \underline{17.3} & 54.4 & 67.3 & \underline{45.5} & 44.3 & 46.3 & \underline{38.9} & 49.7 & 16.8 & 37.6 & 27.0 & 38.7 & 40.6 & 38.5 \\
\rowcolor{cOmni}\multicolumn{18}{l}{\textit{Omnimodal / multimodal knowledge engineering}} \\
\tg{\tmO}{\tsC}{\taN}\quad Omni-SimpleMem & 10.3 & 70.2 & 10.2 & 10.9 & 52.2 & 68.8 & 40.6 & 34.0 & 44.7 & 29.9 & 48.3 & 20.2 & 26.6 & 26.0 & 36.4 & 37.9 & 34.5 \\
\tg{\tmM}{\tsG}{\taN}\quad MMGraphRAG & 1.6 & 32.0 & 10.5 & 1.8 & 32.5 & 51.2 & 23.9 & 17.8 & 21.9 & 14.7 & 31.9 & 9.2 & 14.0 & 16.7 & 19.1 & 18.6 & 18.8 \\
\tg{\tmO}{\tsG}{\taN}\quad M$^3$KG-RAG & 17.2 & 60.8 & 1.5 & 0.0 & 32.8 & 60.9 & 30.4 & 32.6 & 39.3 & 30.1 & 41.8 & 15.1 & 29.7 & 17.7 & 25.7 & 25.9 & 28.8 \\
\midrule
\rowcolor{cAbl}\multicolumn{18}{l}{\textit{Ablations}} \\
\tg{\tmT}{\tsC}{\taY}\quad w/o C1 (no hypergraph) & 31.6 & 81.2 & 11.0 & 7.3 & 95.3 & 74.6 & 51.7 & 55.2 & 57.8 & 42.6 & 58.1 & 25.8 & 51.8 & 43.4 & 53.5 & 50.5 & 49.1 \\
\tg{\tmO}{\tsH}{\taN}\quad w/o C2 (no lifecycle) & \textit{51.6} & \textit{82.8} & \textit{15.8} & \textit{20.9} & 49.6 & \textit{76.0} & \textit{60.9} & 52.3 & 50.5 & \textit{46.6} & \textit{65.2} & \textit{28.3} & \textit{57.3} & 35.3 & 53.3 & 54.5 & 50.4 \\
\tg{\tmO}{\tsH}{\taY}\quad w/o C3 (flat retrieval) & 15.7 & 34.8 & 8.0 & 5.5 & 62.8 & 57.5 & 28.2 & 22.3 & 36.7 & 15.6 & 35.2 & 16.7 & 20.7 & 31.6 & 30.0 & 29.5 & 26.7 \\
\midrule
\tg{\tmO}{\tsH}{\taY}\textbf{KnowledgeClaw (ours)} & \textbf{51.6} & \textbf{82.8} & \textbf{15.8} & \textbf{20.9} & \textbf{95.6} & \textbf{76.0} & \textbf{60.9} & \textbf{64.4} & \textbf{59.2} & \textbf{46.6} & \textbf{65.2} & \textbf{28.3} & \textbf{57.3} & \textbf{47.3} & \textbf{64.5} & \textbf{65.7} & \textbf{55.9} \\
\bottomrule
\end{tabular}
\end{table}

\section{Experiments}
\label{sec:exp}
\makeatletter
\def\subsection{\@startsection{subsection}{2}{\z@}{-0.9ex plus -0.3ex minus -.2ex}{0.2ex plus .2ex}{\normalsize\sc\raggedright}}
\makeatother
\setlength{\intextsep}{5pt plus 1pt minus 1pt}\setlength{\abovecaptionskip}{3pt}

We answer \rC{nine} research questions (RQs). \textbf{RQ1--RQ2:} Does KnowledgeClaw outperform the~$30$ baselines, and does each component count?
\textbf{RQ3--RQ5:} Does C1 join files across modalities (Proposition~1), C2 keep answers current (Proposition~2) and C3 supply the evidence each query needs (Proposition~3)?
\textbf{RQ6--RQ7:} Is it efficient and robust to its settings? \rC{\textbf{RQ8:} Is the lead fair to the baselines?} \textbf{\rN{RQ9}:} How does it solve, \rN{from raw media, a query sixteen of the $30$ baselines miss}?

\subsection{Experimental Setup}
\label{sec:setup}

\textbf{Datasets and Evaluation Metrics.} OmniKE-Bench (\S\ref{sec:bench}) holds $1{,}829$ items over
$3{,}960$ real files of ten personas in five modalities. \rF{Scoring calls no model: set F1 for $\mathcal{T}_1$,
$\mathcal{T}_3$ and $\mathcal{T}_4$, accuracy for $\mathcal{T}_2$ and $\mathcal{T}_5$, balanced accuracy for
$\mathcal{T}_6$}, and OKE averages the ten persona scores (detailed in Appendix~\ref{app:bench}).

\textbf{Baselines and Ablations.} Thirty baselines in the ten groups of Table~\ref{tab:main} span every line of
\S\ref{sec:related_work}, from NaiveGeneration, the frozen model with no corpus, through chunk, graph and hypergraph RAG
to memory agents, agentic KE and omnimodal KE; \rC{each is re-implemented from its code or paper with its prompts and
defaults, and all attempt the six tasks.} The ablations w/o C1 (no hypergraph), w/o C2 (no lifecycle) and w/o C3 (flat retrieval) each remove one component of
\S\ref{sec:method} (detailed in Appendix~\ref{app:baseline}).

\textbf{Implementation Details.} \rC{All methods share one frozen omnimodal model $\Omega$
(\textsc{Gemini 2.5 Flash Lite};~\citealp{Gemini25FlashLite}), encoder $\phi$ (\textsc{Gemini Embedding 2}, $3{,}072$ dimensions) and the answer prompt
$p_{\mathrm{ans}}$, so methods differ only in the evidence they supply}; the omnimodal group and KnowledgeClaw read raw
media natively, all other methods a shared media-to-text bridge. \rF{KnowledgeClaw uses $\theta=0.86$, $\lambda=4$, $k=10$ and $64$ communities for every persona and task} (detailed in Appendix~\ref{app:bl-kc}).

\subsection{Main Results (RQ1)}
\label{sec:rq1}

As shown in Table~\ref{tab:main}, KnowledgeClaw ranks first on all six tasks and all ten personas with
$55.9$ OKE, $+14.6$ over the strongest baseline, MemPro. \rF{Figure~\ref{fig:rq1a} scales each task to its best
method, which is KnowledgeClaw on all six, and in Figure~\ref{fig:rq1c} it beats the
strongest baseline of every persona by $+4.7$ to $+21.1$
(two-sided sign test, $p=0.002$).} The five strongest baselines have a lifecycle (MemPro, OKF,
LLM Wiki, Graphiti) or native media reading (Omni-SimpleMem), never both, and none exceeds $41.3$ OKE: the
three properties of OmniKE pay off only when they are all combined.

\subsection{Ablation Study (RQ2)}
\label{sec:rq2}

\begin{figure}[h]
\centering
\includegraphics{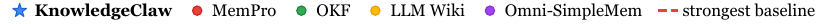}\\[0.3mm]
\begin{subfigure}[t]{0.333\textwidth}\centering\includegraphics{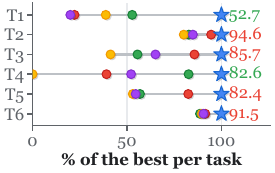}\caption{Share of the best per task}\label{fig:rq1a}\end{subfigure}%
\begin{subfigure}[t]{0.333\textwidth}\centering\includegraphics{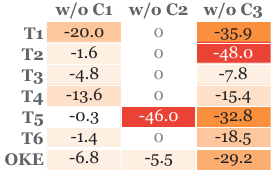}\caption{Ablation minus full model}\label{fig:rq2b}\end{subfigure}%
\begin{subfigure}[t]{0.333\textwidth}\centering\includegraphics{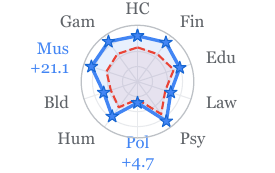}\caption{Per persona}\label{fig:rq1c}\end{subfigure}
\caption{\textbf{RQ1--RQ2.} (a)~\rF{Right: best baseline.} (b)~Table~\ref{tab:main}. (c)~\rF{Task mean; blue: extreme leads.}}
\label{fig:rq1}\label{fig:rq2}
\end{figure}

\leavevmode\rC{Every component is necessary, and each costs most where it acts (Figure~\ref{fig:rq2b})}: without C1, $\mathcal{T}_1$
and $\mathcal{T}_4$ drop most, as only bridged text segments remain; without C2, only $\mathcal{T}_5$
falls, as the superseded version stays; without C3, flat top-$k$ loses on all six tasks. \rC{Without C1 or C2, OKE still stays above every baseline ($49.1$ and $50.4$ against MemPro's $41.3$), so the three components are complementary.}

\subsection{Hypergraph Analysis (RQ3)}
\label{sec:rq3}

\begin{figure}[h]
\centering
\includegraphics{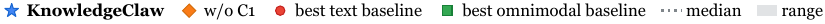}\\[0.3mm]
\begin{subfigure}[t]{0.333\textwidth}\centering\includegraphics{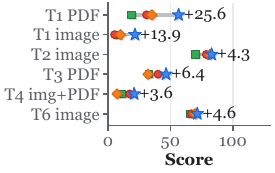}\caption{Per-modality slices}\label{fig:rq3a}\end{subfigure}%
\begin{subfigure}[t]{0.333\textwidth}\centering\includegraphics{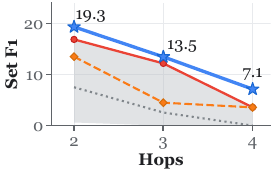}\caption{Join depth}\label{fig:rq3b}\end{subfigure}%
\begin{subfigure}[t]{0.333\textwidth}\centering\includegraphics{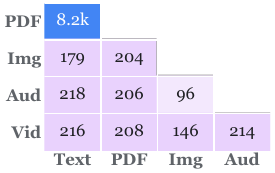}\caption{Cross-modal hyperedges}\label{fig:rq3c}\end{subfigure}
\caption{\textbf{C1 analysis.} (a)~\rC{Label: gain over the best baseline.} (b)~\rP{$\mathcal{T}_3$ and $\mathcal{T}_4$ joins.} (c)~\rC{All personas.}}
\label{fig:rq3}
\end{figure}

\textbf{Typed records lift every modality and hop.} \rC{KnowledgeClaw leads the best text-bridge and
omnimodal baselines on all six modality slices, most on $\mathcal{T}_1$ PDFs and images (Figure~\ref{fig:rq3a}).} \rP{It also leads at every join depth, while the median score across the $30$ baselines falls from $7.5$ to $0.0$ (Figure~\ref{fig:rq3b}).}

\leavevmode\rC{\textbf{Joins cross every modality.} C1 reads all $216$ image, audio and video files of each persona natively
(Table~\ref{tab:app-hg}). Each cell of Figure~\ref{fig:rq3c} counts, over the ten personas, the hyperedges joining
its two modalities: every pair is joined, from $96$ image--audio to $8.2$k text--PDF; $95.4$--$99.2\%$ of files reach another file in two hyperedges (Proposition~1),} and \rP{the
largest join gains fall on shared entities ($+6.0$) and file-name anchors ($+4.9$; Table~\ref{tab:app-rq3}), where similarity on its own has nothing to match.}

\subsection{Lifecycle Analysis (RQ4)}
\label{sec:rq4}

\begin{figure}[h]
\centering
\includegraphics{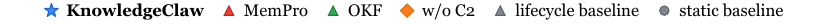}\\[0.3mm]
\begin{subfigure}[t]{0.333\textwidth}\centering\includegraphics{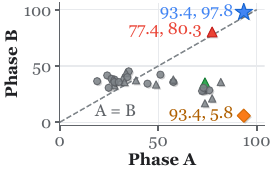}\caption{Before vs.\ after the edit}\label{fig:rq4a}\end{subfigure}%
\begin{subfigure}[t]{0.333\textwidth}\centering\includegraphics{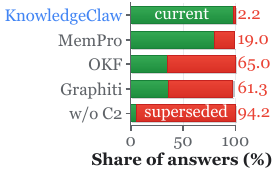}\caption{Phase-B answers}\label{fig:rq4b}\end{subfigure}%
\begin{subfigure}[t]{0.333\textwidth}\centering\includegraphics{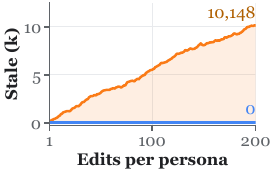}\caption{Edit stream}\label{fig:rq4c}\end{subfigure}
\caption{\textbf{C2 analysis.} (a,\,b)~$\mathcal{T}_5$ accuracy and answers. (c)~\rF{Nodes without a present file, all personas.}}
\label{fig:rq4}
\end{figure}

\rF{Each of the $137$ $\mathcal{T}_5$ items is answered before (phase A) and after (phase B) its edit is committed,
and Figure~\ref{fig:rq4c} streams $2{,}000$ random edits. \textbf{Current after every edit.} KnowledgeClaw scores $93.4$ before and $97.8$ after the edit, while eight of the
nine baselines above $70$ before it fall below $36$ after it (Figure~\ref{fig:rq4a}); it returns the superseded value
for only $2.2\%$, far less than every baseline (Figure~\ref{fig:rq4b}).}

\rF{\textbf{Sound and cheap under a long edit stream.} Over $2{,}000$ edits, Eq.~(\ref{eq:invariant}) never fails and no
node without a present file survives (Proposition~2), while an append-only store accumulates $10{,}148$ such nodes
(Figure~\ref{fig:rq4c}). An edit touches a median $0.33\%$ of the hypergraph and a \textsc{Refresh} takes
$2.6$\,s, against $118$\,s to \textsc{Add} the whole corpus, so each edit costs far less than a full rebuild of the corpus.}

\subsection{Retrieval Analysis (RQ5)}
\label{sec:rq5}

\begin{figure}[h]
\centering
\includegraphics{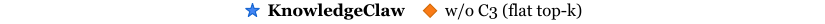}\\[0.3mm]
\begin{subfigure}[t]{0.333\textwidth}\centering\includegraphics{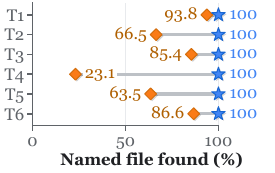}\caption{Named file found}\label{fig:rq5a}\end{subfigure}%
\begin{subfigure}[t]{0.333\textwidth}\centering\includegraphics{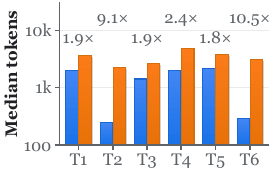}\caption{Evidence length}\label{fig:rq5b}\end{subfigure}%
\begin{subfigure}[t]{0.333\textwidth}\centering\includegraphics{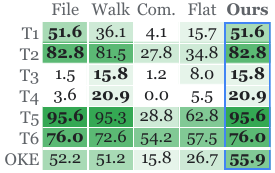}\caption{Forced granularity}\label{fig:rq5c}\end{subfigure}
\caption{\textbf{C3 analysis.} (a)~\rN{Items with all named files in the evidence.} (b)~Per item. (c)~\rC{Bold: best.}}
\label{fig:rq5}
\end{figure}

\rC{A file is \emph{named} when the question carries its handle. Figure~\ref{fig:rq5c} forces one granularity on every
query: anchor cards (File), the walk, community reports (Com.) or flat top-$k$.} \rN{\textbf{Named files arrive, in less evidence.} By Proposition~3, C3 brings every named file's card into the evidence,
where flat top-$k$ misses it on $6$--$77\%$ (Figure~\ref{fig:rq5a})}, with $1.8$--$10.5\times$ shorter evidence for each item
(Figure~\ref{fig:rq5b}).

\rC{\textbf{Routing matches the best fixed granularity.} Each forced granularity fails on some task, file only on the
joins ($1.5$ and $3.6$) and walking on $\mathcal{T}_1$ ($36.1$). The router equals the best fixed one on all six tasks and
reaches $55.9$ OKE, above File ($52.2$), Walk ($51.2$), Flat ($26.7$) and Com.\ ($15.8$) (Figure~\ref{fig:rq5c}).}

\subsection{Efficiency and Robustness (RQ6--RQ7)}
\label{sec:rq6}\label{sec:rq7}

\begin{wrapfigure}[13]{l}{0.55\textwidth}
\centering
\includegraphics{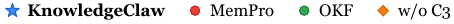}\\[0.3mm]
\begin{subfigure}[t]{0.5\linewidth}\centering\includegraphics{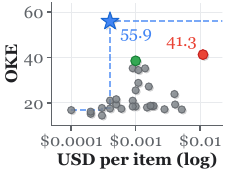}\caption{Query cost}\label{fig:rq6a}\end{subfigure}%
\begin{subfigure}[t]{0.5\linewidth}\centering\includegraphics{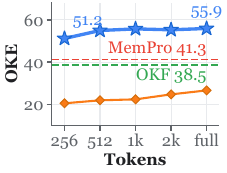}\caption{Evidence budget}\label{fig:rq6b}\end{subfigure}
\caption{\textbf{Efficiency \rF{(RQ6)}.} (a)~Dashed: the best OKE at or below each cost.
(b)~Dashed: MemPro and OKF.}
\label{fig:rq6}
\end{wrapfigure}
\leavevmode\rF{Figure~\ref{fig:rq6a} places each method (grey: other baselines) by its query cost (Table~\ref{tab:cost}) and OKE, and
Figure~\ref{fig:rq6b} cuts the evidence of each item to its first $B$ tokens and answers it again.} \textbf{The best score at a low cost.} KnowledgeClaw spends $\$0.32$ per persona to ingest, $\$0.0004$ per query
and $3$ minutes to build (Table~\ref{tab:cost}), less than MemPro on all three\rF{, by $27.8\times$ per query} (see Figure~\ref{fig:rq6a}).

\leavevmode\rF{\textbf{Short evidence is enough.} Cut to the first $256$
tokens, its evidence still yields $51.2$ OKE, above every baseline even at full context, and $512$ tokens give $55.0$, while flat
top-$k$ stays within $20.6$--$26.7$ at every budget (Figure~\ref{fig:rq6b}); the routed evidence is also shorter than flat top-$k$'s and scores $+7.8$ to $+48.0$ points higher, on every task (Table~\ref{tab:app-rq5} in Appendix~\ref{app:exp-rq5})}.

\begin{wrapfigure}[11]{r}{0.38\textwidth}
\centering
\includegraphics{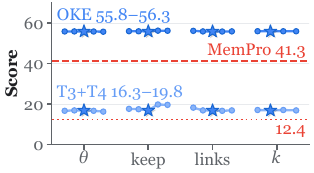}
\caption{\rF{\textbf{Robustness (RQ7).} One hyperparameter varied; star: default.}}
\label{fig:rq7c}\label{fig:rq7}
\end{wrapfigure}
\leavevmode\rF{Figure~\ref{fig:rq7c} varies one hyperparameter at a time around its default (star): the anchor threshold $\theta$, the
keep ratio, the cap on linked works and $k$. \textbf{Stable across settings.} Over its $15$ further settings, OKE stays
within $55.8$--$56.3$ and $\mathcal{T}_3$+$\mathcal{T}_4$ within $16.3$--$19.8$, always above MemPro's $41.3$ and $12.4$
(Figure~\ref{fig:rq7c}): no setting has to be tuned per persona or per task to keep KnowledgeClaw ahead of every baseline. No setting changes $\mathcal{T}_1$, $\mathcal{T}_2$, $\mathcal{T}_5$ or $\mathcal{T}_6$, and OKE moves by at most
$0.4$ from the default (full sweep in Table~\ref{tab:app-rq67}), so none of the defaults is tuned to any task, and none of them needs to be.}

\subsection{\texorpdfstring{\rC{Matched Inputs and Scorers (RQ8)}}{Matched Inputs and Scorers (RQ8)}}
\label{sec:rqfair}

\begin{figure}[h]
\centering
\includegraphics{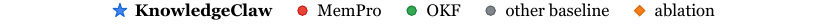}\\[0.3mm]
\begin{subfigure}[t]{0.333\textwidth}\centering\includegraphics{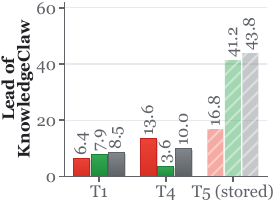}\caption{Lead over +H+R}\label{fig:rqfaira}\end{subfigure}%
\begin{subfigure}[t]{0.333\textwidth}\centering\includegraphics{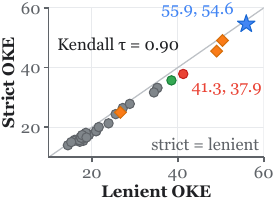}\caption{Strict vs.\ lenient scorer}\label{fig:rqfairb}\end{subfigure}%
\begin{subfigure}[t]{0.333\textwidth}\centering\includegraphics{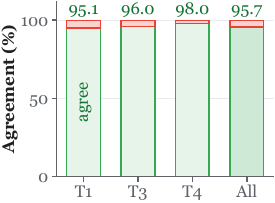}\caption{Matcher vs.\ LLM judge}\label{fig:rqfairc}\end{subfigure}
\caption{\rC{\textbf{Matched inputs and scorers.} (a)~KnowledgeClaw minus each baseline given +H+R; grey: StandardRAG;
light: stored $\mathcal{T}_5$ (not rerun).} \rF{(b)~All $34$ rows of Table~\ref{tab:main}. (c)~Cohen's $\kappa=0.91$.}}
\label{fig:rqfair}
\end{figure}

\begin{wraptable}[16]{l}{0.56\textwidth}
\caption{\rC{\textbf{Matched inputs (RQ8).} \textbf{+H}: the ingested text of each named file is put before
the baseline's own evidence; \textbf{+H+R}: KnowledgeClaw's record card instead. Baseline $\mathcal{T}_5$ (the lifecycle, C2) is not rerun. Bold: best.}}
\label{tab:fair}
\footnotesize
\setlength{\tabcolsep}{3pt}
\begin{tabular*}{\linewidth}{@{\extracolsep{\fill}}l*{7}{c}@{}}
\toprule
\textbf{Method} & $\mathcal{T}_1$ & $\mathcal{T}_2$ & $\mathcal{T}_3$ & $\mathcal{T}_4$ & $\mathcal{T}_5$ & $\mathcal{T}_6$ & \textbf{OKE} \\
\midrule
\rowcolor{cOurs}\textbf{KnowledgeClaw} & \textbf{51.6} & \textbf{82.8} & \textbf{15.8} & \textbf{20.9} & \textbf{95.6} & \textbf{76.0} & \textbf{55.9} \\
\midrule
MemPro & 11.4 & 78.2 & 13.5 & 8.2 & 78.8 & 69.5 & 41.3 \\
\quad + H & 28.3 & 78.8 & 13.2 & 7.3 & 78.8 & 73.8 & 45.9 \\
\quad + H + R & 45.2 & 81.0 & 13.5 & 7.3 & 78.8 & 75.8 & 50.7 \\
\midrule
OKF & 27.2 & 68.5 & 8.8 & 17.3 & 54.4 & 67.3 & 38.5 \\
\quad + H + R & 43.7 & 78.2 & 8.8 & 17.3 & 54.4 & 72.6 & 45.6 \\
\midrule
StandardRAG & 6.1 & 27.2 & 9.5 & 12.7 & 51.8 & 55.7 & 21.0 \\
\quad + H + R & 43.1 & 76.5 & 10.0 & 10.9 & 51.8 & 71.8 & 44.4 \\
\bottomrule
\end{tabular*}
\end{wraptable}
\leavevmode\rC{We give the strongest baselines the inputs KnowledgeClaw uses (Table~\ref{tab:fair}) and rescore every
method (Figure~\ref{fig:rqfair}), to test whether its lead comes from the system or from the evaluation. Builds and
retrieval are replayed, so only the evidence changes, and every answer goes through the same answerer and scorer as
in Table~\ref{tab:main} (the full tables are in Appendix~\ref{app:exp-fair}).}

\leavevmode\rC{\textbf{Ahead with matched inputs.} Given the same handle lookup and KnowledgeClaw's record cards, MemPro rises
from $41.3$ to $50.7$ OKE, still $5.2$ below KnowledgeClaw, which stays first on all six tasks (Table~\ref{tab:fair});
OKF and StandardRAG reach only $45.6$ and $44.4$. The records give the largest lift, on $\mathcal{T}_1$, where they
raise MemPro from $11.4$ to $45.2$ and StandardRAG from $6.1$ to $43.1$: the typed catalogue of C1 is what the
baselines lack, yet none reaches KnowledgeClaw's $51.6$, which also reads the anchor file natively and routes the query. Figure~\ref{fig:rqfaira} plots its lead over each baseline
given +H+R, remaining where lookup and records cannot help, on the joins ($\mathcal{T}_4$ $20.9$ vs.\ at most $17.3$) and the lifecycle ($\mathcal{T}_5$ $95.6$ vs.\ $78.8$), which only the walk and commits supply.}

\leavevmode\rF{\textbf{Independent of the scorer, agreed by a judge.} Both checks reuse the stored predictions of
Table~\ref{tab:main}, so no answer is regenerated. Each point of Figure~\ref{fig:rqfairb} is one row of
Table~\ref{tab:main} at its lenient and strict OKE (diagonal: equal scores). KnowledgeClaw sits top right: under exact
matching it stays first on every task, its margin grows from $14.6$ to $16.7$ OKE, and the two scorers agree closely on the ranking
of all $34$ rows (Kendall $\tau_b=0.90$; per-task scores in Table~\ref{tab:app-score}). In Figure~\ref{fig:rqfairc}, on sampled answers, the
matcher and an LLM judge agree on $95.1$, $96.0$ and $98.0\%$ of the decisions on $\mathcal{T}_1$, $\mathcal{T}_3$ and
$\mathcal{T}_4$ ($95.7\%$ of all decisions; Table~\ref{tab:app-judge}), and the deterministic matcher is the stricter of the two. Of the values that C1 copies into records, $97.8\%$ occur in the source text that it
read (Appendix~\ref{app:exp-fair}).}

\newpage
\begin{table}[t!]
\caption{\rN{\textbf{Case study} on item \texttt{T2\_building\_3261e3912859}. Frames, waveform and transcript come from
the files, answers are saved predictions, and the last two rows follow their saved traces.}}
\label{tab:case}
\footnotesize
\setlength{\tabcolsep}{3pt}\renewcommand{\arraystretch}{1.12}
\begin{tabular}{@{}p{0.9in}p{3.55in}>{\centering\arraybackslash}p{0.85in}@{}}
\toprule
\multicolumn{3}{@{}p{5.5in}@{}}{\textbf{Query} ($\mathcal{T}_2$, building). \textit{Which candidate description is the
authoritative description of the video} \texttt{f75bfdd876eb3d22.mp4}\textit{?} The four candidates open four lectures
of one MIT class by the same speaker and differ only in the topic: (A)~transportation systems, (B)~the grid,
(C)~solar, (D)~a just energy transition. \textbf{Gold:} (D).}\\[1mm]
\multicolumn{3}{@{}c@{}}{\includegraphics{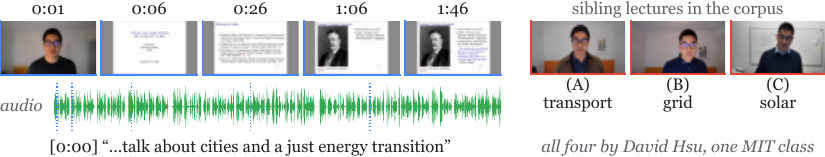}}\\
\midrule
\textbf{Method} & \textbf{How the video reaches the answerer, and what is retrieved} & \textbf{Answer} \\
\midrule
NaiveGeneration & No corpus: the query names only a hashed file name, which carries no topic. & (C) solar \,{\color{red!70!black}\ding{55}} \\
StandardRAG & Text bridge of the video ($30$\,s windows) cut into $1{,}200$-token chunks; top-$5$ by similarity to the query. & (B) grid \,{\color{red!70!black}\ding{55}} \\
LightRAG & Entity-relation graph from the text bridge; query keywords retrieve entities, relations and their chunks. & (A) transport \,{\color{red!70!black}\ding{55}} \\
HippoRAG\,2 & Graph of triples from the text bridge; personalized PageRank from nodes matched to the query. & (A) transport \,{\color{red!70!black}\ding{55}} \\
HyperGraphRAG & $n$-ary hyperedges from the text bridge; query entities and segments retrieve hyperedges and chunks. & (B) grid \,{\color{red!70!black}\ding{55}} \\
Mem0 & Memories distilled from the text bridge; recalled by similarity to the query. & (A) transport \,{\color{red!70!black}\ding{55}} \\
Cognee & Knowledge graph and chunk summaries from the text bridge; the \mbox{top-$5$} triplets nearest to the query. & (A) transport \,{\color{red!70!black}\ding{55}} \\
MMGraphRAG & Text graph over the bridged transcript (raw media only for images); \mbox{top-$5$} nodes by similarity. & (A) transport \,{\color{red!70!black}\ding{55}} \\
w/o C3 & Our hypergraph with flat top-$k$: ten hyperedges, five of them identifiers of other videos, and six passages,
none of them from the named video ($10.5$k characters). & (B) grid \,{\color{red!70!black}\ding{55}} \\
\midrule
\rowcolor{cOurs}\textbf{KnowledgeClaw} & \raisebox{-0.5\height}{\includegraphics{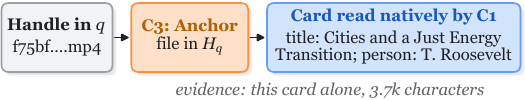}} & \textbf{(D) just transition}\,{\color{green!45!black}\ding{51}} \\
\bottomrule
\end{tabular}
\end{table}

\subsection{\texorpdfstring{Case Study (\rN{RQ9})}{Case Study (RQ9)}}
\label{sec:rq8}

Table~\ref{tab:case} traces an item whose video is one of four sibling lectures with the same speaker, class and
near-identical openings, which similarity to a query naming only a file cannot separate: sixteen of the $30$ baselines
choose a sibling's description. KnowledgeClaw resolves the handle to the anchor file and answers from its card alone,
a record that C1 read from the raw video, so the topic, the one field in which the four candidates differ, is in the
evidence, and it chooses the gold description (D) from $2.9\times$ less evidence than w/o C3 ($3.7$k vs.\ $10.5$k characters), all of it from the named video. \rN{This case isolates lookup (C3) and native reading (C1); Tables~\ref{tab:main} and~\ref{tab:fair} measure the overall gain.}

\section{Conclusion}
\label{sec:conclusion}

We formalized Omnimodal Knowledge Engineering (OmniKE), which asks a live knowledge base for omnimodality, a closed-loop
lifecycle and multi-granular application at once, and released OmniKE-Bench, six model-free tasks over the real files
of ten personas in five modalities. Our KnowledgeClaw reads every file natively into an omnimodal typed
hypergraph (C1), keeps it sound under edits with a closed-loop lifecycle agent (C2), and routes each query to the granularity it
needs (C3). Against $30$ baselines it ranks first on all six tasks and all ten personas and stays current after every edit.
\rC{Given its handle lookup and records, the strongest baseline still trails it by $5.2$ OKE.}

\clearpage

\makeatletter
\def\subsection{\@startsection{subsection}{2}{\z@}{-1.8ex plus -0.5ex minus -.2ex}{0.8ex plus .2ex}{\normalsize\sc\raggedright}}
\makeatother
\setlength{\intextsep}{12pt plus 2pt minus 2pt}\setlength{\abovecaptionskip}{10pt}
\subsection*{AI use statement}

Large language models enter this work only as studied components: the frozen backbone shared by all evaluated systems
(\S\ref{sec:setup}) and the stem paraphraser of OmniKE-Bench (\S\ref{sec:bench}), which never writes a gold answer. Beyond
these roles, they were used only to polish language and to assist with coding. The authors conceived the
research, designed and ran every experiment, drew every conclusion, and reviewed, tested and verified all AI-assisted text and code,
for which they take full responsibility.

\subsection*{Ethics statement}

OmniKE-Bench is built only from public sources that permit redistribution (Table~\ref{tab:sources}): public-domain
government records, open-access papers, and open museum, media and course collections. Every released file keeps its
original licence, including the non-commercial terms of some sources. The benchmark involves no human subjects and
collects no private data: its ten personas are assembled from public material, speech and video come only from public
proceedings and openly licensed recordings, and every file is stripped of metadata, original names and source URLs. No
gold answer is produced by a model, so it adds no model-generated misinformation. OmniKE-Bench is intended for research
use; systems that, like KnowledgeClaw, index personal files should run only on data their users own or are entitled to process,
and every use must honour the licence of each source file.

\subsection*{Reproducibility statement}

All material needed to reproduce our results is given in the appendices. Appendix~\ref{app:prompt} lists every prompt of
KnowledgeClaw verbatim with its inputs and output limits; Appendix~\ref{app:proofs} proves the three propositions of
\S\ref{sec:method}; Appendix~\ref{app:algorithm} gives the full algorithm of construction, lifecycle commits and retrieval.
Appendix~\ref{app:bench} specifies the six tasks, the seven-stage construction of OmniKE-Bench from its licensed sources
and the model-free metrics; a SHA-256 manifest seals the release, and one command rebuilds every item from the raw files.
Appendix~\ref{app:baseline} fixes the shared deployment protocol (one frozen model and encoder, temperature $0$ by default,
cached calls) and the configuration of every baseline, KnowledgeClaw and its ablations. Appendix~\ref{app:exp} reports
the evaluation protocol, cost, and the setting and full results of every analysis, all recomputed from saved predictions and traces.



\bibliography{iclr2027_conference}
\bibliographystyle{iclr2027_conference}

\newpage
\appendix
\section*{Appendix} 
\captionsetup[figure]{position=bottom,skip=5pt}\captionsetup[table]{position=top,skip=4pt}
\setlength{\intextsep}{6pt plus 1pt minus 1pt}\setlength{\textfloatsep}{6pt plus 1pt minus 1pt}
\renewcommand{\arraystretch}{1.0}\setlength{\tabcolsep}{1.5pt}\renewcommand{\tabularxcolumn}[1]{m{#1}}

\section{Prompts Used in KnowledgeClaw}\label{app:prompt}

This section gives, verbatim, every prompt that KnowledgeClaw sends to the frozen model $\Omega$
(\textsc{Gemini~2.5~Flash~Lite}); Table~\ref{tab:prompts} places each in \S\ref{sec:method}. In the boxes,
\slot{\textnormal{\dots}} marks a slot filled at run time and \media{\dots} a raw media part. All calls run at
temperature $0$, in JSON mode for JSON replies. C1 owns four prompts and C3 three; \rF{C2 adds none: \textsc{Del} calls no model, and
\textsc{Add} and \textsc{Refresh} run C1's prompts on the added files, so edits and construction share
one extraction path.}

\begin{table}[H]
\caption{The prompts of KnowledgeClaw; \emph{Max} is the output-token limit of the call.}\label{tab:prompts}
\small
\begin{tabular*}{\linewidth}{@{\extracolsep{\fill}}lllllr@{}}
\toprule
Prompt & Used in & Input & Output & Calls & Max \\
\midrule
$p_{\textit{image}}$ & C1, Eq.~(\ref{eq:read}) & image, ${\le}1{,}536$\,px & description, visible text & per image & 1,000 \\
$p_{\textit{audio}}$ & C1, Eq.~(\ref{eq:read}) & audio, $16$\,kHz mono & summary, timed transcript & per recording & 3,000 \\
$p_{\textit{video}}$ & C1, Eq.~(\ref{eq:read}) & ${\le}6$ keyframes, audio & description, transcript & per video & 3,000 \\
$p_{\mathrm{rec}}$ & C1, Eq.~(\ref{eq:record}) & first $9{,}000$ chars of $\tilde a$ & JSON record & per file & 2,500 \\
$p_{\mathrm{route}}$ & C3, Eq.~(\ref{eq:route}) & query $q$ & one label & if no rule routes $q$ & 5 \\
$p_{\mathrm{com}}$ & C3, communities & entities, propositions & report, ${\le}200$ words & per community & 400 \\
$p_{\mathrm{ans}}$ & C3, answering & evidence $\mathcal{X}_q$, question & JSON answer & per query & 600 \\
\bottomrule
\end{tabular*}
\end{table}

\subsection{Reading prompts $p_{\mu(a)}$}\label{app:prompt-read}

\leavevmode\rC{Eq.~(\ref{eq:read}) reads each image, audio and video file once, natively, under the prompt of its modality
(Figure~\ref{fig:prompt-read}).} Since the reading $\tilde a$ is what the record prompt catalogues, each prompt
asks for what tells a file apart from its look-alikes rather than for a caption: the object type, depicted
content, technique and every legible word of an image, with an inventory for a floor plan; the speakers and a
time-marked transcript of a recording; and both, with on-screen text, for a video, sent as up to six
time-labelled keyframes and its audio track. Each reading is then split into segments of about
$1{,}000$ tokens that keep their file's modality and together form the passage index $I_\sigma$ that C3 searches.

\begin{figure}[H]
\begin{promptbox}{pmCa}{pmCaF}{Image reading prompt $p_{\textit{image}}$}
You are indexing one image for a personal knowledge base, so that it can later be told apart from similar images and matched to its catalogue record. Write plain prose (at most 250 words), then one line "Visible text:" with every legible word verbatim (or "none").
\par Cover, when present: what kind of object the image is (photograph, photomicrograph, electron micrograph, engraving, drawing, painting, portrait, map, chart, floor plan, diagram, screenshot, poster, document scan, object photo, ...); what it depicts -{}- people (how many, age group, sex, activity, clothing), organisms, anatomy, objects, instruments, buildings, places, setting and apparent period; technique (stain, magnification, colour or black-and-white).
\par For a floor plan or technical diagram, add a line "Inventory:" that counts what is drawn, by type: rooms (each room type with its count), walls, windows, doors, stairs, columns, railings, and every fixed fitting or furniture item (sinks, toilets, showers, bathtubs, cabinets, closets, stoves, appliances, sauna benches, ...) with its count.
\par Do not guess names, places or dates that are not visible.
\pgap\media{image}
\end{promptbox}\par\vspace{3pt}
\begin{promptbox}{pmCa}{pmCaF}{Audio reading prompt $p_{\textit{audio}}$}
You are indexing one audio recording for a personal knowledge base. First give a 1-3 sentence summary: what the recording is (lecture, interview, talk, song, sound effect, ...), who speaks (names if said), and the topic or musical character (genre, instruments, mood, tempo). Then write "Transcript:" and transcribe all speech verbatim, with a time mark such as [0:30] about every 30 seconds. If there is no speech, write "Transcript: none".
\pgap\media{audio}
\end{promptbox}\par\vspace{3pt}
\begin{promptbox}{pmCa}{pmCaF}{Video reading prompt $p_{\textit{video}}$}
You are indexing one video for a personal knowledge base. You are given keyframes (with their times) and the audio track. First describe what the video shows in 2-4 sentences: kind of video (lecture, interview, demonstration, gameplay, animation, ...), people, setting, and all on-screen text verbatim. Then write "Transcript:" and transcribe all speech verbatim, with a time mark such as [0:30] about every 30 seconds (or "Transcript: none").
\pgap Frame at 0:10: \media{keyframe}\quad Frame at 0:30: \media{keyframe}\quad \dots\quad Frame at 1:50: \media{keyframe}\quad Audio track: \media{audio}
\end{promptbox}
\caption{Reading prompts $p_{\mu(a)}$ of Eq.~(\ref{eq:read}) for image, audio and video files; each
prompt is sent together with the raw media part(s) shown last; the reply is the surface $\tilde a$ read by the record prompt.}\label{fig:prompt-read}
\end{figure}
\vskip-\intextsep\vskip-\dp\strutbox 
\pagebreak
\subsection{Record prompt $p_{\mathrm{rec}}$}\label{app:prompt-record}

Eq.~(\ref{eq:record}) writes each file's record with one call of the prompt in
Figure~\ref{fig:prompt-record} on the first $9{,}000$ characters of $\tilde a$. \rN{The files of all ten personas share one
schema $\Gamma$ of $13$ general cataloguing fields and dated events. As a join matches forms, not topics,
\emph{subject} asks for the preferred form of standard vocabularies (arXiv codes, MeSH descriptors and check tags),
and \emph{element} for canonical class names.}

\begin{figure}[H]
\begin{promptbox}{pmCa}{pmCaF}{Record prompt $p_{\mathrm{rec}}$}
You are cataloguing one file of a personal knowledge base. From the file content below, write its catalogue record. Copy names, identifiers and dates exactly as they appear; never invent a value that the content does not support.
\pgap Return JSON only:
\pj{0pt}{\{"title": "<title as printed, or a short descriptive title>",}
\pj{0.6em}{"kind": "<what the file is: preprint, journal article, report, bill, thesis, abstract, photograph, photomicrograph, drawing, floor plan, lecture recording, interview, song, ...>",}
\pj{0.6em}{"summary": "<one sentence>",}
\pj{0.6em}{"fields": \{}
\pj{1.2em}{"author": ["<each author's full name, in order>"],}
\pj{1.2em}{"affiliation": ["<each distinct affiliation, the full string as printed>"],}
\pj{1.2em}{"subject": ["<the controlled-vocabulary subject terms a library catalogue would assign to this file, each in the vocabulary's exact preferred form: arXiv category codes (e.g. cs.CL, eess.AS, q-bio.NC) for arXiv preprints; 5-15 MeSH descriptors for biomedical articles, written as MeSH writes them (plural nouns, inverted forms with commas), covering the population studied, the conditions, organisms, interventions, settings and methods, plus check tags (Humans, Male, Female, age groups) when the study population supports them; policy areas for government reports; subject headings otherwise>"],}
\pj{1.2em}{"keyword": ["<author keywords or key topics>"],}
\pj{1.2em}{"identifier": ["<DOI, arXiv id, report or bill number, PMID, ...>"],}
\pj{1.2em}{"venue": ["<journal, conference, publisher or series>"],}
\pj{1.2em}{"organization": ["<other organisations named>"],}
\pj{1.2em}{"person": ["<other people named: speakers, editors, people shown>"],}
\pj{1.2em}{"place": ["<places>"],}
\pj{1.2em}{"depicts": ["<for media: what is shown or heard: people, organisms, objects, activities>"],}
\pj{1.2em}{"object\_type": ["<for media: the kind of object, e.g. Photographs, Portraits, Engravings, Maps>"],}
\pj{1.2em}{"element": ["<for plans, diagrams and building photos: every room type and building element shown, each as a canonical singular class name (e.g. Bedroom, Kitchen, Door, Window, Wall, Stairs, Balcony, Roof)>"],}
\pj{1.2em}{"language": ["<language of the content>"]}
\pj{0.6em}{\},}
\pj{0.6em}{"events": [\{"event": "<submitted | received | revised | accepted | published | published online | published print | created | recorded | issued | ...>", "date": "<YYYY-MM-DD, YYYY-MM or YYYY>"\}]\}}
\pgap Rules: leave a field as [] when the content does not state it. An arXiv stamp such as "arXiv:2205.01234v2 [cs.CL] 24 May 2022" gives the identifier, the primary subject category and the submission date of that version. Journal front matter gives received / accepted / published dates.
\pgap FILE:
\par\slot{content}
\end{promptbox}
\caption{Record prompt $p_{\mathrm{rec}}$ of Eq.~(\ref{eq:record}); \slot{content} is the opening of $\tilde a$.}\label{fig:prompt-record}
\end{figure}

\paragraph{From record to hyperedges.} \leavevmode\rF{Before any write, fields outside $\Gamma$, placeholder
values (``unknown'', ``n/a'', ``none'', ``null''), values of $400$ or more characters and duplicates are dropped, and
an event is kept only if its date parses.} The title and each non-empty field become one \emph{record} hyperedge
holding the file and the field's values, and each event an \emph{event} hyperedge dated YYYY-MM-DD, YYYY-MM or YYYY.
Every value enters through $\iota$, so a value shared by two records is one entity with two source files;
Eq.~(\ref{eq:xmodal}) then adds a cross-modal hyperedge for each entity shared across modalities.

\subsection{Routing prompt $p_{\mathrm{route}}$}\label{app:prompt-route}

\leavevmode\rF{Eq.~(\ref{eq:route}) routes by rule a query that names a file or entity, is short, or has a lookup or
synthesis cue; others go to Figure~\ref{fig:prompt-route}, whose one-word reply picks the subgraph or
community tier or both (\emph{hybrid}).}

\begin{figure}[H]
\begin{promptbox}{pmCc}{pmCcF}{Routing prompt $p_{\mathrm{route}}$}
Classify the query into ONE retrieval strategy over a personal knowledge hypergraph.
\par subgraph: a precise fact, definition, who/when/where/which. community: open-ended synthesis or a thematic overview. hybrid: ambiguous, or needs both.
\par Reply with exactly one word: subgraph, community or hybrid.
\par Query: \slot{query}
\end{promptbox}
\caption{Routing prompt $p_{\mathrm{route}}$, the last case of Eq.~(\ref{eq:route}).}\label{fig:prompt-route}
\end{figure}

\vskip-\intextsep\vskip-\dp\strutbox 
\pagebreak
\subsection{Community report prompt $p_{\mathrm{com}}$}\label{app:prompt-community}

The community tier summarises each of the $64$ largest Louvain communities with the prompt of
Figure~\ref{fig:prompt-community}. Its slots hold the names of up to $15$ member entities, up to $12$
propositions, each the text of a record hyperedge on a member, and the modalities those hyperedges were read
from. Louvain runs with a fixed seed on the entity graph whose edge weight counts the record hyperedges two
entities share. The reports, at most $200$ words in four fixed parts, form the dense index that the community tier
searches with $\phi(q)$; any edit invalidates them, and they are rebuilt on next use.

\begin{figure}[H]
\begin{promptbox}{pmCc}{pmCcF}{Community report prompt $p_{\mathrm{com}}$}
You are summarising one community of a personal knowledge hypergraph.
\par Entities: \slot{entities}
\par Propositions:
\par\slot{propositions}
\par Modalities: \slot{modalities}
\par Write a report of at most 200 words, plain prose: (1) THEME, one sentence; (2) KEY ENTITIES, the 3-5 most central and their role; (3) SUMMARY of the facts and relationships; (4) CROSS-MODAL NOTES if several modalities occur. No generic boilerplate.
\end{promptbox}
\caption{Community report prompt $p_{\mathrm{com}}$ of the community tier (\S\ref{sec:c3}).}\label{fig:prompt-community}
\end{figure}

\subsection{Answer prompt $p_{\mathrm{ans}}$}\label{app:prompt-answer}

\leavevmode\rC{Every method of Table~\ref{tab:main} that reads the corpus answers through the same prompt}
(Figure~\ref{fig:prompt-answer}): a system message, then the evidence, the question and the JSON output
contract of the task's answer type, so that methods differ only in the evidence they supply. KnowledgeClaw
fills the context with $\mathcal{X}_q$, anchor cards first. On phase B of a $\mathcal{T}_5$ item it retrieves
again from the edited $\mathcal{K}$, and the question carries the update notice shown in the box.
\rF{The contract follows the item's answer type: a choice index for the four-way items of $\mathcal{T}_2$ and
$\mathcal{T}_5$, the entity names for $\mathcal{T}_1$, the asked value for the joins of $\mathcal{T}_3$ and
$\mathcal{T}_4$, and the covered candidates for $\mathcal{T}_6$; the reply is parsed into the prediction that
\S\ref{sec:metrics} scores.} For the video item of Table~\ref{tab:case}, KnowledgeClaw sends only
the card of the named video, $3{,}659$ characters of catalogue record and timestamped transcript, and $\Omega$
chooses the gold option (D) from it alone.

\begin{figure}[H]
\begin{promptbox}{pmCc}{pmCcF}{Answer prompt $p_{\mathrm{ans}}$}
{\color{pmCcF}\bfseries System:}\enspace You may use the CONTEXT below (retrieved automatically over the released corpus; it may be irrelevant, incomplete, or -{}- in the ablation condition -{}- deliberately missing one modality). Do not assume anything not stated in the context or the question.
\pgap{\color{pmCcF}\bfseries User:}\enspace CONTEXT:
\par\slot{evidence}
\pgap QUESTION: \slot{question}\qquad{\color{black!55}\itshape (phase B of $\mathcal{T}_5$:)} QUESTION: (An update to the underlying record may have been filed since any earlier question.) \slot{question}
\pgap{\color{pmCcF}\bfseries $\mathcal{T}_2$, $\mathcal{T}_5$:}\enspace Options:
\par [1] \slot{option 1}
\par [2] \slot{option 2}
\par [3] \slot{option 3}
\par [4] \slot{option 4}
\pgap Respond with JSON only: \{"choice": <1-4>\}
\pgap{\color{pmCcF}\bfseries $\mathcal{T}_1$:}\enspace Respond with JSON only: \{"entities": ["<entity name>", "<entity name>", ...]\}. Each element must be ONLY the entity's own name/surface form, exactly as it appears -{}- for example a person's name, an organisation, a subject term, or a physical part or feature. Do NOT list the entity TYPES you were asked about (e.g. do not answer with "author" or "subject" themselves), do NOT wrap entities in objects with their own type/modality fields, and do NOT answer "unknown" -{}- if you cannot find any, respond with an empty list.
\pgap{\color{pmCcF}\bfseries $\mathcal{T}_3$, $\mathcal{T}_4$:}\enspace Answer the question directly. Respond with JSON only: \{"answer": ["<value>"]\} -{}- a list containing the specific value(s) the question asks for (usually exactly one, e.g. a single date, name or identifier), not a list of unrelated entities found in the context.
\pgap{\color{pmCcF}\bfseries $\mathcal{T}_6$:}\enspace Candidates:
\par - \slot{candidate 1}
\par - \dots
\pgap Respond with JSON only: \{"covered": ["<candidate>", ...]\}. Each element must be one of the candidate strings above, copied exactly, as a plain string (not wrapped in an object).
\end{promptbox}
\caption{Answer prompt $p_{\mathrm{ans}}$ of \S\ref{sec:c3}, shared by every method that reads the corpus:
a system message and a user message that ends with the output contract of the item's task ($600$ output tokens).}\label{fig:prompt-answer}
\end{figure}
\vskip-\intextsep\vskip-\dp\strutbox 
\pagebreak
\clearpage

\section{Theoretical Proof}
\label{app:proofs}

The notation is that of \S\ref{sec:method}; $\mathcal{K}_t=(\mathcal{D}_t,\mathcal{V}_t,\mathcal{E}_t,\pi_t)$
is the artefact after the $t$-th edit, $\mathcal{K}_0$ is empty, and an \emph{element} is an entity or hyperedge $x\in\mathcal{V}_t\cup\mathcal{E}_t$. A \emph{path} of length $m$ between
nodes $x_0$ and $x_m$ is an alternating sequence $x_0\in e_1\ni x_1\in e_2\ni\dots\in e_m\ni x_m$ of nodes and
hyperedges of $\mathcal{E}_t$, and $d_t(x_0,x_m)$ is the length of a shortest one ($\infty$ if none).
Three facts of the construction are used throughout: (F1) node identity is $\iota$, a deterministic function
of a value, so $\mathcal{V}_t$ holds at most one node per key; (F2) a record write changes $\mathcal{K}$ only as
in Eq.~(\ref{eq:record}), creating an element with provenance $\{a\}$ or adding $a$ to an existing provenance;
(F3) every edit is one of the three cases of Eq.~(\ref{eq:edits}), and edits are committed one after another, so each
step of a proof acts on one fixed, committed artefact $\mathcal{K}_{t-1}$.

\subsection{Proof of Proposition 1}
\label{app:proof1}

\textbf{Proposition 1.} \textit{In $\mathcal{K}$, any two files $a,b\in\mathcal{D}$, of the same or different modalities, whose records name a common entity $v$ are joined within two hyperedges, namely their record hyperedges containing $v$.}%
\vspace{-2mm}
\begin{proof}
We first show that merging is order-free. \emph{Lemma.} After the records of a set $A$ of files are
merged, in any order, $\mathcal{V}\supseteq\{\iota(z):z\in Z_{a,\gamma},\ a\in A,\ \gamma\in\Gamma\}$ and, for
each such node $v$, $\pi(v)\supseteq\{a\in A:\exists\gamma,\ v\in e_{a,\gamma}\}$. Indeed, by (F1) and (F2)
the write of $a$ reaches the unique node of key $\iota(z)$: the first write creates it and every later one
unions its file into $\pi(v)$. Set union is commutative and associative, so the result does not depend on
the order of the writes, nor on whether they come from one \textsc{Add} or from several, so building $\mathcal{K}$
at once or by later \textsc{Add}s gives the same $\mathcal{K}$.

Now let $a,b\in\mathcal{D}_t$ hold values $z\in Z_{a,\gamma}$ and $z'\in Z_{b,\gamma'}$ with
$\iota(z)=\iota(z')=v$. By the lemma the node $v$ exists and the record hyperedges
$e_{a,\gamma}\ni a,v$ and $e_{b,\gamma'}\ni v,b$ are both in $\mathcal{E}_t$, whence the path
\begin{equation}
a\ \in\ e_{a,\gamma}\ \ni\ v\ \in\ e_{b,\gamma'}\ \ni\ b,\qquad d_t(a,b)\le2 .
\label{eq:p1path}
\end{equation}
If moreover $\mu(a)\neq\mu(b)$, then $a,b\in\pi(v)$ gives $|\mu(\pi(v))|\ge2$, so the cross-modal hyperedge
$e^{\times}_v=\{v\}\cup\pi(v)$ of Eq.~(\ref{eq:xmodal}) contains both files and $d_t(a,b)=1$. No step depends
on the modalities: Eq.~(\ref{eq:read}) maps every modality to a text surface, one record call maps every
surface onto the same schema $\Gamma$, and $\iota$ acts on values alone, so the bound holds for all
$|\mathcal{M}|^2=25$ modality pairs alike. For person names the walk works in the quotient of
$\mathcal{K}_t$ by the alias classes $[v]$; replacing $v$ by $[v]$ in Eq.~(\ref{eq:p1path}) gives the same
bound for two files that write ``M.~Langtry'' and ``Max Langtry'' alike.

\emph{Chains.} If $a_0,\dots,a_m\in\mathcal{D}_t$ and each consecutive pair names a common entity, then
concatenating the $m$ paths of Eq.~(\ref{eq:p1path}) gives $d_t(a_0,a_m)\le2m$: a join over $m$ records is
a walk of at most $2m$ steps over one structure, whatever the modalities along the way, as the joins of
$\mathcal{T}_3$ and $\mathcal{T}_4$ require.

\emph{Stability.} The join holds exactly while both records are present. If an edit removes a file
$c\notin\{a,b\}$, Eq.~(\ref{eq:del}) only shrinks $\pi(v)\supseteq\{a,b\}$ to a non-empty set, so $v$ is
kept, while $e_{a,\gamma}$ and $e_{b,\gamma'}$, whose provenance is $\{a\}$ and $\{b\}$, are untouched; if it
removes $a$, the hyperedge $e_{a,\gamma}$ leaves with it, and so does the path. A store of segments, by
contrast, has no node for $v$: $a$ and $b$ meet only if one similarity search happens to return both, which
no structure of that segment store ensures.
\end{proof}
Empirically, $95.4$--$99.2\%$ of each persona's files lie within two hyperedges of another file (Table~\ref{tab:app-hg}).

\subsection{Proof of Proposition 2}
\label{app:proof2}

\textbf{Proposition 2.} \textit{After any sequence of edits $\Delta_1,\dots,\Delta_t$, all of $\mathcal{K}_t$ rests on present files, i.e., it satisfies Eq.~(\ref{eq:invariant}), so no evidence is stale: no later query draws evidence from a deleted or superseded file.}%
\vspace{-2mm}
\begin{proof}
We first show by induction on $t$ that Eq.~(\ref{eq:invariant}) holds after every edit. For $t=0$,
$\mathcal{K}_0$ is empty and the claim is vacuous; assume it holds for $\mathcal{K}_{t-1}$. By (F3) there are
exactly three cases to check.

\emph{Case} $\Delta_t=\textsc{Add}(A)$. By (F2) a write of $a\in A$ gives each element it creates the provenance
$\{a\}$ and adds $a$ to the provenance of each element that exists; a cross-modal hyperedge takes
$\pi(e^{\times}_v)=\pi(v)$. Every provenance thus stays non-empty and within
$\mathcal{D}_{t-1}\cup A=\mathcal{D}_t$, and index keys are added only for the elements just written and for the
segments of $A$, which lie in $\Sigma_t$; nothing is removed.

\emph{Case} $\Delta_t=\textsc{Del}(h)$. Let $R=\eta_{t-1}(h)$, so that $\mathcal{D}_t=\mathcal{D}_{t-1}\setminus R$.
By Eq.~(\ref{eq:del}) with $\eta_{t-1}(h)=R$, the edit removes, with its index keys, every element whose
provenance lies within $R$, and deletes $R$ from every other provenance, which therefore stays non-empty:
\begin{equation}
\emptyset\neq\pi_t(x)=\pi_{t-1}(x)\setminus R\subseteq\mathcal{D}_{t-1}\setminus R=\mathcal{D}_t
\qquad\text{for every surviving element }x .
\label{eq:p2del}
\end{equation}
The segments of $R$ leave $I_\sigma$ with their files. Dropping orphans only removes nodes, and recomputing
$e^{\times}_v$ from the pruned $\pi(v)$ keeps $\pi(e^{\times}_v)=\pi(v)\subseteq\mathcal{D}_t$. Every element $x$
with $\pi_{t-1}(x)\cap R=\emptyset$ keeps its provenance unchanged, so the edit is local: it touches only the
elements that $R$ supports.

\emph{Case} $\Delta_t=\textsc{Refresh}(h\to A')$. It is \textsc{Del}$(h)$ followed by \textsc{Add}$(A')$ and
preserves the invariant because both do; re-pointing $\eta_t(h)=A'\subseteq\mathcal{D}_t$ changes no
provenance, and a later query naming $h$ resolves to $A'$, so it reads the new values and never the old
ones. This completes the induction.

Now let $c\notin\mathcal{D}_t$ be a superseded or deleted file. The evidence $\mathcal{X}_q$ of
\S\ref{sec:c3} is assembled from (i) the cards of files in $H_q$ and of linked files, all in
$\mathcal{D}_t$ because $\eta_t$ maps every handle into $\mathcal{D}_t$ and the walk scores only present
files; (ii) hyperedges of $\mathcal{E}_t$ and entities of $\mathcal{V}_t$, whose provenance lies in
$\mathcal{D}_t$ by Eq.~(\ref{eq:invariant}); (iii) passages keyed in $I_\sigma$, which belong to files of
$\mathcal{D}_t$ by Eq.~(\ref{eq:invariant}); and (iv) the keyword index and the community summaries, which
every edit invalidates and which are rebuilt from $\mathcal{K}_t$. The record and event hyperedges that
stated the values of $c$ had provenance $\{c\}$ and were removed with it, and a value of $c$ that survives
in $\mathcal{K}_t$ is, up to $\iota$, stated by a present file. Hence $c$ contributes to no part of
$\mathcal{X}_q$, for any query $q$ and any sequence $\Delta_1,\dots,\Delta_t$, so no evidence is stale.
\end{proof}
Empirically, Eq.~(\ref{eq:invariant}) held after each of the $2{,}000$ edits of the stress test in \S\ref{sec:rq4}.

\subsection{Proof of Proposition 3}
\label{app:proof3}

\textbf{Proposition 3.} \textit{For every query $q$, each file $a\in H_q$ that $q$ names enters its evidence $\mathcal{X}_q$ as $\mathrm{card}(a)$, whatever $k$, $\lambda$, $\theta$ and the size of $\mathcal{D}_t$, which top-$k$ similarity with any $k$ below that size cannot ensure.}%
\vspace{-2mm}
\begin{proof}
\emph{Part 1: KnowledgeClaw.} Let $q$ name a handle $h$ and let $a\in\eta_t(h)$. By Eq.~(\ref{eq:route}),
$a\in H_q$, hence $H_q\cup V_q\neq\emptyset$ and Eq.~(\ref{eq:route}) routes $q$ to the \textit{file} or the
\textit{subgraph} granularity, never to the community tier alone. At either granularity the evidence
$\mathcal{X}_q$ opens with the card of every file of $H_q$; the character budget applies only to the linked
works and passages that follow, and no later step removes a card. Hence
\begin{equation}
\Pr\bigl[\,\mathrm{card}(a)\in\mathcal{X}_q\,\bigr]=1\qquad\text{for every }a\in H_q,
\label{eq:p3ours}
\end{equation}
whatever $k$, $\lambda$, $\theta$ or the size of $\mathcal{D}_t$, and $\mathcal{X}_q$ enters the answer prompt
whole. By Proposition~2, $a\in\mathcal{D}_t$, so this card is also current. At the \textit{subgraph}
granularity the named file also seeds the walk: for every file $b$ that shares a record entity
$u\in\mathcal{V}(a)$ with $a$, the third sum of Eq.~(\ref{eq:link}) adds
$\log(1+|\mathcal{D}_t|/|\pi(u)|)>0$ to $s_q(b)$, since $|\pi(u)|\le|\mathcal{D}_t|$ by
Eq.~(\ref{eq:invariant}). Every file that Proposition~1 joins to $a$ through a non-hub entity is thus
ranked by the walk, and the evidence is anchored on the named file before any linked work is added. The guarantee covers a handle that names several files as
well: every $a\in\eta_t(h)$, e.g.\ a PDF and its text twin after a \textsc{Refresh}, enters as its own card.

\emph{Part 2: top-$k$ similarity.} Let a retriever return the $k$ items of a corpus most similar to
$\phi(q)$ under a fixed encoder, and let the corpus hold $k$ files $b_1,\dots,b_k$ whose surfaces embed in
the direction of $\phi(q)$, $\cos(\phi(q),\phi(\tilde b_i))=1$, beside the named file $a$ with
$\cos(\phi(q),\phi(\tilde a))<1$. The retriever returns $b_1,\dots,b_k$ and misses $a$. The same happens,
without exact ties, whenever $k$ look-alike files lie closer to $\phi(q)$ than $a$ does, as the
same-collection candidates of a $\mathcal{T}_2$ item do. More generally, suppose $a$ and $\ell$ look-alike
files are exchangeable to the encoder given $q$, i.e.\ the joint law of their scores is invariant under
permutation, as it is when a content-hash handle carries no meaning an encoder could use; ties are broken
uniformly. By symmetry each of the $\ell+1$ files is equally likely to hold each rank, and the top $k$ hold
$\min(k,\ell+1)$ of them, so
\begin{equation}
\Pr\bigl[\,a\in\mathrm{top}\text{-}k(q)\,\bigr]=\frac{\min(k,\,\ell+1)}{\ell+1}<1\qquad\text{whenever }\ell\ge k .
\label{eq:p3topk}
\end{equation}
Fusing several similarity lists, as Eq.~(\ref{eq:rrf}) does, does not lift this bound: each list is a
top-$2k$ by similarity, and an item absent from every list has fused score $0$. So no choice of $k$ smaller
than the corpus lets similarity search ensure that a named file is retrieved, while Eq.~(\ref{eq:p3ours})
holds for every $k$; KnowledgeClaw therefore uses similarity only to add evidence, never to decide
whether a named file is read. Propositions~1--3 thus give the join, maintenance and lookup
guarantees of C1--C3.
\end{proof}
\rN{Empirically, flat top-$k$ misses a named file on $33.5\%$ of the $\mathcal{T}_2$ items and $76.9\%$ of the $\mathcal{T}_4$
items, while C3 brings every named file into the evidence (Table~\ref{tab:app-rq5}), exactly as Eq.~(\ref{eq:p3ours}) predicts.}
\pagebreak
\clearpage

\section{KnowledgeClaw Algorithm Details}
\label{app:algorithm}

Algorithm~\ref{alg:kc} gives the full workflow of KnowledgeClaw (\S\ref{sec:method}) in three phases:
\textbf{(1) Omnimodal hypergraph construction (C1).} Each file $a$ is read once into a surface $\tilde a$ by
Eq.~(\ref{eq:read}) and catalogued by one call of $\Omega$; the records merge into $\mathcal{K}$ by
Eq.~(\ref{eq:record}), with cross-modal hyperedges and twins by Eq.~(\ref{eq:xmodal}).
\textbf{(2) Closed-loop lifecycle (C2).} Each edit $\Delta_t$ of Eq.~(\ref{eq:edits}) is committed in place:
\textsc{Del} retracts provenance by Eq.~(\ref{eq:del}) with no model call, \textsc{Add} reuses phase~1 on the
added files only, and each commit restores the soundness invariant of Eq.~(\ref{eq:invariant}).
\textbf{(3) Intent-routed multi-granular retrieval (C3).} A query is anchored and routed by Eq.~(\ref{eq:route})
to a tier; the subgraph tier walks from the anchors by Eq.~(\ref{eq:link}) and fuses two dense paths by
Eq.~(\ref{eq:rrf}); one call of $\Omega$ answers from the evidence $\mathcal{X}_q$.

\begin{algorithm}[h!t]
\small
\caption{KnowledgeClaw: construct, maintain and query an omnimodal typed hypergraph}
\label{alg:kc}
\begin{algorithmic}[1]
\Require Persona files $\mathcal{D}$, edit stream $\Delta_1,\Delta_2,\dots$, queries $q$, frozen model $\Omega$, encoder $\phi$
\Ensure Hypergraph $\mathcal{K}_t=(\mathcal{D}_t,\mathcal{V}_t,\mathcal{E}_t,\pi_t)$ and answers $\hat y_q$
\vspace{1mm}
\State \textbf{// 1: Omnimodal Hypergraph Construction (C1)}, \ \textsc{Add}$(\mathcal{D})$ on an empty $\mathcal{K}$
\Function{Add}{$A$}
  \For{each $a\in A$ \textbf{in parallel}} \Comment{model calls write nothing}
    \State $\tilde a\gets\mathrm{proj}(a)$ if $\mu(a)\in\{\textit{text},\textit{pdf}\}$, else $\Omega(\mathrm{raw}(a);p_{\mu(a)})$;\quad $(\sigma_{a,1},\dots,\sigma_{a,n_a})\gets\mathrm{split}(\tilde a)$
    \State Catalogue record: $\bigl(\{Z_{a,\gamma}\}_{\gamma\in\Gamma},\ \text{events}\bigr)\gets\Omega(\tilde a;\,p_{\mathrm{rec}})$
  \EndFor
  \For{each $a\in A$, each field $\gamma\in\Gamma$} \Comment{sequential merge, Eq.~(\ref{eq:record})}
    \State $e_{a,\gamma}\gets\{a\}\cup\{\iota(z):z\in Z_{a,\gamma}\}$;\quad $\mathcal{E}\gets\mathcal{E}\cup\{e_{a,\gamma}\}$;\quad $\pi(e_{a,\gamma})\gets\{a\}$
    \State $\pi(v)\gets\pi(v)\cup\{a\}$ for $v\in e_{a,\gamma}\cap\mathcal{V}$;\quad add alias key $\iota_{\mathrm{p}}(v)$ to class $[v]$ for person names
  \EndFor
  \State Add a dated \emph{event} hyperedge $\{a\}$ per event; add $e^{\times}_v$ and twins $a\approx b$ by Eq.~(\ref{eq:xmodal})
  \State Insert $\phi(v),\phi(e),\phi(\sigma_{a,j})$ into $I_{\mathcal{V}},I_{\mathcal{E}},I_{\sigma}$ and $\tilde a$ into BM25
\EndFunction
\vspace{1mm}
\State \textbf{// 2: Closed-loop Lifecycle (C2)}, \ for each edit $\Delta_t$ of Eq.~(\ref{eq:edits})
\Function{Del}{$h$} \Comment{no model call}
  \For{each $a\in\eta(h)$, each $x\in\mathcal{V}\cup\mathcal{E}$ with $a\in\pi(x)$}
    \State \textbf{if} $\pi(x)=\{a\}$ \textbf{then} delete $x$ and its index keys \textbf{else} $\pi(x)\gets\pi(x)\setminus\{a\}$ \Comment{Eq.~(\ref{eq:del})}
  \EndFor
  \State Drop orphans; recompute touched $e^{\times}_v$; $\mathcal{D}\gets\mathcal{D}\setminus\eta(h)$
\EndFunction
\State \textbf{function} \textsc{Refresh}$(h\!\to\!A')$: \ \Call{Del}{$h$};\ \Call{Add}{$A'$};\ $\eta(h)\gets A'$
\State \rF{After each commit: assert Eq.~(\ref{eq:invariant})}; mark derived indices for rebuild on next use
\vspace{1mm}
\State \textbf{// 3: Intent-routed Multi-granular Retrieval (C3)}, \ for each query $q$
\State $H_q\gets\bigcup_{h\in\mathrm{hdl}(q)}\eta(h)$;\quad $V_q\gets\{\iota(\psi)\ \text{or nearest in}\ I_{\mathcal{V}}\ \text{at}\ \cos\ge\theta:\psi\in\Psi_q\}$;\quad $g\gets g(q)$ by Eq.~(\ref{eq:route})
\If{$g=\textit{file}$}
  \State $\mathcal{X}_q\gets$ full cards of the files in $H_q$
\ElsIf{$g=\textit{subgraph}$}
  \State Score each $b\notin H_q$ (nor its twin) by $s_q(b)$ of Eq.~(\ref{eq:link}), skipping hubs; merge twins into works $w$
  \State $W_q\gets$ works with $\bar s_q(w)\ge0.6\max_{w'}\bar s_q(w')$, at most six
  \State $L_{q,\mathcal{V}}\gets$ hyperedges of the top-$2k$ entities of $I_{\mathcal{V}}$;\quad $L_{q,\mathcal{E}}\gets$ top-$2k$ hyperedges of $I_{\mathcal{E}}$
  \State $\mathcal{X}_q\gets$ cards of $H_q$ $\Vert$ cards of $W_q$ $\Vert$ top-$k$ hyperedges by $\mathrm{rrf}_q$ (Eq.~\ref{eq:rrf}) $\Vert$ four passages of $I_\sigma$
\Else
  \State $\mathcal{X}_q\gets$ the three nearest reports of the $64$ largest Louvain communities, each $\Omega(\cdot;p_{\mathrm{com}})$
\EndIf
\State \Return $\hat y_q=\Omega(q,\mathcal{X}_q;\,p_{\mathrm{ans}})$
\end{algorithmic}
\end{algorithm}

\textit{Complexity analysis.}
Let $N=|\mathcal{D}|$, $n_{\mathrm{m}}$ the number of media files, $R=\sum_a|\mathcal{V}(a)|$ the total record size and $d$ the embedding dimension.
First, construction issues exactly $N+n_{\mathrm{m}}$ calls of $\Omega$ on bounded inputs (Table~\ref{tab:prompts}),
i.e.\ $O(N)$ parallel calls, plus $O(R)$ to merge and $O\bigl((|\mathcal{V}|+|\mathcal{E}|+|\Sigma|)\,d\bigr)$ to index;
no call sees two files together, so no step is quadratic in $N$.
Second, an edit costs only what it touches: $\textsc{Del}(h)$ visits the nodes whose provenance holds a file of $\eta(h)$,
$O(\sum_{a\in\eta(h)}|\mathcal{V}(a)|)$ with no model call, and $\textsc{Add}(A)$ costs $O(|A|)$ calls, independent of $|\mathcal{D}_t|$,
so $T$ edits avoid the $O(TN)$ calls of rebuilding.
Third, a query costs at most two calls of $\Omega$ (routing only if no rule applies, and the answer); the file tier is $O(|H_q|)$,
the walk visits only files linked to an anchor, $O\bigl(\sum_{v\in V_q\cup\mathcal{V}(H_q)}|\pi([v])|\bigr)$, and the two dense paths cost
$O\bigl((|\mathcal{V}|+|\mathcal{E}|)\,d\bigr)$. The evidence is bounded (six works, $k$ hyperedges, four passages), so a query
costs $\$0.0004$ and a $1{,}024$-token budget keeps $55.8$ of the full $55.9$ OKE (\S\ref{sec:rq6}); all three phases thus scale at most linearly with the corpus.

\clearpage

\section{OmniKE-Bench Detail}
\label{app:bench}\label{app:sources}

This appendix details the tasks, construction and metrics of \S\ref{sec:bench}. Persona $d$ has corpus $\mathcal{D}_d$; a \emph{unit} $u$ is one released work
(a document as PDF and text, an image, or a clip), a \emph{record} $r$ one of its files with modality $m(r)\in\mathcal{M}$,
and $\mathcal{K}_d=\mathcal{S}(\mathcal{D}_d)$ the artefact that a system $\mathcal{S}$ commits at \textsc{ingest} time.

\begin{table}[H]
\centering
\caption{\textbf{The six OmniKE tasks.} \rC{All gold comes from human-made metadata or description, deterministic
parsing or official transcripts, never from a model. $M(t)$: evidence modalities (items).}}
\label{tab:tasks}
\small
\begin{tabularx}{\linewidth}{@{}l>{\raggedright\arraybackslash\hsize=1.25\hsize}X>{\raggedright\arraybackslash\hsize=1.6\hsize}X>{\raggedright\arraybackslash\hsize=0.5\hsize}X>{\raggedright\arraybackslash\hsize=0.65\hsize}X@{}}
\toprule
& \textbf{Query $\rightarrow$ answer} & \textbf{Gold source} & \textbf{Metric} & $M(t)$ \\
\midrule
\multicolumn{5}{@{}l}{\textit{\textbf{Construction}}} \\
$\mathcal{T}_1$ & PDF or image $+$ closed type inventory $\rightarrow$ its entities of those types
 & catalogued authors, affiliations, subject terms, depicted persons; parsed floor-plan elements & set F1 & pdf 344, image 56 \\
$\mathcal{T}_2$ & media file $\rightarrow$ its authoritative description among $4$ candidates
 & catalogue record, creator description or transcript; all four candidates from one collection, each of a different subject & accuracy & image 397, audio/video 3 \\
\multicolumn{5}{@{}l}{\textit{\textbf{Application}}} \\
$\mathcal{T}_3$ & 2--4-hop question $\rightarrow$ one catalogued value
 & deterministic join over catalogue records; each join condition is one hop & set F1 & image 275, pdf 125 \\
$\mathcal{T}_4$ & as $\mathcal{T}_3$, but the chain joins image and PDF records (2--3 hops)
 & as above & set F1 & image$+$pdf 110 \\
$\mathcal{T}_6$ & PDF or image $+$ candidate labels $\rightarrow$ those assigned to it
 & human-assigned controlled-vocabulary labels; negatives never co-assigned with its labels & balanced acc. & image 304, pdf 78 \\
\multicolumn{5}{@{}l}{\textit{\textbf{Maintenance}}} \\
$\mathcal{T}_5$ & a document's title, authors or abstract; answer, commit $\Delta$, answer again
 & official version chain; the corpus holds the prior version, $\Delta$ the next & accuracy, both phases & pdf 137 \\
\bottomrule
\end{tabularx}
\end{table}

\subsection{Task Design}
\label{app:bench-task}

\paragraph{Canonical contracts.} Adapters lift each source onto nine contracts: REC (a record with its unit, modality
and parent), ORD (order by date, version or recording time), MEN (an entity of a catalogued class), ALN (a cross-modal
alignment), IDN (an agent identity with its identifier basis), EVT (a dated event), STA (a state value in one version),
NRM (a validity norm and its deviations) and COV (an assigned controlled-vocabulary term). Every row carries its
provenance, and builders read contracts only, so each task is one builder that runs unchanged over all ten personas alike.

\noindent\begin{minipage}[t]{0.5836\textwidth}\setlength{\parskip}{.5pc}
\paragraph{Instantiation.} \leavevmode\rC{$\mathcal{T}_1$--$\mathcal{T}_3$ run in every persona, the other three tasks wherever their structure
yields at least $20$ items (Table~\ref{tab:corpus}), never padded with model-made items: $\mathcal{T}_4$ needs two modalities
sharing authority-controlled entities (in healthcare, NLM subject headings of CDC images and PubMed Central papers),
$\mathcal{T}_5$ an official version chain and $\mathcal{T}_6$ a controlled vocabulary with enough balanced labels.}

\paragraph{$\mathcal{T}_1$ (knowledge acquisition).} Given a unit $u$ (a PDF or an image) and a closed type inventory
$\Gamma=\smash[b]{\Gamma_{d,m(u)}}$, return the entities of those types in $u$, $\mathcal{T}_1:(u,\Gamma)\mapsto\smash[t]{\hat E}$, with gold
$E^\ast(u)=\{\mathrm{surf}(x): x\in\mathrm{MEN}(u),\ \mathrm{cls}(x)\in\Gamma,\ \mathrm{own}(x,u)\}$. The rule $\mathrm{own}(x,u)$
keeps only what the unit itself states: a whole-word surface in its own human text or aligned transcript, or, for an image,
a row from a human structural layer or a class naming what is depicted. \rN{$\Gamma_{d,m}$ holds every class with such a row
in $(d,m)$, never tailored to any single item}; on average, an item asks for $2.4$ entity types and holds $6.0$ gold entities of those types.
\end{minipage}\hfill
\raisebox{\ht\strutbox}[\ht\strutbox][0pt]{\begin{minipage}[t]{0.39\textwidth}
\captionof{table}{\textbf{Items per persona and task.} \rC{Each persona has $90$ documents (PDF and text), $180$ images, $18$ audio
and $18$ video clips ($2.9$, $2.4$\,min on average); $15.8$\,h of media, $6.9$\,GB in all. A blank cell: the task is not run in that persona.}}
\label{tab:corpus}\label{tab:items}
\small
\begin{tabular*}{\linewidth}{@{\extracolsep{\fill}}lccccccr@{}}
\toprule
 & $\mathcal{T}_1$ & $\mathcal{T}_2$ & $\mathcal{T}_3$ & $\mathcal{T}_4$ & $\mathcal{T}_5$ & $\mathcal{T}_6$ & \textbf{All} \\
\midrule
\textsc{HC}  & 40 & 40 & 41 & 110 &    & 166 & 397 \\
\textsc{Fin} & 41 & 40 & 41 &     & 29 &     & 151 \\
\textsc{Edu} & 40 & 40 & 41 &     & 22 & 46  & 189 \\
\textsc{Law} & 36 & 40 & 41 &     &    &     & 117 \\
\textsc{Psy} & 40 & 40 & 40 &     &    & 170 & 290 \\
\textsc{Pol} & 40 & 40 & 41 &     &    &     & 121 \\
\textsc{Hum} & 41 & 40 & 41 &     &    &     & 122 \\
\textsc{Bld} & 40 & 40 & 32 &     & 29 &     & 141 \\
\textsc{Mus} & 41 & 40 & 41 &     & 29 &     & 151 \\
\textsc{Gam} & 41 & 40 & 41 &     & 28 &     & 150 \\
\midrule
Total & 400 & 400 & 400 & 110 & 137 & 382 & 1{,}829 \\
\bottomrule
\end{tabular*}
\end{minipage}}

\begin{table}[t]
\centering
\caption{\textbf{Sources per persona.} PD: public domain (US federal work); WC: Wikimedia Commons, pages before
30 Nov 2022, each file keeping its licence (PD, CC0, CC BY or CC BY-SA); OCW: MIT OpenCourseWare. \textsuperscript{\textsc{nc}}:
CC BY-NC-SA, kept per file. No source carries a no-derivatives term.}
\label{tab:sources}
\small
\begin{tabularx}{\linewidth}{@{}l>{\raggedright\arraybackslash\hsize=0.95\hsize}X>{\raggedright\arraybackslash\hsize=1.15\hsize}X>{\raggedright\arraybackslash\hsize=1.0\hsize}X>{\raggedright\arraybackslash\hsize=0.9\hsize}X@{}}
\toprule
\textbf{Persona} & \textbf{text / pdf} & \textbf{image} & \textbf{audio} & \textbf{video} \\
\midrule
Healthcare & PubMed Central OA (CC BY) & CDC Public Health Image Library (PD) & PriMock57 (CC BY); Senate HELP hearings (PD) & Senate HELP hearings (PD) \\
Finance & arXiv q-fin & WC & FOMC press conferences (PD); Earnings-21/22 (CC BY-SA) & FOMC press conferences with official transcripts (PD) \\
Education & arXiv & Smithsonian Open Access (CC0) & OCW\textsuperscript{\textsc{nc}} & OCW\textsuperscript{\textsc{nc}} \\
Law & SCOTUS opinions (PD); DOAJ CC-BY journals & WC & SCOTUS oral arguments, official transcripts (PD) & Senate Judiciary hearings (PD) \\
Psychology & PsyArXiv, PMC & Wellcome Collection (PD/CC0/CC BY/BY-SA) & OCW 9.00\textsuperscript{\textsc{nc}} & OCW 9.00, captioned lectures\textsuperscript{\textsc{nc}} \\
Politics & CRS reports with version chains (PD); World Bank OKR (CC BY) & WC & VoxPopuli, European Parliament (CC0) & White House briefings, official captions (PD) \\
Humanities & DOAJ CC-BY humanities journals & Met and Cleveland Museum Open Access with curatorial records (CC0) & Open Yale Courses\textsuperscript{\textsc{nc}}; LibriVox with Gutenberg text (PD) & Open Yale Courses\textsuperscript{\textsc{nc}} \\
Building & arXiv; building codes (CC0); NIST (PD) & CubiCasa5K plans\textsuperscript{\textsc{nc}}; CMP facades (CC BY-SA) & OCW planning\textsuperscript{\textsc{nc}} & OCW planning; site footage\textsuperscript{\textsc{nc}} \\
Music & arXiv & Met musical instruments (CC0) & OpenMIC-2018 labels (CC BY); FMA, no NC/ND & OCW 21M, captioned lectures\textsuperscript{\textsc{nc}} \\
Games & arXiv & WC screenshots; OpenGameArt (CC0) & OpenGameArt, creator descriptions (CC0) & WC, captioned game videos (CC BY) \\
\bottomrule
\end{tabularx}
\end{table}

\paragraph{$\mathcal{T}_2$ (cross-modal coreference).} Given a media record $r$, choose among four candidates the
description joined to $r$ by an ALN row, \mbox{$\mathcal{T}_2:(r,\{c_1,\dots,c_4\})\mapsto\hat\imath$}. All candidates come
from the same parent $p(r)$ (transcript, explicit parent, partition key or collection), so distractors are of the same kind
and place; descriptions have at least $40$ characters, and no two candidates share a subject group.

\paragraph{$\mathcal{T}_3$ (multi-hop retrieval).} Given $q$ encoding a join path
$\pi=r_0\xrightarrow{j_1}\cdots\xrightarrow{j_h}v$, $h\in\{2,3,4\}$, return the terminal value,
\mbox{$\mathcal{T}_3:(q,\mathcal{K}_d)\mapsto\hat a$}. Each hop is one contract join (an ALN step, the one other record sharing
a field, a unique-entity lookup or a terminal value step). A path is admissible only if every
record-valued hop yields one unvisited record, the terminal hop yields one value and no earlier record
gives the answer, so every hop is needed. The $400$ items have $300$, $72$ and $28$ paths of $2$, $3$ and $4$ hops.
\textbf{$\mathcal{T}_4$ (cross-modal reasoning)} is the same over paths through two or more modalities, one
non-textual; its $110$ items, disjoint from $\mathcal{T}_3$, join image and PDF records via NLM headings.

\paragraph{$\mathcal{T}_5$ (lifecycle consistency).} Over an official version chain, the system answers $q$ on a document's
state attribute (phase A, gold $v_{\mathrm{old}}$), commits the released next version
$\Delta=(\textsc{Refresh},A_\Delta)$ with $A_\Delta\cap\mathcal{D}_d=\emptyset$, and answers again (phase B, gold
$v_{\mathrm{new}}\neq v_{\mathrm{old}}$). Attributes are titles ($49$ items), abstracts ($69$) and author lists ($19$); spelling-only
pairs are dropped, and no stem or option carries a version label, date or other order cue, so only the
committed edit separates the two phases.

\paragraph{$\mathcal{T}_6$ (knowledge discovery).} Given a unit $u$ and candidates $C=C^+\cup C^-$ from one vocabulary, return the covered ones, \mbox{$\mathcal{T}_6:(u,C)\mapsto\hat C^+\subseteq C$}. $C^+$ holds the unit's
human-assigned terms and $C^-$ terms never co-assigned with them, with $|C^+|=|C^-|\geq2$; vocabularies are MeSH ($171$
items), LCSH ($127$), Smithsonian object types ($46$), LC Names ($24$) and three others ($14$).

\subsection{Benchmark Construction}
\label{app:bench-build}

The four phases of Figure~\ref{fig:omnike}C run as seven stages, each SHA-256-ordered, fully cached and score-blind.
\textbf{(1) Collect.} Sources (Table~\ref{tab:sources}) obey three rules. \rC{Gold must be human-made: ASR, OCR
and model descriptions may be system input but never gold, ruling out three ASR-captioned video archives.} Media
must be redistributable, and crowd-written gold is limited to pages created before 30 Nov 2022.
\textbf{(2) Normalise and lift.} Every file is re-encoded, renamed by content hash and stripped of metadata; images
come with their own catalogue record, never cut from a paper. Adapters map each annotation to one contract row.
\textbf{(3) Instantiate, rewrite and verify.} A template filled from contract rows yields a stem, its slots and its
answer. \rF{GPT-5-nano paraphrases the stem blind to answer and options; six deterministic checks require verbatim
slots, numbers and dates, no answer in the stem and no task-declared pattern, e.g.\ a version label
in $\mathcal{T}_5$; failing items are discarded, never repaired.}
\textbf{(4) Balance and (5) recompute.} No answer value exceeds $10\%$ of a task or persona share,
$\mathcal{T}_2$ and $\mathcal{T}_5$ rotate gold positions, and frequent $\mathcal{T}_6$ labels are positive in $1/3$ to $2/3$
of their uses. An independent second implementation recomputes every answer from
the raw sources and agrees on all $1{,}829$ items.
\textbf{(6) Gate and (7) freeze.} Gates confirm that no released file keeps an original name, source URL or gold sentence;
image-level records drop a redaction-box probe on $\mathcal{T}_2$ from $77.3$ to chance ($25.0$--$27.5$). Each persona is then cut to Table~\ref{tab:corpus}'s size,
keeping every unit an item touches, so no gold answer changes; a SHA-256 manifest seals the release, and one command
rebuilds every item from the raw sources, so the benchmark itself is fully reproducible from public data.

\subsection{Evaluation Metric Design}
\label{app:bench-metric}

\paragraph{Protocol.} The harness builds $\mathcal{K}_d$ once per persona and answers every task, each item
in a fresh context, with a value list ($\mathcal{T}_1,\mathcal{T}_3,\mathcal{T}_4$), an option ($\mathcal{T}_2$, $\mathcal{T}_5$ per
phase) or a candidate subset ($\mathcal{T}_6$); \rF{a missing or unparseable answer scores $0$.} Reads are not sandboxed (MMGraphRAG
reopens images at query time); \rF{answers are scored by the deterministic functions below, never by a model.}

\paragraph{Lenient value matching.} A predicted value $p$ and a gold value $g$ have similarity $\sigma(p,g)$, where
$\mathrm{nrm}(\cdot)$ is the normalisation (lower-case, trim punctuation, drop a leading article, collapse spaces),
$\mathrm{tok}(\cdot)$ the lower-cased token set and $\gamma(g)\in\{\text{year},\text{month},\text{day}\}$ the precision of a gold date:
\begin{equation}
\sigma(p,g)=\begin{cases}
\mathbf{1}\bigl[\mathrm{date}(p)\equiv_{\gamma(g)}g\bigr] & g \text{ a date of granularity } \gamma(g),\\
1 & \mathrm{nrm}(p)=\mathrm{nrm}(g),\\
2|\mathrm{tok}(p)\cap\mathrm{tok}(g)|/(|\mathrm{tok}(p)|+|\mathrm{tok}(g)|) & \text{otherwise},
\end{cases}
\end{equation}
\rF{so ``May 24, 2022'' matches ``2022-05-24'' and a short affiliation its full catalogued form. Pairs with
$\sigma\geq\tau=0.5$ are matched one-to-one, greedily by decreasing $\sigma$, giving the match set $\mathcal{M}(\hat A,A^\ast)$.}

\paragraph{Item and task scores.} With $\hat A$, $A^\ast$ the predicted and gold values and $\hat C^+$ the selected
candidates, $s_2=\mathbf{1}[\hat\imath=g]$ for gold option $g$,
$s_5=\tfrac12(\mathbf{1}[\hat a_{\mathrm A}=g_{\mathrm A}]+\mathbf{1}[\hat a_{\mathrm B}=g_{\mathrm B}])$, and
\begin{equation}
s_{1,3,4}=\frac{2|\mathcal{M}(\hat A,A^\ast)|}{|\hat A|+|A^\ast|},\qquad
s_6=\frac{|\hat C^+\cap C^+|}{2|C^+|}+\frac{|C^-\setminus\hat C^+|}{2|C^-|}.
\end{equation}
On $\mathcal{T}_5$ always returning the old value, or always the new, scores exactly $50$, as does selecting all or
none of the candidates on $\mathcal{T}_6$, so only a committed edit and real discrimination rise above half credit. As $s\in[0,1]$ has variance at most $1/4$, the worst-case $95\%$ half-width of an $n$-item mean is $98/\sqrt n$: $\pm4.9$ for $\mathcal{T}_1$--$\mathcal{T}_3$,
$\pm5.0$ for $\mathcal{T}_6$, $\pm8.4$ for $\mathcal{T}_5$ and $\pm9.3$ for $\mathcal{T}_4$; per-persona scores are descriptive.

\noindent\begin{minipage}[c]{0.51\textwidth}
\paragraph{Admission references.} Table~\ref{tab:refs} gives the mechanical floor $f$ (best of position, length,
overlap and prior heuristics), the zero-model shortcut probe $P$, closed-book $C$, flat retrieval $R$ and
$R_{-m}$, with $C$ and $R$ the best of two models before the cut. $C$ is $0.0$ on the three free-form tasks and below the floor on $\mathcal{T}_2$ and $\mathcal{T}_5$, so neither
the stem nor memorised knowledge answers it; \rP{$R$ reaches only $2.5$ and $0.0$ on the joins $\mathcal{T}_3$ and
$\mathcal{T}_4$, so retrieval alone does not compose records}; and $R_{-m}$ falls below $R$ wherever $R$ scores, by up to
$56.0$ on $\mathcal{T}_2$, so the named modality is needed. Neither $P$ nor $C$ exceeds $R$ on any task, so no shortcut
beats reading the corpus, the skill OmniKE-Bench measures.
\end{minipage}\hfill
\begin{minipage}[c]{0.46\textwidth}
\captionof{table}{\textbf{Reference scores.} $R$: media bridge and three BM25 passages; $R_{-m}$: $R$ without the item's modality.}
\label{tab:refs}
\small
\begin{tabular*}{\linewidth}{@{\extracolsep{\fill}}lcccccc@{}}
\toprule
 & $\mathcal{T}_1$ & $\mathcal{T}_2$ & $\mathcal{T}_3$ & $\mathcal{T}_4$ & $\mathcal{T}_5$ & $\mathcal{T}_6$ \\
\midrule
Items $n$         & 400 & 400 & 400 & 110 & 137 & 382 \\
Floor $f$         & 12.1 & 33.5 & 2.0 & 0.0 & 50.0 & 50.0 \\
Probe $P$         & 0.1 & 26.6 & 0.3 & 0.0 & 25.4 & 49.2 \\
Closed-book $C$   & 0.0 & 32.3 & 0.0 & 0.0 & 35.8 & 58.4 \\
Retrieval $R$     & 5.3 & 82.3 & 2.5 & 0.0 & 64.2 & 69.9 \\
$R_{-m}$          & 0.0 & 26.3 & 0.0 & 0.0 & 32.8 & 50.3 \\
\bottomrule
\end{tabular*}
\end{minipage}

\clearpage

\section{Baseline Detail}
\label{app:baseline}
\begingroup\small\setlength{\parskip}{3pt plus 1pt minus 1pt}

This appendix specifies how every row of Table~\ref{tab:main} is deployed: the shared protocol
(\S\ref{app:bl-protocol}), the $30$ baselines in the order and grouping of the table
(\S\ref{app:bl-graph}--\S\ref{app:bl-agentic}), and KnowledgeClaw with its three ablations (\S\ref{app:bl-kc}). For each
method we state what \textsc{ingest} builds, what a query retrieves and how the $\mathcal{T}_5$ edit $\Delta$ is committed;
every number is the value used in our runs, and a token is $4$ characters throughout, as in the evidence budgets of \S\ref{sec:rq6}.

\subsection{Shared deployment protocol}
\label{app:bl-protocol}

\leavevmode\rC{\textbf{One frozen stack.} Every method calls the same model $\Omega$ (\textsc{Gemini~2.5~Flash~Lite}) and encoder
$\phi$ (\textsc{Gemini~Embedding~2}, $3{,}072$ dimensions) through one metering proxy that caches each call by the SHA-256 of
model, messages and parameters, at temperature $0$ unless a method's official default differs (MemPro, $0.3$).
\textbf{One input.} A text or PDF file enters as its opening $4{,}000$ tokens, cut at a paragraph break and headed by its
file name, the only handle items use. Text-reading methods receive image, audio and video through the shared media-to-text
bridge (image captions, time-windowed transcripts); the omnimodal group and KnowledgeClaw read raw media natively.
\textbf{One answerer.} Whatever a method retrieves, or gathers with its own agent loop, ends in the answer prompt of
Appendix~\ref{app:prompt-answer} with $600$ output tokens, so methods differ only in the evidence they supply.
\textbf{One protocol.} Per persona a method builds once over the base corpus and answers each item in a fresh context at
$24$-way concurrency. On $\mathcal{T}_5$ all phase-A answers come first, then every item's $\Delta$ (the next released
version, headed ``a new version of $h$'') is committed through the method's own write path, then all phase-B answers, so no
phase-A answer can see an edit. Every method attempts all six tasks.
\textbf{Faithful re-implementation.} Each baseline is re-implemented on this stack from its official code (from the paper
where none is released: OntoKG, M$^3$KG-RAG), with its prompts copied verbatim and its defaults kept; in-process indices
replace databases, the frozen stack replaces each method's own models (NV-Embed, MiniLM, CLIP, Whisper, Qwen), and
gleaning is $0$ wherever offered. Other deviations are listed per method; costs are in Table~\ref{tab:cost}, metered by the same proxy for every method and split into ingest and query by when each call was made, with the shared media-to-text bridge charged once for the whole benchmark rather than per method.}

\subsection{No knowledge engineering, chunk, graph and agentic graph RAG}
\label{app:bl-graph}

\textbf{NaiveGeneration}~\citep{Gemini25FlashLite}. No ingest, retrieval or write: $\Omega$ answers from the question and
options alone, the closed-book control, which shows what the frozen model knows without any corpus.

\textbf{StandardRAG}~\citep{RAG}. \emph{Ingest:} $1{,}200$-token windows with $100$ overlap, embedded by $\phi$.
\emph{Query:} cosine top-$5$ windows. \emph{Edit:} the windows of $\Delta$ are appended beside those of the prior version.

\textbf{RAPTOR}~\citep{RAPTOR}. \emph{Ingest:} sentence-bounded leaves of ${\le}100$ tokens; up to $5$ layers of soft GMM
clustering ($k\le50$ by BIC, membership $p>0.1$, PCA to $10$ dimensions for UMAP, clusters over $3{,}500$ tokens
re-clustered), each cluster summarised in ${\le}100$ tokens. \emph{Query:} collapsed-tree top-$10$ over all layers within
$3{,}500$ tokens. \emph{Edit:} the tree is rebuilt over the base corpus and $\Delta$, which still holds the prior version.

\textbf{LightRAG}~\citep{LightRAG} (v1.4.14). \emph{Ingest:} $1{,}200/100$-token chunks; one call per chunk extracts
entities of $11$ types and keyworded relations, merged by name and pair with map-reduce description summaries.
\emph{Query:} hybrid mode; a keyword call sends low-level keys to the entity store and high-level keys to the relation store
(top-$40$ each, cosine $>0.2$), plus $20$ related chunks under $6{,}000/8{,}000/30{,}000$-token entity, relation and total
budgets. \emph{Edit:} native incremental insert, merging the entities and relations of $\Delta$ by name and pair.

\textbf{GraphRAG}~\citep{GraphRAG}. \emph{Ingest:} $1{,}200/100$-token units; one call per unit extracts entities of four
types and weighted relations, descriptions are summarised in ${\le}500$ words, recursive Louvain (communities above $10$
members split again) stands in for hierarchical Leiden, and each community receives a rated report. \emph{Query:} local
search from the top-$20$ entities within $12{,}000$ tokens, split $15/35/50\%$ over reports, entity-relation tables and
text units. \emph{Edit:} the official update mode extracts $\Delta$, merges by title and pair, and re-summarises.

\textbf{HippoRAG\,2}~\citep{HippoRAG2}. \emph{Ingest:} one passage per file; OpenIE by an NER call and a conditioned
triple call; phrase, fact and passage nodes with synonymy edges at cosine $\ge0.8$. \emph{Query:} the top-$5$ facts pass the
recognition-memory filter ($10$ shots) and seed Personalized PageRank (damping $0.5$, passage weight $0.05$);
returning the top-$5$ passages. \emph{Edit:} incremental indexing, synonymy edges recomputed.

\textbf{KAG/OpenSPG}~\citep{KAG}. \emph{Ingest:} $1{,}200/100$-token chunks; three schema-free calls a chunk (NER,
standardisation, triples) into an entity-chunk graph with a PageRank matrix. \emph{Query:} its solver plans a logical form,
runs ${\le}3$ retrieval steps (exact linking at $0.9$, else fuzzy at $0.8$ with PPR chunks, RRF-fused with
vector chunks at $0.65$), summarising each step, ${\le}5$ calls. \emph{Edit:} incremental extraction, PPR matrix rebuilt.

\textbf{Youtu-GraphRAG}~\citep{Youtu-GraphRAG}. \emph{Ingest:} $1{,}200/100$-token chunks (official $5{,}000/200$), one
schema-guided call each, into a four-level tree of attributes, triples, keywords and ${\le}100$ FastTreeComm communities.
\emph{Query:} the agent splits $q$ into ${\le}4$ sub-questions, retrieves triples by node, relation and $3$-hop paths
(top-$20$), and runs IRCoT for ${\le}2$ steps (official $5$). \emph{Edit:} full rebuild of the tree over the base corpus and $\Delta$.

\subsection{Knowledge graph construction and $n$-ary / hypergraph KE}
\label{app:bl-kgc}

iText2KG, OntoKG, Hyper-KGGen and ATOM construct a graph but define no query path; their facts, each tagged with its source
file, are served by one shared layer, cosine top-$30$ over $\phi$, the same for all four.

\textbf{iText2KG}~\citep{iText2KG}. \emph{Ingest:} $1{,}200$-token chunks are distilled into facts, one section per file;
entity and relation calls follow, entities resolve on exact (name, label) match or cosine $>0.92$ ($0.6$ name $+$ $0.4$
label), relation names at $0.9$, and one round revisits isolated entities (official $3$); the official $0.8/0.7$ thresholds
collapsed relations onto ${\sim}30$ names and were recalibrated. \emph{Edit:} $\Delta$ is ingested onto the graph.

\textbf{OntoKG}~\citep{OntoKG}. \emph{Ingest:} one call designs a modular schema over $8$ seed categories from $24$ sampled
files, one refinement call repairs the types and properties a probe extraction leaves unrouted, and $600$-token chunks are
extracted with the paper's Listing~2 and routed to modules by exact or cosine-$\ge0.9$ match. \emph{Edit:} $\Delta$ is
extracted and routed into the frozen schema.

\textbf{HyperGraphRAG}~\citep{HyperGraphRAG}. \emph{Ingest:} $1{,}200/100$-token chunks; one call per chunk yields $n$-ary
knowledge segments with scored entities, stored as a bipartite hyperedge-entity graph. \emph{Query:} entity and segment
keys parsed from $q$ drive a local (top-$60$ entities) and a global (top-$60$ hyperedges) path at cosine $>0.2$, returning
entities, hyperedges and source chunks, $4{,}000$ tokens each. \emph{Edit:} incremental insert.

\textbf{Hyper-KGGen}~\citep{Hyper-KGGen}. \emph{Ingest:} the official recursive splitter ($1{,}000$ characters, $200$
overlap); per chunk a mention call and a hyperedge call over the known nodes; nodes merge by name and hyperedges by member
set, and the skill library is empty as none is released. \emph{Edit:} $\Delta$ runs the same pipeline and merges in.

\textbf{HyperRAG}~\citep{HyperRAG}. \emph{Ingest:} the HyperGraphRAG hypergraph, built with its own calls. \emph{Query:}
topic entities of $q$ seed an LLM-scored beam search over hyperedges and entities (width $3$, depth $3$, ${\le}100$
candidates per entity) with a sufficiency check per depth; paths and their chunks enter within $12{,}000$ tokens.
\emph{Edit:} incremental insert through the HyperGraphRAG write path, which keeps the prior version in the graph.

\subsection{Memory agents and self-evolving lifecycle}
\label{app:bl-mem}

\textbf{MemGPT/Letta}~\citep{MemGPT}. \emph{Ingest:} $512/50$-token passages into archival memory beside the core
\texttt{persona} and \texttt{human} blocks. \emph{Query:} a tool-calling loop under the verbatim \texttt{letta\_v1} system
prompt searches archival memory (top-$5$, ${\le}20$) and edits core memory for ${\le}5$ steps. \emph{Edit:} passages are
appended.

\textbf{A-MEM}~\citep{A-MEM}. \emph{Ingest:} one note per $1{,}200/100$-token chunk, analysed into keywords, context and
tags; each new note runs the evolution call over its $5$ nearest notes, strengthening links or updating neighbours.
\emph{Query:} top-$5$ notes and up to $5$ linked notes. \emph{Edit:} $\Delta$ is added and evolved.

\textbf{Mem0}~\citep{Mem0}. \emph{Ingest:} a fact-extraction call per chunk, then one update call over each fact's top-$5$
memories decides \textsc{Add}, \textsc{Update}, \textsc{Delete} or \textsc{None}. \emph{Query:} top-$100$ memories with their
sources. \emph{Edit:} the same path, so $\Delta$ can update or delete memories.

\textbf{Graphiti (Zep)}~\citep{Zep}. \emph{Ingest:} one episode per chunk at a reference time; entity extraction,
resolution (exact name, $3$-gram Jaccard $\ge0.9$, else an LLM dedupe over $15$ candidates), fact extraction with validity
times, and edge resolution that invalidates contradicted facts. \emph{Query:} RRF of BM25 and cosine over facts and
entities (top-$10$ each) and BM25 episodes, with validity dates. \emph{Edit:} $\Delta$ is ingested at a later time and
invalidates what it contradicts. To fit the shared budget, a fact call has $3{,}072$ output tokens and $15$ facts.

\textbf{MIRIX}~\citep{MIRIX}. \emph{Ingest:} a meta-manager routes each chunk to the core, episodic, semantic, resource and
knowledge-vault managers, each writing its store by a JSON function call against its BM25 top-$10$; procedural memory stays
empty, as released. \emph{Query:} a topic call, then core memory, the $10$ latest episodes and BM25 top-$10$ per store.
\emph{Edit:} the managers insert, merge or replace items.

\textbf{MemOS}~\citep{MemOS}. \emph{Ingest:} the structured document reader turns each chunk into key-value-tag memories,
with a $20$-item working memory. \emph{Query:} a task-parse call yields keys, tags and $2$--$5$ expansions; graph recall
(key or ${\ge}2$ tags) and vector recall are reranked by cosine to the top-$20$. \emph{Edit:} new memories are added.

\textbf{MemEvolve}~\citep{MemEvolve}. It evolves agent memory and has no document ingest, so each chunk enters its
fusion-memory provider as a successful trajectory ``memorise $\langle$file$\rangle$''. \emph{Ingest:} a summarise call
writes insight, pattern, playbook and checklist nodes, linked at similarity $\ge0.75$ and consolidated every $50$ ingests.
\emph{Query:} a focus call, hybrid search ($0.2$ TF-IDF $+$ $0.8$ cosine, top-$5$, $3$-neighbour expansion) and a synthesis
call give a ${\le}350$-character note. \emph{Edit:} $\Delta$ is written through the same path.

\textbf{ATOM}~\citep{ATOM}. \emph{Ingest:} $380$-token chunks stamped with the date each file states; atomic facts, then
temporal $5$-tuples $(s,p,o,t_{\mathrm{start}},t_{\mathrm{end}})$, merged with entities at cosine $\ge0.9$ and served by the
shared fact layer of \S\ref{app:bl-kgc}. \emph{Edit:} $\Delta$ is observed later; a relation its old file supported that the
new file does not re-observe loses that source, and is closed at $t_{\mathrm{end}}$ once none remains.

\textbf{MemPro}~\citep{MemPro}. The released evolved program is used without re-evolution. \emph{Ingest:} $2{,}000$-token
pages, each abstracted by its memory agent, into dense and BM25 indices. \emph{Query:} planning over all abstracts, keyword,
vector and page search (top-$5$), focused excerpts of ${\le}3$ pages, integration and reflection, for one research
iteration (official $3$) to fit the shared budget. \emph{Edit:} new pages are abstracted against the memory and appended, so the prior version's pages stay searchable ($19.0\%$ of its phase-B answers superseded, Table~\ref{tab:app-t5}).

\subsection{Agentic knowledge engineering and omnimodal knowledge engineering}
\label{app:bl-agentic}

OpenClaw, LLM Wiki and OKF build markdown knowledge bases for an agent to read, so they share one reader: hybrid search
($0.7$ cosine $+$ $0.3$ BM25, top-$6$ over $1{,}500$-character windows), open and list tools, ${\le}5$ steps and
${\le}24{,}000$ characters gathered, so they differ only in the knowledge base each one writes.

\textbf{OpenClaw}~\citep{OpenClaw}. \emph{Ingest:} the corpus is split into $32$ sessions; each file is one agent turn under
the verbatim \texttt{AGENTS.md} memory rules, appending to the session's daily note, with the pre-compaction flush beyond
$32{,}000$ tokens. \emph{Edit:} $\Delta$ runs as a new session that writes a new note.

\textbf{Graphify}~\citep{Graphify}. \emph{Ingest:} the verbatim skill prompt extracts nodes and edges from batches of $22$
files (one per image); Louvain (seed $42$) is its fallback for Leiden. \emph{Query:} term hits in label and source pick $5$
start nodes, expanded by BFS to depth $2$ within $2{,}000$ tokens. \emph{Edit:} \texttt{--update} on unseen files.

\textbf{Cognee}~\citep{Cognee}. \emph{Ingest:} \texttt{cognify}; per file a graph-extraction and a summary call into
document, chunk, summary, entity and type nodes. \emph{Query:} graph completion; top-$100$ per collection, unmatched elements
penalised $6.5$, and the $5$ best triplets by summed distance. \emph{Edit:} \texttt{cognify} run on $\Delta$ alone.

\textbf{LLM Wiki}~\citep{LLMWiki}. The idea file gives no prompts, so our schema instantiates it. \emph{Ingest:} in waves of
$16$, each source gets a page and a plan over its $12$ nearest entity and concept pages (${\le}5$ updates, ${\le}3$ new),
then one merge rewrite per touched page. \emph{Edit:} a new version plans against its old source page first, then merges.

\textbf{OKF}~\citep{OKF}. \emph{Ingest:} one concept document per file, written by the reference instruction with the
concept list in view, plus per-directory indices and a log. \emph{Edit:} a new version refines its base concept in place, with the existing document as input and the update logged, without ever duplicating the concept page.

\textbf{Omni-SimpleMem}~\citep{Omni-SimpleMem}. \emph{Ingest:} multimodal atomic units from raw media: $500/50$-token text
chunks, redundancy-filtered and summarised; two captions per image; audio transcripts; video at $1$\,fps through the HSV
scene trigger, ${\le}24$ captioned frames and its audio. \emph{Query:} dense top-$10$ and BM25 top-$10$. \emph{Edit:} $\Delta$
adds units beside those of the prior version, which both searches can still return after the edit is committed.

\textbf{MMGraphRAG}~\citep{MMGraphRAG}. \emph{Ingest:} a text graph ($1{,}200/100$-token chunks, transcripts included)
fused with an image graph built from the raw images (description, image entities, enhancement, whole-image matching by
spectral clustering), one graph per persona. \emph{Query:} top-$5$ nodes at $0.2$ with their tables, plus a vision answer
over each of ${\le}3$ retrieved images. \emph{Edit:} a full rebuild over the base corpus and $\Delta$, as at ingest.

\textbf{M$^3$KG-RAG}~\citep{M3KG-RAG}. \emph{Ingest:} $2{,}000$-token text items and a raw-media caption per image, audio
and video file; triples by the extractor and normaliser, descriptions by the searcher-inspector (pass $\ge7$, ${\le}3$
rounds). \emph{Query:} items within $L_2\le1.1$ of the top-$5$ plus named files, $2$-hop expansion scored by
$\cos(h,q)+\cos(t,q)$, and GRASP pruning. \emph{Edit:} $\Delta$ adds its items, captions and triplets to the graph built at ingest time.

\subsection{KnowledgeClaw and its ablations}
\label{app:bl-kc}

\textbf{KnowledgeClaw.} \emph{Ingest (C1):} text and PDF from the shared projection; image (${\le}1{,}536$\,px), audio
($16$\,kHz) and video (${\le}6$ keyframes, one per $20$\,s, with audio) read natively by $\Omega$ (Appendix~\ref{app:prompt-read}), all $2{,}160$ media files of the benchmark;
surfaces split into ${\sim}1{,}000$-token segments; one record call per file over its first $9{,}000$ characters fills $13$
typed fields, title and events; record, event and cross-modal hyperedges and twins by
Eqs.~(\ref{eq:record})--(\ref{eq:xmodal}); indices $I_{\mathcal{V}},I_{\mathcal{E}},I_{\sigma}$ and BM25. \emph{Query (C3):}
anchors at $\theta=0.86$ and routing by Eq.~(\ref{eq:route}); $7{,}000$-character anchor cards; the walk of
Eq.~(\ref{eq:link}) keeps works within $0.6$ of the best, at most six; RRF with $\lambda=4$, $k=10$, and four passages
(Eq.~\ref{eq:rrf}), within $22{,}000$ characters; $64$ Louvain communities (seed $0$) reported in ${\le}200$ words; one
retrieval pass and one answer call. \rF{\emph{Edit (C2):} a new version of a live handle $h$ commits as
$\textsc{Refresh}(h\!\to\!A')$: \textsc{Del} with provenance pruning, \textsc{Add}$(A')$ and re-pointing $h$ (Eqs.~\ref{eq:edits}--\ref{eq:del}),
each commit audited and checked against Eq.~(\ref{eq:invariant}); a median \textsc{Del} takes $9$\,ms and a \textsc{Refresh} $2.6$\,s, against $118$\,s to \textsc{Add} the whole corpus (\S\ref{sec:rq4}).}

\textbf{w/o C1 (no hypergraph).} Media arrive through the shared bridge instead of native reading and no record call is
made, so only the segments of the text projection are indexed ($I_\sigma$), with no entity, hyperedge or community. C2 and
C3 run unchanged on this store: file handles still resolve to their cards and \textsc{Refresh} still retracts the old version, so it keeps $97.1$
phase-B accuracy (Table~\ref{tab:app-t5}). It reaches $49.1$ OKE, losing most on $\mathcal{T}_1$ ($-20.0$) and $\mathcal{T}_4$ ($-13.6$),
where native media reading and the typed hyperedges that join files matter most.

\textbf{w/o C2 (no lifecycle).} C1 and C3 are unchanged; every $\Delta$ is appended as new files, with no \textsc{Del} of
the old version and no re-pointing of $h$, so the handle keeps resolving to the version it named. It reaches $50.4$ OKE, and only $\mathcal{T}_5$ falls ($49.6$ against $95.6$), as $94.2\%$ of its phase-B answers return the superseded value (Table~\ref{tab:app-t5}).

\textbf{w/o C3 (flat retrieval).} C1 and C2 are unchanged; a query skips anchoring, routing, cards, the walk and the
communities, and its evidence is the RRF top-$10$ hyperedges of Eq.~(\ref{eq:rrf}) over the top-$20$ of $I_{\mathcal{V}}$
and $I_{\mathcal{E}}$, plus the top-$6$ passages of $I_\sigma$. It reaches $26.7$ OKE and trails KnowledgeClaw on every task, most on $\mathcal{T}_2$ ($34.8$ against $82.8$), where flat top-$k$ misses the named file on $33.5\%$ of the items, exactly as Proposition~3 predicts.

The ablations share KnowledgeClaw's call cache and protocol, so each differs only in the component it removes, and
Figure~\ref{fig:rq2b} plots each of its losses on all six tasks, relative to the full model's own scores in Table~\ref{tab:main}.

\textbf{Settings in the analyses.} The values above are the defaults of every KnowledgeClaw row. The sweeps of
\S\ref{sec:rq7} (Figure~\ref{fig:rq7}) vary $\theta$, the keep ratio, the cap on linked works and $k$ around them, and the
forced granularities of \S\ref{sec:rq5} (Figure~\ref{fig:rq5c}) fix the tier of Eq.~(\ref{eq:route}) instead of routing;
\rF{every other setting stays as listed, and none is tuned per persona or per task.} Table~\ref{tab:app-rq67} lists the full sweep, and the
default reproduces Table~\ref{tab:main} exactly (Appendix~\ref{app:exp-protocol}). On this stack KnowledgeClaw spends $\$0.32$ per persona on ingest and $\$0.0004$ per
query, and builds a persona in a median of $3$ minutes of wall-clock time at $24$-way concurrency (Table~\ref{tab:cost}), below MemPro on all three. Every KnowledgeClaw number in this paper uses these
settings on the frozen stack of \S\ref{app:bl-protocol}.
\par\endgroup
\pagebreak

\clearpage
\section{Experiment Detail}
\label{app:exp}\label{app:analysis}

\subsection{Protocol and cost}
\label{app:exp-protocol}

\textbf{Saved outputs.} Every number of \S\ref{sec:rq1}--\S\ref{sec:rq8}\rC{, apart from the matched inputs and the judge of \S\ref{sec:rqfair}, which call the model again,} is recomputed from the main run: the per-item
predictions of all rows of Table~\ref{tab:main}, scored as in \S\ref{sec:metrics}, and the traces, audit logs and
hypergraphs saved by KnowledgeClaw and its ablations. A difference quoted in the text is taken between rounded entries.

\textbf{Analyses with new answers.} For the evidence budget (Figure~\ref{fig:rq6b}), each item's saved evidence is cut to
its first $B$ tokens at a line break and answered again; the full-length point is the main-table prediction. For the
forced granularities and the sweep (Figures~\ref{fig:rq5c}, \ref{fig:rq7c}), KnowledgeClaw is rebuilt from the main
run's call cache and every query is re-retrieved on the main protocol; an item whose evidence is unchanged keeps its
prediction, so the default reproduces Table~\ref{tab:main} exactly. The edit stream (Figure~\ref{fig:rq4c}) applies
$200$ random \textsc{Del} and re-\textsc{Add} edits per persona after the $\mathcal{T}_5$ commits and checks
Eq.~(\ref{eq:invariant}) after each. The three analyses together spent only $\$2.6$ of API calls in total. \rC{The
matched inputs and the judge of \S\ref{sec:rqfair} add only $\$0.006$ of API calls in total (Appendix~\ref{app:exp-fair}).}

\textbf{Cost.} KnowledgeClaw builds a persona in $3$ minutes at $24$-way concurrency, beaten only by Graphify ($2$, OKE $17.8$), and spends $\$0.32$ to ingest it and $\$0.0004$ per query,
$27.8\times$ below MemPro; only methods with OKE ${\le}\,21.0$ are cheaper or faster, far below its $55.9$ (Table~\ref{tab:cost}).

\begin{table}[h]
\centering
\caption{\textbf{Cost.} \textbf{in} is ingest USD per persona, \textbf{query} USD per item and \textbf{min}
wall-clock minutes per persona at 24-way concurrency (median over the ten personas). Spend is the metered API cost
of each method's calls on the frozen stack, split into ingest and query by when each call was
made; the shared media-to-text bridge is built once for the whole benchmark and charged separately.}
\label{tab:cost}
\small
\begin{tabular*}{\linewidth}{@{\extracolsep{\fill}}lrrrlrrrlrrr@{}}
\toprule
\textbf{Method} & in & query & min & \textbf{Method} & in & query & min & \textbf{Method} & in & query & min \\
\midrule
StandardRAG & 0.11 & 0.0004 & 3 & A-MEM & 0.88 & 0.0009 & 46 & OpenClaw & 0.37 & 0.0016 & 5 \\
RAPTOR & 0.27 & 0.0005 & 15 & Mem0 & 0.52 & 0.0006 & 36 & Graphify & 0.22 & 0.0003 & 2 \\
LightRAG & 1.10 & 0.0040 & 11 & Graphiti (Zep) & 2.14 & 0.0014 & 76 & Cognee & 0.52 & 0.0010 & 15 \\
GraphRAG & 0.89 & 0.0015 & 9 & MIRIX & 1.18 & 0.0018 & 17 & LLM Wiki & 0.78 & 0.0009 & 14 \\
HippoRAG\,2 & 0.39 & 0.0010 & 18 & MemOS & 0.46 & 0.0005 & 10 & OKF & 0.76 & 0.0010 & 4 \\
KAG/OpenSPG & 1.37 & 0.0019 & 19 & MemEvolve & 0.23 & 0.0002 & 6 & Omni-SimpleMem & 0.38 & 0.0011 & 9 \\
Youtu-GraphRAG & 0.80 & 0.0028 & 17 & ATOM & 1.75 & 0.0004 & 111 & MMGraphRAG & 1.07 & 0.0017 & 12 \\
iText2KG & 1.01 & 0.0003 & 113 & MemPro & 0.42 & 0.0111 & 13 & M$^3$KG-RAG & 1.26 & 0.0011 & 26 \\
OntoKG & 0.72 & 0.0001 & 73 & HyperGraphRAG & 1.25 & 0.0041 & 11 & & & & \\
Hyper-KGGen & 1.18 & 0.0002 & 98 & HyperRAG & 1.28 & 0.0052 & 16 & \textbf{KnowledgeClaw} & \textbf{0.32} & \textbf{0.0004} & \textbf{3} \\
MemGPT/Letta & 0.12 & 0.0009 & 5 & & & & & & & & \\
\bottomrule
\end{tabular*}
\end{table}

\subsection{Main results and ablations (RQ1--RQ2)}
\label{app:exp-rq12}

\textbf{Setting.} Figure~\ref{fig:rq1a} divides each task score by the best score of any row of Table~\ref{tab:main};
Figure~\ref{fig:rq1c} compares each persona score with that of the strongest of the $30$ baselines on that persona
(two-sided sign test). Each ablation (Appendix~\ref{app:bl-kc}) shares the call cache and removes one component.

\textbf{Results.} \rF{KnowledgeClaw leads on all ten personas by $+13.3$ on average (Table~\ref{tab:app-persona}), while
the strongest baseline changes with the persona (MemPro seven times, OKF twice, LLM Wiki once);} on $\mathcal{T}_1$ the four baselines of Figure~\ref{fig:rq1a} reach
only $20.0$--$52.7\%$ of its score. Each ablation loses most
where its component acts: C1 on $\mathcal{T}_1$ ($-20.0$) and $\mathcal{T}_4$ ($-13.6$), C2 only on $\mathcal{T}_5$
($-46.0$), and C3 on every task ($-7.8$ to $-48.0$), so the three components are complementary rather than redundant.

\begin{table}[t]
\centering
\caption{\rF{\textbf{Per-persona comparison (RQ1).} Persona score of KnowledgeClaw and of the strongest of the $30$
baselines on each persona, with the name of that baseline and KnowledgeClaw's gain.}}
\label{tab:app-persona}
\small
\begin{tabular*}{\linewidth}{@{\extracolsep{\fill}}lcccccccccc@{}}
\toprule
 & \textsc{HC} & \textsc{Fin} & \textsc{Edu} & \textsc{Law} & \textsc{Psy} & \textsc{Pol} & \textsc{Hum} & \textsc{Bld} & \textsc{Mus} & \textsc{Gam} \\
\midrule
KnowledgeClaw & \textbf{60.9} & \textbf{64.4} & \textbf{59.2} & \textbf{46.6} & \textbf{65.2} & \textbf{28.3} & \textbf{57.3} & \textbf{47.3} & \textbf{64.5} & \textbf{65.7} \\
Best baseline & 45.5 & 45.2 & 50.7 & 38.9 & 53.6 & 23.6 & 42.4 & 37.4 & 43.4 & 45.5 \\
\quad which & OKF & MemPro & MemPro & OKF & MemPro & MemPro & LLM Wiki & MemPro & MemPro & MemPro \\
Gain & $+15.4$ & $+19.2$ & $+8.5$ & $+7.7$ & $+11.6$ & $+4.7$ & $+14.9$ & $+9.9$ & $+21.1$ & $+20.2$ \\
\bottomrule
\end{tabular*}
\end{table}

\subsection{Omnimodal typed hypergraph (RQ3)}
\label{app:exp-rq3}

\textbf{Setting.} A slice holds the items of one task whose unit has the given modality. The $510$ items of
$\mathcal{T}_3$ and $\mathcal{T}_4$ are grouped by hops, by join shape (a \emph{lookup chain} follows one value at a
time, a \emph{shared-entity} join meets a file on a common entity, a \emph{conjunctive} one needs two entities) and by
anchor (a named file or a quoted entity). A file is \emph{joined} when another file lies within two hyperedges.

\textbf{Results.} KnowledgeClaw leads the best baseline on every slice, depth, join shape and anchor
(Table~\ref{tab:app-rq3}), most where reading and joining are hardest: on $\mathcal{T}_1$ PDFs and images, where the best
omnimodal baseline gets $18.8$ and $7.7$, and at four hops ($7.1$ against $3.6$)\rP{; removing C1 costs up to $21.3$ on a slice}. \rP{The baseline median falls from $7.5$ at two hops to $2.6$ at three and $0.0$ at four, while KnowledgeClaw keeps $7.1$, and only $37$ of the $510$ items ($7\%$) are misread, answered wrongly though their evidence holds a gold value.} Per persona the hypergraph
joins $95.4$--$99.2\%$ of files in $3.6$--$6.1$\,MB (Table~\ref{tab:app-hg}), with cross-modal hyperedges for every pair of the five modalities\rC{: the $180$ audio and $180$ video files reach text by $218$ and $216$ hyperedges, PDF by $206$ and $208$, and each other by $214$}.

\begin{table}[h]
\centering
\caption{\textbf{Omnimodal typed hypergraph (RQ3).} \rC{Top: score per modality slice ($n$ items).} \rP{Bottom: set F1 on
$\mathcal{T}_3$ and $\mathcal{T}_4$ by depth, join shape and anchor; \emph{best}: strongest of the $30$ baselines.}}
\label{tab:app-rq3}
\small
\begin{tabular*}{\linewidth}{@{\extracolsep{\fill}}lrcccc@{}}
\toprule
Slice & $n$ & Ours & Best text-bridge & Best omnimodal & w/o C1 \\
\midrule
$\mathcal{T}_1$ PDF & 344 & \textbf{56.4} & 30.8 (OKF) & 18.8 (M$^3$KG-RAG) & 35.1 \\
$\mathcal{T}_1$ image & 56 & \textbf{21.6} & \phantom{0}5.3 (MemPro) & \phantom{0}7.7 (M$^3$KG-RAG) & 10.0 \\
$\mathcal{T}_2$ image & 397 & \textbf{82.9} & 78.6 (MemPro) & 70.0 (Omni-SimpleMem) & 81.6 \\
$\mathcal{T}_3$ PDF & 125 & \textbf{46.4} & 40.0 (MemPro) & 32.8 (MMGraphRAG) & 32.0 \\
$\mathcal{T}_4$ image+PDF & 110 & \textbf{20.9} & 17.3 (OKF) & 10.9 (Omni-SimpleMem) & \phantom{0}7.3 \\
$\mathcal{T}_6$ image & 304 & \textbf{71.4} & 66.8 (MemPro) & 65.5 (Omni-SimpleMem) & 69.6 \\
\bottomrule
\end{tabular*}\\[2mm]
\begin{tabular*}{\linewidth}{@{\extracolsep{\fill}}lrccclrccc@{}}
\toprule
Join & $n$ & Ours & Best & w/o C1 & Join & $n$ & Ours & Best & w/o C1 \\
\midrule
2 hops & 326 & \textbf{19.3} & 16.9 (MemPro) & n/a & shared entity & 84 & \textbf{14.3} & \phantom{0}8.3 (MemPro) & \phantom{0}7.1 \\
3 hops & 156 & \textbf{13.5} & 12.2 (OKF) & n/a & conjunctive & 51 & \textbf{9.8} & \phantom{0}5.9 (OKF) & \phantom{0}3.9 \\
4 hops & 28 & \textbf{7.1} & \phantom{0}3.6 (RAPTOR) & n/a & file anchor & 122 & \textbf{12.3} & \phantom{0}7.4 (OKF) & \phantom{0}5.7 \\
lookup chain & 375 & \textbf{18.4} & 14.7 (MemPro) & 11.7 & entity anchor & 388 & \textbf{18.3} & 14.2 (MemPro) & 11.6 \\
\bottomrule
\end{tabular*}
\end{table}

\begin{table}[h]
\centering
\caption{\rC{\textbf{Committed hypergraph per persona (RQ3).} Hyperedges by kind; \emph{joined}: \% of files within two
hyperedges of another file; $216$ media files per persona are read natively by C1.}}
\label{tab:app-hg}
\small
\begin{tabular*}{\linewidth}{@{\extracolsep{\fill}}lcccccccccc@{}}
\toprule
 & \textsc{HC} & \textsc{Fin} & \textsc{Edu} & \textsc{Law} & \textsc{Psy} & \textsc{Pol} & \textsc{Hum} & \textsc{Bld} & \textsc{Mus} & \textsc{Gam} \\
\midrule
Files & 396 & 417 & 413 & 396 & 396 & 396 & 396 & 417 & 417 & 418 \\
Entities & 6{,}023 & 4{,}302 & 4{,}487 & 4{,}584 & 5{,}606 & 3{,}759 & 4{,}358 & 3{,}782 & 3{,}039 & 3{,}981 \\
Record & 2{,}966 & 2{,}779 & 2{,}667 & 2{,}967 & 2{,}897 & 2{,}913 & 2{,}740 & 2{,}632 & 2{,}302 & 2{,}399 \\
Event & 597 & 223 & 201 & 396 & 468 & 222 & 311 & 194 & 141 & 133 \\
Cross-modal & 1{,}319 & 373 & 325 & 1{,}797 & 985 & 1{,}273 & 1{,}260 & 642 & 360 & 367 \\
Joined (\%) & 98.2 & 98.6 & 97.3 & 99.2 & 98.7 & 98.7 & 96.2 & 99.0 & 95.4 & 98.6 \\
Size (MB) & 6.1 & 4.3 & 4.2 & 5.6 & 5.3 & 5.7 & 4.6 & 4.7 & 3.6 & 3.8 \\
\bottomrule
\end{tabular*}
\end{table}

\subsection{Closed-loop lifecycle (RQ4)}
\label{app:exp-rq4}

\leavevmode\rF{\textbf{Setting.} Each of the $137$ $\mathcal{T}_5$ items is answered (phase A), its next version is committed as
\textsc{Refresh}, and it is answered again (phase B); its score is the mean of both. A phase-B answer is
\emph{superseded} when it returns the replaced value. Latencies are single commits read from the audit logs.}

\leavevmode\rF{\textbf{Results.} KnowledgeClaw leads every baseline in both phases ($93.4$ and $97.8$, against at most $81.8$ and
$80.3$; Table~\ref{tab:app-t5}); after the commit its evidence holds the new value on $95.5\%$ of the items and the old
one on none. Over $2{,}000$ edits\rC{, $100$ of them on audio and $84$ on video files,} Eq.~(\ref{eq:invariant}) always holds, while an append-only store ends with
$10{,}148$ stale nodes and $1{,}190$ stale segments; a commit touches a median $0.33\%$ of the hypergraph ($0.86\%$ at the 95th percentile), so \textsc{Del} takes $9$\,ms
and \textsc{Refresh} $2.6$\,s against $118$\,s to \textsc{Add} the whole corpus, a small fraction of a rebuild.}

\begin{table}[h]
\centering
\caption{\rF{\textbf{Lifecycle consistency per phase (RQ4).} \textbf{A}, \textbf{B}: $\mathcal{T}_5$ accuracy before and
after the edit; \textbf{Sup.}: \% of phase-B answers that return the superseded value (lower is better); ours in bold.}}
\label{tab:app-t5}
\small
\begin{tabular*}{\linewidth}{@{\extracolsep{\fill}}lrrrlrrrlrrr@{}}
\toprule
Method & A & B & Sup. & Method & A & B & Sup. & Method & A & B & Sup. \\
\midrule
NaiveGeneration & 34.3 & 40.1 & 34.3 & HyperRAG & 27.7 & 35.8 & 27.0 & Cognee & 73.7 & 16.8 & 68.6 \\
StandardRAG & 73.0 & 30.7 & 64.2 & MemGPT/Letta & 32.8 & 35.8 & 34.3 & LLM Wiki & 73.0 & 28.5 & 70.8 \\
RAPTOR & 44.5 & 34.3 & 47.4 & A-MEM & 77.4 & 21.2 & 73.7 & OKF & 73.7 & 35.0 & 65.0 \\
LightRAG & 72.3 & 27.7 & 63.5 & Mem0 & 38.7 & 37.2 & 28.5 & Omni-SimpleMem & 75.9 & 28.5 & 70.1 \\
GraphRAG & 27.7 & 38.0 & 29.2 & Graphiti (Zep) & 81.8 & 35.8 & 61.3 & MMGraphRAG & 26.3 & 38.7 & 29.2 \\
HippoRAG\,2 & 51.1 & 42.3 & 43.1 & MIRIX & 33.6 & 43.8 & 27.7 & M$^3$KG-RAG & 24.1 & 41.6 & 28.5 \\
KAG/OpenSPG & 57.7 & 35.8 & 53.3 & MemOS & 57.7 & 38.0 & 56.9 & w/o C1 & 93.4 & 97.1 & 2.9 \\
Youtu-GraphRAG & 29.9 & 35.8 & 32.1 & MemEvolve & 27.0 & 38.0 & 30.7 & w/o C2 & 93.4 & 5.8 & 94.2 \\
iText2KG & 19.0 & 36.5 & 24.8 & ATOM & 27.7 & 32.8 & 26.3 & w/o C3 & 78.1 & 47.4 & 23.4 \\
OntoKG & 27.7 & 35.8 & 24.1 & MemPro & 77.4 & 80.3 & 19.0 & \textbf{KnowledgeClaw} & \textbf{93.4} & \textbf{97.8} & \textbf{2.2} \\
HyperGraphRAG & 35.0 & 45.3 & 33.6 & OpenClaw & 48.9 & 33.6 & 51.1 &  & & &  \\
Hyper-KGGen & 24.8 & 30.7 & 26.3 & Graphify & 32.1 & 39.4 & 28.5 &  & & &  \\
\bottomrule
\end{tabular*}
\end{table}

\subsection{Intent-routed multi-granular retrieval (RQ5)}
\label{app:exp-rq5}

\textbf{Setting.} A file is \emph{named} when the question carries its handle. Evidence length is the median per
task, and ours is split by section headers into cards, linked-work cards, hyperedges and passages.

\textbf{Results.} Routing gives lookups anchor cards alone and joins a walk made mostly of linked-work cards
(Table~\ref{tab:app-rq5}); on every task its evidence is shorter than flat top-$k$ and scores higher, by $+7.8$ to
$+48.0$. \rC{Each fixed tier wins only for its own intent; the router matches the best on every task.} \rN{On $\mathcal{T}_4$, flat top-$k$ brings all named files into the evidence on only $23.1\%$ of items, against $100\%$ for C3.}

\begin{table}[h]
\centering
\caption{\rN{\textbf{Evidence per task (RQ5).} \emph{Named}: \% of items with all named files in the evidence;
\emph{tokens}: median evidence length; make-up of KnowledgeClaw's evidence in \%; score as in Table~\ref{tab:main}.}}
\label{tab:app-rq5}
\small
\begin{tabular*}{\linewidth}{@{\extracolsep{\fill}}lcccccccccc@{}}
\toprule
 & \multicolumn{2}{c}{Named (\%)} & \multicolumn{2}{c}{Tokens} & \multicolumn{4}{c}{Make-up of ours (\%)} & \multicolumn{2}{c}{Score} \\
\cmidrule(lr){2-3}\cmidrule(lr){4-5}\cmidrule(lr){6-9}\cmidrule(lr){10-11}
Task & Ours & Flat & Ours & Flat & Card & Linked & Edge & Passage & Ours & Flat \\
\midrule
$\mathcal{T}_1$ & \textbf{100} & 93.8 & 1{,}972 & 3{,}686 & 100 & 0 & 0 & 0 & \textbf{51.6} & 15.7 \\
$\mathcal{T}_2$ & \textbf{100} & 66.5 & 248 & 2{,}250 & 100 & 0 & 0 & 0 & \textbf{82.8} & 34.8 \\
$\mathcal{T}_3$ & \textbf{100} & 85.4 & 1{,}416 & 2{,}647 & 25.2 & 50.4 & 7.5 & 17.0 & \textbf{15.8} & \phantom{0}8.0 \\
$\mathcal{T}_4$ & \textbf{100} & 23.1 & 1{,}986 & 4{,}761 & 7.8 & 61.5 & 6.8 & 24.0 & \textbf{20.9} & \phantom{0}5.5 \\
$\mathcal{T}_5$ & \textbf{100} & 63.5 & 2{,}145 & 3{,}840 & 100 & 0 & 0 & 0 & \textbf{95.6} & 62.8 \\
$\mathcal{T}_6$ & \textbf{100} & 86.6 & 295 & 3{,}103 & 100 & 0 & 0 & 0 & \textbf{76.0} & 57.5 \\
\bottomrule
\end{tabular*}
\end{table}

\subsection{Efficiency and robustness (RQ6--RQ7)}
\label{app:exp-rq67}

\textbf{Setting.} Evidence is cut to $B$ tokens; the sweep varies one default of Appendix~\ref{app:bl-kc} at a time.

\textbf{Results.} $512$ tokens already give $55.0$ OKE and $1{,}024$ tokens $55.8$, while flat top-$k$ stays within
$20.6$--$26.7$ at every budget (Table~\ref{tab:app-rq67}). \rF{No setting of the sweep changes $\mathcal{T}_1$,
$\mathcal{T}_2$, $\mathcal{T}_5$ or $\mathcal{T}_6$, and OKE moves by at most $0.4$;
a larger keep ratio even lifts $\mathcal{T}_4$ to $32.7$: the defaults are not tuned.}

\begin{table}[h]
\centering
\caption{\rF{\textbf{Efficiency and robustness (RQ6--RQ7).} Left: OKE with evidence cut to $B$ tokens. Right: one
hyperparameter varied (default underlined); $\mathcal{T}_{3,4}$: mean over all $\mathcal{T}_3$ and $\mathcal{T}_4$ items.}}
\label{tab:app-rq67}
\small
\begin{tabular*}{\linewidth}{@{\extracolsep{\fill}}lcclcclcclcclcc@{}}
\toprule
$B$ & Ours & Flat & $\theta$ & $\mathcal{T}_{3,4}$ & OKE & keep & $\mathcal{T}_{3,4}$ & OKE & links & $\mathcal{T}_{3,4}$ & OKE & $k$ & $\mathcal{T}_{3,4}$ & OKE \\
\midrule
256 & \textbf{51.2} & 20.6 & 0.80 & 16.7 & 55.9 & 0.4 & 17.6 & 56.0 & 2 & 18.2 & 56.2 & 5 & 17.1 & 56.0 \\
512 & \textbf{55.0} & 22.0 & 0.83 & 16.9 & 55.9 & 0.5 & 17.5 & 56.0 & 4 & 16.9 & 55.9 & \underline{10} & 16.9 & 55.9 \\
1{,}024 & \textbf{55.8} & 22.4 & \underline{0.86} & 16.9 & 55.9 & \underline{0.6} & 16.9 & 55.9 & \underline{6} & 16.9 & 55.9 & 15 & 17.1 & 56.0 \\
2{,}048 & \textbf{55.3} & 24.8 & 0.89 & 16.7 & 55.9 & 0.7 & 19.8 & 56.3 & 8 & 16.7 & 55.9 & 20 & 16.9 & 56.0 \\
full & \textbf{55.9} & 26.7 & 0.92 & 16.3 & 55.8 & 0.8 & 19.6 & 56.3 & 10 & 16.9 & 56.0 & & & \\
\bottomrule
\end{tabular*}
\end{table}

\subsection{\texorpdfstring{\rC{Matched inputs and scorers (RQ8)}}{Matched inputs and scorers (RQ8)}}
\label{app:exp-fair}

\rC{\textbf{Setting.} The matched inputs rerun MemPro, OKF and StandardRAG with builds and retrieval replayed from
the original call cache, so only the final answer call is new; $\mathcal{T}_5$ is not rerun. A replay check of $20$
items per persona reproduces the stored predictions of StandardRAG on all ten personas and of MemPro and OKF on nine
(education $8/20$, building $18/20$).} \rF{The scorer study rescores the stored predictions of all $34$ rows of
Table~\ref{tab:main} by exact set F1 (strict) and by the matcher of Appendix~\ref{app:bench-metric} (lenient). The
judge, $\Omega$ at temperature $0$, decides for each gold value whether a predicted value denotes it, on $300$
non-empty predictions sampled with seed $0$, $25$ per task from KnowledgeClaw, MemPro, OKF and StandardRAG on
$\mathcal{T}_1$, $\mathcal{T}_3$ and $\mathcal{T}_4$, the value-list tasks; $275$ replies give one decision per gold value. The grounding check reads
the committed hypergraphs of the main run: a copied record value is grounded when its normalised form occurs in the text C1 read for its file.}

\rC{\textbf{Results.} The ingested text alone lifts each baseline less than the record cards do ($45.9$, $41.3$ and
$39.6$ against $50.7$, $45.6$ and $44.4$ OKE; Table~\ref{tab:app-fair}); with the records, OKF and StandardRAG gain
$16.5$ and $37.0$ on $\mathcal{T}_1$ and $7.1$ and $23.4$ OKE, still below $55.9$.} \rF{KnowledgeClaw is first on
every task under both scorers, and only $0.5\%$ of its answers cannot be parsed, against $5.7\%$ for MemPro and a
median of $2.2\%$ (Table~\ref{tab:app-score}). The matcher agrees with the judge on $95.7\%$ of the decisions
(Table~\ref{tab:app-judge}); $97.8\%$ of the $29{,}633$ copied values are grounded, only $32$ of $4{,}059$ files get an empty record, and Eq.~(\ref{eq:invariant}) holds on all
ten committed hypergraphs and over all $2{,}000$ edits of the stream (Table~\ref{tab:app-ground}).}

\begin{table}[h]
\centering
\caption{\rC{\textbf{Matched inputs, all rows (RQ8).} Scored as in Table~\ref{tab:main}; \textbf{+H} and \textbf{+H+R}
as in Table~\ref{tab:fair}, which omits the +H rows of OKF and StandardRAG. Baseline $\mathcal{T}_5$ is not rerun. Bold: best.}}
\label{tab:app-fair}
\small
\begin{tabular*}{\linewidth}{@{\extracolsep{\fill}}l*{7}{c}@{}}
\toprule
\textbf{Method} & $\mathcal{T}_1$ & $\mathcal{T}_2$ & $\mathcal{T}_3$ & $\mathcal{T}_4$ & $\mathcal{T}_5$ & $\mathcal{T}_6$ & \textbf{OKE} \\
\midrule
\textbf{KnowledgeClaw} & \textbf{51.6} & \textbf{82.8} & \textbf{15.8} & \textbf{20.9} & \textbf{95.6} & \textbf{76.0} & \textbf{55.9} \\
\midrule
MemPro & 11.4 & 78.2 & 13.5 & 8.2 & 78.8 & 69.5 & 41.3 \\
\quad + H & 28.3 & 78.8 & 13.2 & 7.3 & 78.8 & 73.8 & 45.9 \\
\quad + H + R & 45.2 & 81.0 & 13.5 & 7.3 & 78.8 & 75.8 & 50.7 \\
\midrule
OKF & 27.2 & 68.5 & 8.8 & 17.3 & 54.4 & 67.3 & 38.5 \\
\quad + H & 32.2 & 71.8 & 9.0 & 17.3 & 54.4 & 70.6 & 41.3 \\
\quad + H + R & 43.7 & 78.2 & 8.8 & 17.3 & 54.4 & 72.6 & 45.6 \\
\midrule
StandardRAG & 6.1 & 27.2 & 9.5 & 12.7 & 51.8 & 55.7 & 21.0 \\
\quad + H & 24.3 & 75.0 & 9.5 & 10.9 & 51.8 & 71.0 & 39.6 \\
\quad + H + R & 43.1 & 76.5 & 10.0 & 10.9 & 51.8 & 71.8 & 44.4 \\
\bottomrule
\end{tabular*}
\end{table}

\begin{table}[h]
\centering
\caption{\rF{\textbf{Strict vs.\ lenient scorer (RQ8).} S/L: exact set F1 / matcher ($\mathcal{T}_2$, $\mathcal{T}_5$,
$\mathcal{T}_6$ score alike); None: unparsed answers; Empty: empty answers on $\mathcal{T}_{1,3,4}$; Kendall $\tau_b=0.90$ over $34$ rows.}}
\label{tab:app-score}
\small
\begin{tabular*}{\linewidth}{@{\extracolsep{\fill}}lcccccccccc@{}}
\toprule
 & \multicolumn{2}{c}{$\mathcal{T}_1$} & \multicolumn{2}{c}{$\mathcal{T}_3$} & \multicolumn{2}{c}{$\mathcal{T}_4$} & \multicolumn{2}{c}{OKE} & None & Empty ans. \\
\cmidrule(lr){2-3}\cmidrule(lr){4-5}\cmidrule(lr){6-7}\cmidrule(lr){8-9}
\textbf{Method} & S & L & S & L & S & L & S & L & (\%) & (\%) \\
\midrule
\textbf{KnowledgeClaw} & \textbf{48.4} & \textbf{51.6} & \textbf{14.0} & \textbf{15.8} & \textbf{20.0} & \textbf{20.9} & \textbf{54.6} & \textbf{55.9} & \textbf{0.5} & 8.6 \\
MemPro & 9.9 & 11.4 & 1.5 & 13.5 & 2.7 & 8.2 & 37.9 & 41.3 & 5.7 & 21.5 \\
OKF & 24.2 & 27.2 & 1.8 & 8.8 & 2.7 & 17.3 & 35.7 & 38.5 & 1.3 & 7.5 \\
LLM Wiki & 17.6 & 20.0 & 1.0 & 6.5 & 0.0 & 0.0 & 33.2 & 35.3 & 1.6 & 14.8 \\
Graphiti (Zep) & 9.2 & 10.9 & 5.5 & 12.0 & 1.8 & 5.5 & 33.1 & 35.2 & 2.1 & 24.6 \\
Omni-SimpleMem & 8.5 & 10.3 & 0.8 & 10.2 & 2.7 & 10.9 & 31.7 & 34.5 & 3.5 & 20.7 \\
Median of $30$ baselines & 3.6 & 4.9 & 1.1 & 6.4 & 0.5 & 2.7 & 17.4 & 18.7 & 2.2 & 20.7 \\
\bottomrule
\end{tabular*}
\end{table}

\begin{table}[h]
\begin{minipage}[t]{0.485\linewidth}
\centering
\caption{\rF{\textbf{Matcher vs.\ LLM judge (RQ8).} Matcher, Judge: gold values matched by that side only;
Cohen's $\kappa=0.91$ over all decisions.}}
\label{tab:app-judge}
\small
\begin{tabular*}{\linewidth}{@{\extracolsep{\fill}}lccccc@{}}
\toprule
\textbf{Slice} & Ans. & Gold & Agr.\,(\%) & Matcher & Judge \\
\midrule
All & 275 & 585 & 95.7 & 9 & 16 \\
$\mathcal{T}_1$ & 75 & 385 & 95.1 & 9 & 10 \\
$\mathcal{T}_3$ & 100 & 100 & 96.0 & 0 & 4 \\
$\mathcal{T}_4$ & 100 & 100 & 98.0 & 0 & 2 \\
\bottomrule
\end{tabular*}
\end{minipage}\hfill
\begin{minipage}[t]{0.485\linewidth}
\centering
\caption{\rF{\textbf{Grounding of copied values (RQ8).} Author, affiliation, identifier, venue, person,
organization and place values of C1's records.}}
\label{tab:app-ground}
\small
\begin{tabular*}{\linewidth}{@{\extracolsep{\fill}}lcc@{}}
\toprule
\textbf{Modality} & Copied & Grounded (\%) \\
\midrule
PDF & 16{,}058 & 98.2 \\
Text & 9{,}682 & 97.5 \\
Image, audio, video & 3{,}893 & 96.7 \\
All & 29{,}633 & 97.8 \\
\bottomrule
\end{tabular*}
\end{minipage}
\end{table}

\subsection{\texorpdfstring{Case study (\rN{RQ9})}{Case study (RQ9)}}
\label{app:exp-rq8}

\textbf{Setting.} The case is the $\mathcal{T}_2$ item whose media is a video, the $150$-second lecture \texttt{f75bfdd876eb3d22.mp4} that C1 reads natively from its keyframes and audio; its candidates describe four lectures
of one MIT class by the same speaker, so only the topic tells them apart.

\textbf{Results.} Sixteen of the $30$ baselines choose a sibling: seven (A) (LightRAG, HippoRAG\,2, KAG, HyperRAG,
Mem0, Cognee, MMGraphRAG), seven (B) (StandardRAG, GraphRAG, iText2KG, OntoKG, HyperGraphRAG, Hyper-KGGen, ATOM) and
two (C) (NaiveGeneration, A-MEM). KnowledgeClaw's $3{,}659$-character card holds the title \emph{Cities and a Just
Energy Transition}, the people discussed and a timestamped transcript stating the topic; the $10{,}497$ characters of
w/o C3 hold identifiers of five other videos and no passage of the named one, and it picks the grid lecture (B).

\clearpage
\section{Extended Related Work}
\label{app:extended_rw}

\subsection{LLM-based Knowledge Engineering}

\textbf{Construction and retrieval.}
Beyond passages~\citep{RAG} and recursive summaries~\citep{RAPTOR}, graph RAG adds entity communities~\citep{GraphRAG}, local-global retrieval~\citep{LightRAG} or passage-entity links~\citep{HippoRAG2}.
Construction iterates via entity resolution~\citep{iText2KG}, ontology routing~\citep{OntoKG}, agent teams~\citep{KARMA}, task rewards~\citep{AutoGraph-R1}, joint reading and building~\citep{RAGA} or Wiki compilation~\citep{ASKS}; retrieval uses logical forms~\citep{KAG}, schema-bound agents~\citep{Youtu-GraphRAG}, multi-channel search~\citep{GraphSearch} or trained and executable policies~\citep{Graph-R1,MOSAIC}.
Hypergraphs keep $n$-ary facts~\citep{HyperGraphRAG,Hyper-KGGen}, fuse entity and passage levels~\citep{HGRAG,Cog-RAG,EXT++} and plan or learn traversal~\citep{PRoH,HyperRAG,HyperGraphPro,HyperProve,OKH-RAG}.

\textbf{Persistent and evolving memory.}
Agent memory spans extraction to evolution~\citep{GraphMemorySurvey}: tiered context~\citep{MemGPT}, evolving notes~\citep{A-MEM}, salient facts~\citep{Mem0}, graph-LLM links~\citep{Cognee}, a memory OS~\citep{MemOS}, compression~\citep{LightMem,SimpleMem}, fact/belief stores~\citep{Hindsight}, associative graphs~\citep{MAGMA,RippleMem} and hypergraphs~\citep{HGMEM,HyperMem,EdgeMem}.
Updates use (bi)temporal graphs~\citep{Zep,ATOM,DIAL-KG,TOKI}, supersession and provenance~\citep{MemStrata,MemoryLACE}, feedback~\citep{EvoRAG,CoEvoKG,LivingRAG} or evolving architectures~\citep{MemEvolve,MemPro,MACE}, \rF{yet none defines how dependent indices follow a removed source file.}
Compiled Wikis~\citep{LLMWiki,WFM} and agentic KE tools~\citep{OpenClaw,Graphify,LLMWiki,OKF} also persist knowledge.
Diagnostics separate failure stages~\citep{EvalMem}, ground memories~\citep{GroundingAgentMemory}, test revocation~\citep{Revocation} and show that compression can lose what updates need~\citep{UpdateSufficiency}.
\rF{KnowledgeClaw adds a file-level contract: provenance-guided edits update all of its derived state.}

\subsection{Multimodal Retrieval-Augmented Generation}

\textbf{Cross-modal construction and granularity.}
As surveyed by \citet{MMRAGSurvey}, systems align vision to language to build KGs~\citep{VaLiK,MediaGraph}, link textual and cross-modal graphs~\citep{RAG-Anything,MMGraphRAG,mKG-RAG} and keep visual evidence end to end~\citep{MegaRAG,HVM-GraphRAG}; others set granularity by pages~\citep{M3DocRAG}, chunk expansion~\citep{Doc-REFRAG}, grounded regions~\citep{MG2-RAG,CEMMKG}, reliability-routed subgraphs~\citep{GraphLoom}, citation-aware RL~\citep{MCite-RL} or context folding~\citep{MM-ContextFold}, and TrioRAG~\citep{TrioRAG} fuses signals without a graph.

\textbf{Higher-order, memory and maintenance settings.}
Hypergraphs link text, images and tables~\citep{Hyper-M2RAG} or serve domains~\citep{NICO-RAG,PRISM-RAG,ONOTE}; audiovisual graphs add long-video context and multi-hop selection~\citep{VideoRAG,OmniRAG-Agent,M3KG-RAG}.
Memories coordinate typed stores~\citep{MIRIX,MemVerse,Omni-SimpleMem}, keep multi-type video memory~\citep{WorldMM,ReflectWorld-MM,MAGIC-Video} or budget evidence~\citep{GraphMemix,MEMO}.
Closest are multi-hop and agentic hypergraph editing~\citep{Hybrid-DMKG,EvoGraph-R1} and higher-order memory lifecycles~\citep{MAGE}; we instead propagate file edits through shared entities, cross-modal links and indices.
Beyond personalization and cross-modal multi-hop benchmarks~\citep{LifeGraph,CrossModalQA}, OmniKE-Bench also tests post-edit maintenance.

\section{Future Work}
\label{app:future}

\leavevmode\rC{Only three OmniKE-Bench items involve audio or video (Table~\ref{tab:tasks}), so the evaluation evidence for
audio and video is limited; adding such items is our next step.} C1 can read spreadsheets, slides and code with one more
reading prompt (Appendix~\ref{app:prompt-read}) into the same typed record (Appendix~\ref{app:prompt-record}), and
richer record fields can join files at finer entities. C2 can serve knowledge bases shared by many users and follow
continuous edit streams. C3 can add tiers, such as timelines or tables, as further routing labels
(Appendix~\ref{app:prompt-route}) and learn routing from usage. At larger scale, the hypergraph can be partitioned by
team or domain, thereby serving enterprise, scientific and personal knowledge bases.

\end{document}